\documentclass{article}

\usepackage{microtype}
\usepackage{graphicx}
\usepackage{subfigure}
\usepackage{booktabs} 

\usepackage{algorithm}
\usepackage[utf8]{inputenc} 
\usepackage[T1]{fontenc}    
\usepackage{url}            
\usepackage{booktabs}       
\usepackage{amsfonts}       
\usepackage{nicefrac}       
\usepackage{microtype}      
\usepackage[table]{xcolor}         
\usepackage{natbib}
\usepackage[
  colorlinks,
  citecolor=blue,
  linkcolor=red,
  anchorcolor=red,
  urlcolor=blue
]{hyperref}    
\usepackage{authblk}
\usepackage{multirow}

\usepackage{color-edits}
\definecolor{redorange}{RGB}{255, 68, 51}
\addauthor[Ruohan]{rz}{redorange}

\usepackage{bbm}
\usepackage{bm}
\usepackage{pifont}
\usepackage{makecell}
\newcommand{\cmark}{\ding{51}}%
\newcommand{\xmark}{\ding{55}}%

\usepackage{hyperref}

\usepackage[accepted]{icml2025}

\usepackage{amsmath}
\usepackage{amssymb}
\usepackage{mathtools}
\usepackage{amsthm}
\usepackage[capitalize,noabbrev]{cleveref}

\theoremstyle{plain}
\newtheorem{theorem}{Theorem}[section]
\newtheorem{proposition}[theorem]{Proposition}
\newtheorem{lemma}[theorem]{Lemma}

\theoremstyle{definition}
\newtheorem{definition}[theorem]{Definition}
\newtheorem{assumption}[theorem]{Assumption}
\theoremstyle{remark}
\newtheorem{remark}[theorem]{Remark}

\usepackage[textsize=tiny]{todonotes}

\renewcommand{\hat}{\widehat}
\newcommand{\ind}{\,\rotatebox[origin=c]{90}{$\models$}\,}

\newcommand{\PP}{\mathbb{P}}

\newcommand{\EE}{\mathbb{E}}
\newcommand{\ZZ}{\mathbb{Z}}

\newcommand{\D}{\mathcal{D}}
\newcommand{\A}{\mathcal{A}}

\newcommand{\Dk}{\mathcal{D}^{(k)}}

\newcommand{\V}{\mathcal{V}}
\newcommand{\Vd}{\mathcal{V}_{\delta}}
\newcommand{\hVd}{\widehat{\mathcal{V}}_{\delta}}
\newcommand{\hVdl}{\widehat{\mathcal{V}}_{\delta}^{\text{LN}}}

\newcommand{\hVdk}{\widehat{\mathcal{V}}_{\delta}^{(k)}}
\newcommand{\dd}{\delta}
\newcommand{\al}{\alpha}
\newcommand{\ve}{\varepsilon}
\newcommand{\po}{\pi_0}
\newcommand{\als}{\alpha^*}
\newcommand{\etas}{\eta^*}

\newcommand{\hGpik}{\widehat{G}_\pi^{(k)}}

\newcommand{\bgpi}{\bar{g}_\pi}

\newcommand{\hgpik}{\hat{g}_\pi^{(k)}}

\newcommand{\hapik}{\hat{\alpha}_\pi^{(k)}}

\newcommand{\hpik}{\widehat{\pi}_0^{(k)}}
\newcommand{\hpi}{\widehat{\pi}_{\text{LN}}}
\newcommand{\given}{\,| \,}
\newcommand{\biggiven}{\, \big| \,}
\newcommand{\Biggiven}{\, \Big| \,}
\newcommand{\bigggiven}{\, \bigg| \,}

\newcommand{\Regret}{\mathcal{R}_{\dd}}

\newcommand{\thes}{\theta^*}
\newcommand{\hepik}{\hat{\eta}_\pi^{(k)}}

\newcommand{\X}{\mathcal{X}}

\newcommand{\kl}{\textnormal{KL}}
\newcommand{\LN}{\textnormal{LN}}
\newcommand{\bmalpha}{\bm{\alpha}}
\newcommand{\bmeta}{\bm{\eta}}
\newcommand{\bmtheta}{\bm{\theta}}

\newcommand{\argmin}[1]{\underset{#1}{\arg\!\min}}
\newcommand{\argmax}[1]{\underset{#1}{\arg\!\max}}
\def\##1\#{\begin{align}#1\end{align}}
\def\$#1\${\begin{align*}#1\end{align*}}
\newcommand{\cD}{\mathcal{D}}
\newcommand{\indc}{\mathbbm{1}}
\newcommand{\cbnd}{C_0(\bar{\alpha}, \underline{\alpha}, \bar{\eta}, \delta,\ve)}
\newcommand{\cbndb}{C_1(\bar{\alpha}, \underline{\alpha}, \bar{\eta}, \delta,\ve)}
\newcommand{\cbndc}{C_2(\bar{\alpha}, \underline{\alpha}, \bar{\eta}, \delta,\ve)}

\icmltitlerunning{Distributionally Robust Policy Learning under Concept Drifts}

\begin{document}

\twocolumn[
\icmltitle{Distributionally Robust Policy Learning under Concept Drifts}



\icmlsetsymbol{equal}{*}

\begin{icmlauthorlist}
\icmlauthor{Jingyuan Wang}{nyu}
\icmlauthor{Zhimei Ren}{upenn}
\icmlauthor{Ruohan Zhan}{ukst}
\icmlauthor{Zhengyuan Zhou}{nyu,arena}
\end{icmlauthorlist}

\icmlaffiliation{nyu}{Stern School of Business, New York University}
\icmlaffiliation{upenn}{Department of Statistics and Data Science, University of Pennsylvania}
\icmlaffiliation{ukst}{IEDA, Hong Kong University of Science and Technology}
\icmlaffiliation{arena}{Arena Technologies}

\icmlcorrespondingauthor{Zhengyuan Zhou}{zhengyuanzhou24@gmail.com}

\icmlkeywords{Machine Learning, ICML}

\vskip 0.3in
]



\printAffiliationsAndNotice{}  

\begin{abstract}
Distributionally robust policy learning aims to find a policy that performs well 
    under the worst-case distributional shift, and yet most existing methods for 
    robust policy learning consider the worst-case {\em joint} distribution of 
    the covariate and the outcome. The joint-modeling strategy can be unnecessarily conservative
    when we have more information on the source of distributional shifts. This paper studies
    a more nuanced problem --- robust policy learning under the \emph{concept drift}, 
    when only the conditional relationship between the outcome and the covariate changes. 
    To this end, we first provide a doubly-robust estimator for evaluating
    the worst-case average reward of a given policy under a set of perturbed conditional distributions. 
    We show that the policy value estimator enjoys asymptotic normality even if the nuisance parameters 
    are estimated with a slower-than-root-$n$ rate.
    We then propose a learning algorithm that outputs the policy maximizing the 
    estimated policy value within a given policy class $\Pi$, and show
    that the sub-optimality gap of the proposed algorithm is of the order 
    $\kappa(\Pi)n^{-1/2}$, where $\kappa(\Pi)$ is the entropy integral of $\Pi$ under the Hamming distance
    and $n$ is the sample size. A matching lower bound is provided to show the optimality of the rate.
    The proposed methods are implemented and evaluated in numerical studies, 
    demonstrating substantial improvement compared with existing benchmarks.
\end{abstract}

\begin{table*}[t]
\centering
\begin{tabular}{c|c|c|c|c|c}
\toprule 
& \makecell{Distribution shift} & \makecell{Unknown $\pi_0$} & 
\makecell{General $\mathcal{X}$} & \makecell{Upper bound} & \makecell{Lower bound}\\
\midrule 
\citet{athey2021policy} & \xmark & \cmark & \cmark & --- & --- \\
\citet{zhou2023offline} & \xmark & \cmark & \cmark & --- & --- \\
\citet{si2023distributionally} & Joint & \xmark & \cmark & --- & --- \\
\citet{kallus2022doubly} & Joint & \cmark & \cmark & --- & ---\\
\midrule
\citet{mu2022factored} & Separate & \cmark & \xmark & $O\big(\sqrt{\frac{\log n \log(|\mathcal{X}||\mathcal{A}|)}{n}}\big)$ & \xmark\\
This work & Separate & \cmark & \cmark & $O\big({\frac{\kappa(\Pi)}{\sqrt{n}}}\big)$ & 
$\Omega\big(\sqrt{\frac{\text{Ndim}(\Pi)}{n}}\big)$\\
\bottomrule
\end{tabular}
\caption{Comparison of results in the offline policy learning literature. 
``Unknown $\pi_0$'' refers to whether an algorithm assumes knowledge of the 
behavior policy $\pi_0$. ``General $\mathcal{X}$'' refers to whether an algorithm 
allows for general types of covariates.~\citet{athey2021policy,zhou2023offline,si2023distributionally,kallus2022doubly}
have the regret upper and lower bounds for the specific problems they consider that 
are not directly comparable to ours, so we do not include them in the table. 
$|\mathcal{X}|$ refers to the cardinality of the covariate support 
(if finite) and $|\mathcal{A}|$ to that of the action set. $\kappa(\Pi)$ and $\text{Ndim}(\Pi)$
are the entropy integral under Hamming distance and the Natarajan dimension of a policy class 
$\Pi$, with the relation $\kappa(\Pi) = O(\log(d) \text{Ndim}(\Pi))$, where $d$ is the dimension of the covariate 
space.
}
\label{tav:summary}
\end{table*}

\section{Introduction}
\label{sec:intro}
In a wide range of fields, the abundance of 
user-specific historical data provides opportunities
for learning efficient individualized policies. Examples 
include learning the optimal personalized 
treatment from electronic health record data~\citep{murphy2003optimal, kim2011battle, chan2012optimizing}, 
or obtaining an individualized advertising strategy
using past customer behavior data~\citep{bottou2013counterfactual,kallus2016dynamic}.
Driven by such a practical need, a line of works have been devoted to 
developing efficient policy learning algorithms using 
historical data --- a task often known as 
{\em offline policy learning}~\citep{dudik2011doubly,zhang2012estimating,swaminathan2015batch, swaminathan2015counterfactual,swaminathan2015self,kitagawa2018should,athey2021policy,zhou2023offline,zhan2023policy,bibaut2021risk,jin2021pessimism,jin2022policy}. 

Most existing methods for offline policy learning deliver performance guarantees 
under the premise that the  target environment remains 
the same as that from which the historical data is collected.
It has been widely observed, however, that 
such a condition is hardly met in practice
(see e.g.,~\citet{recht2019imagenet,namkoong2023diagnosing,liu2023need,jin2023diagnosing}
and the references 
therein). Under distribution shift, a policy learned in one environment 
often shows degraded performance when deployed in another environment. 
To address this issue, there is an emerging body of research 
on {\em robust policy learning}, which aims at finding
a policy that still performs well when the target distribution is 
perturbed. 
Pioneering works in this area 
consider the case where the {\em joint distribution} of the covariates and 
the outcome is shifted from the training distribution, and researchers devise  
algorithms that output a policy achieving reliable worst-case 
performance under the aforementioned shifts~\citep{si2023distributionally,kallus2022doubly}. 
The joint modeling approach, however,   
ignores the {\em type} of distributional shifts, and the resulting worst-case value can be unnecessarily conservative in practice.

Indeed, distributional shifts can be categorized into 
two classes by their sources: (1) the shift in the covariate $X$,  
and/or (2) the shift in the conditional 
relationship between the outcome $Y$ and the covariate $X$.
The two types of distributional shifts are different in nature, 
have different implications on the objectives,
and call for distinct treatment~\citep{namkoong2023diagnosing,liu2023need,jin2023diagnosing,ai2024not}. 
To be concrete, imagine that the distribution of covariates changes while that of 
$Y\given X$ remains invariant --- in this case, the distribution shift 
is identifiable/estimable since the covariates are often accessible in the target environment.  
As a result, it is often unnecessary to account for the worst-case covariate shift 
rather than directly correcting for it. 
Alternatively, when the $Y\given X$ distribution changes 
but the $X$ distribution remains invariant, the distribution shift is
no longer identifiable, where we can instead apply the worst-case consideration to guarantee performance.
This latter setting, known as  {\em concept drift},  occurs
due to sudden external shocks~\citep{widmer1996learning,lu2018learning,gama2014survey}.
For example, in advertising, the customer behavior can evolve over time 
as the environment changes, while the population remains largely the same.
In personalized product recommendation, similar population segments in developed and emerging markets may prefer different product features.
In these applications, with the one extra bit of information that 
the shift is only in the conditional reward distribution, can we obtain 
a more efficient policy learning algorithm?

Motivated by the above situations, 
we study robust policy learning under concept drift. 
Most existing methods for robust policy learning~\citep{si2023distributionally,kallus2022doubly}  
model the distributional shift jointly without distinguishing 
the sources, and the corresponding algorithms turn out to be suboptimal. 
The reason behind their suboptimality
is that the worst-case distributions under the two models --- the 
joint-shift model and the concept-drift model ---
can be substantially different, so it would be a ``waste'' of our budget to  
consider adversarial distributions that are not feasible under concept drift. 
It is worth mentioning that a recent paper by~\citet{mu2022factored} 
accounts for the sources of distributional shifts in policy learning; their approach,
however, applies only when the covariates take {\em a finite number of 
values}, and therefore is limited in its applicability.  
When the covariate space is infinite, it remains unclear 
how to efficiently learn a robust policy under concept drift.
The current work aims to fill in the gap by answering the question:\emph{How can we efficiently learn a policy with 
optimal worst-case average performance under concept drift 
with minimal assumptions?}
We provide a rigorous answer to the above question.
Specifically, we assume the covariate distribution remains
the same in the training and target environments,\footnote{Otherwise, the covariate shift can be easily adjusted by covariate matching discussed earlier.} while the 
$Y\given X$ distribution shift is bounded in KL-divergence by a
pre-specified constant $\delta$. Our goal is to find a 
policy that maximizes the worst-case averaged outcome 
over all possible target distributions satisfying the previous condition.

\subsection{Our Contributions}\label{sec:our-contributions}
Towards robust policy learning under concept drift, we make 
the following contributions.

\textbf{Policy Evaluation.} Given a policy, we present a doubly-robust estimator for the 
worst-case policy value under concept drift. 
We prove that the estimator is 
asymptotic normal under mild conditions on the estimation rate of the 
nuisance parameter.
Our approach involves solving the dual form of a distributionally 
robust optimization problem and  
taking a de-biased step to deal with the slow convergence of the optimizer, 
thereby obtaining an estimator with root-n convergence rate.

\textbf{Policy Learning.} We propose
a robust policy learning algorithm that outputs a policy 
maximizing the estimated policy value over a policy class $\Pi$.
Compared with the oracle optimal policy, 
the policy provided by our algorithm with high probability 
has a regret/suboptimality gap of the order $\kappa(\Pi)/ \sqrt{n}$, 
where $\kappa(\Pi)$ is a measure quantifying the policy class complexity (to be formalized 
shortly) and $n$ is the number of samples. Compared with~\citet{mu2022factored}, 
our algorithm and theory apply to general covariate spaces and potentially 
infinite policy classes, while their method is restricted to finite covariate space and policy class. We complement the upper bound with a matching lower bound, thus 
establishing the minimax optimality of our proposed algorithm. 
We summarize the comparison between our result and prior work in Table~\ref{tav:summary}
for better demonstration.

\textbf{Implementation and Empirics.} We provide efficient implementation 
of our robust policy learning algorithm, and compare its empirical 
performance with existing benchmarks in numerical studies. Our 
proposed method exhibits substantial improvement.

\subsection{Related Works}\label{sec:related-works}
\textbf{Offline Policy Learning.}
There is a long list of works devoted to
offline policy learning. Most of them assume 
no distributional shifts (e.g.,~\citet{dudik2011doubly,zhang2012estimating,swaminathan2015batch,swaminathan2015counterfactual,swaminathan2015self,kitagawa2018should,
athey2021policy,zhou2023offline}).~\citet{zhan2023policy,jin2021pessimism,jin2022policy}
allow the data to be adaptively collected, but the distribution 
over the covariate and the (potential) outcomes remain invariant in 
the training and target environment. 

As mentioned earlier, \citet{si2023distributionally,kallus2022doubly} study 
robust policy learning when the joint distribution of $(X,Y)$
ranges in the neighborhood of the training distribution;~\citet{mu2022factored}
consider the case when the covariate shift and $Y\given X$ shift 
are specified separately; their method, however, is restricted to finite 
covariate space, and their sub-optimality gap is logarithmic factors slower than parametric rates. 
More recently,~\citet{guo2024distributionally} considers  
a pure covariate shift with a focus on policy evaluation, where the setup and the goal are different from ours.

\textbf{Distributionally Robust Optimization.}
More broadly, 
our work is closely related to DRO, where the goal is to 
learn a model that has good performance under the 
worst-case distribution (e.g.,~\citet{bertsimas2004price,delage2010distributionally,hu2013kullback,duchi2019distributionally,dudik2011doubly,zhang2023optimal}).
The major focus of the aforementioned works involves parameter estimation and 
prediction in supervised settings; we however take a decision-making perspective
and aim at learning an individualized policy with optimal worst-case performance guarantees.

\section{Preliminaries}\label{sec:problem-formulation}
Consider a set of $M$ actions denoted by $[M]$
and let $\mathcal{X}\subseteq\mathbb{R}^d$. 
Throughout the paper, we follow the potential outcome framework~\citep{imbens2015causal}, 
where $Y(a)\in\mathcal{Y}_a\subseteq\mathbb{R}$ denotes the 
potential outcome had action $a$ been taken for any $a \in [M]$. 
We posit the underlying data-generating distribution $P$ on the joint 
covariate-outcome random vector $(X,Y(1),\cdots,Y(M))\in\X\times
\prod_{a=1}^M\mathcal{Y}_a$. 
Consider a data set $\D=\{(X_i,A_i,Y_i)\}_{i\in[n]}$ consisting of $n$ 
i.i.d.~draws of $(X,A,Y)$,
where $X_i\in\mathcal{X}$ is the observed contextual vector, $A_i\in[M]$ the action, 
and $Y_i = Y(A_i)$ the realized reward. 
The actions are selected by the {\em behavior policy} $\po$, where
$\po(a\given x):=\PP(A_i = a \given X= x)$ is the {\em propensity score}, for any $a\in [M],x\in\X$.
We make the following assumptions for $\po$ and $P$. 

\begin{assumption}\label{assum:pi0Ydistri}
The behavior policy $\po$ and the joint distribution $P$
satisfy the following.
(1) {\em Unconfoundedness:} $(Y(1),\cdots,Y(M))\ind A \given X$. (2) {\em Overlap:} for some $\ve>0$, $\po(a\given x)\geq\ve$, for all $(a,x)\in [M]\times\mathcal{X}$. (3) {\em Bounded reward support:} there exists $\bar{y}>0$, such that $0\leq Y(a)\leq \bar{y}$ for all $a \in [M]$.
\end{assumption}

The above assumptions are standard in the 
literature~\citep[see e.g.,][]{athey2021policy,zhou2023offline,si2023distributionally, kallus2022doubly}. 
In particular, the unconfoundedness assumption guarantees identifiability, and 
the overlap assumption ensures sufficient exploration when collecting the training dataset.
The bounded reward support is assumed for the ease of exposition, and 
can be relaxed to the sub-Gaussian reward straightforwardly.

\subsection{The KL-distributionally Robust Formulation}
Given the training set $\D = \{(X_i,A_i,Y_i)\}_{i \in [n]}$ 
and a policy class $\Pi$, we aim to learn a policy $\pi \in \Pi$  
that achieves high expected reward in a 
target environment that may deviate from the data-collection environment where $\D$ is collected. 
While distribution shift can take place in various forms, 
we focus primarily on the concept drift, 
where only the conditional reward distribution $Y(a)\given X$ differs 
in the training and target environments.
The distance between distributions is quantified by the 
KL divergence. 
\begin{definition}[KL divergence]
The KL divergence between two 
distributions $Q$ and $P$ is defined as 
$D_{\kl}(Q \,\|\, P) = \EE_{Q}[\log\frac{dQ}{dP}]$, 
where $\frac{dQ}{dP}$ is the Radon-Nikodym derivative of $Q$ with respect to $P$.
\end{definition}
We define an uncertainty set of neighboring distributions 
around $P$, whose conditional outcome distribution is bounded in 
KL divergence from $P$. Given a radius $\delta >0$, 
the uncertainty set of the conditional distribution is defined as $\mathcal{P}(P_{Y \given X},\dd):= \big\{Q_{Y\given X}:
D_{\text{KL}}(Q_{Y \given X} \,\|\, P_{Y \given X})\leq\dd \big\}$,
where $P_{Y \given X}$ and $Q_{Y \given X}$ refers to the 
distribution of $(Y(1),\dots,Y(M)) \given X$ under $P$ and 
$Q$ respectively.
The distributionally robust policy value for any policy $\pi$ 
at level $\delta$ is defined as 
\begin{align}\label{eqn:objectiveYshift}
\Vd(\pi):=\EE_{P_X}
\bigg[\inf_{Q_{Y\given X}\in\mathcal{P}(P_{Y\given X}, \dd)}
\EE_{Q_{Y\given X}}\Big[Y\big(\pi(X)\big)\Biggiven X\Big]\bigg].
\end{align}
The optimal policy in $\Pi$ is the one that maximizes $\Vd(\pi)$, i.e. $\pi^*_{\dd}:=\arg\!\max_{\pi\in\Pi}~\Vd(\pi)$.\footnote{When the 
supremum cannot be attained, we can always construct a sequence of 
policies whose policy values converge to the supremum, and 
all the arguments go through with a limiting argument.}

Under this formulation, our goal is to 
learn a ``robust'' policy with a high value of $\Vd(\pi)$ 
using a dataset drawn from~$P$.
The task here is two-fold: we need to (i) estimate the policy value $\Vd(\pi)$ 
for a given policy $\pi$, and (ii) find a near-optimal robust policy $\hat{\pi}\in\Pi$ 
whose policy value is close to the optimal policy $\pi^*_{\dd}$.
Here, the performance of a learned policy $\hat{\pi}$
is measured by the sub-optimality gap (regret):
\#\label{eq:regret}
\Regret(\hat{\pi}):=\Vd(\pi_\dd^*)-\Vd(\hat{\pi}).
\#
In the following sections, we tackle each task sequentially.

\subsection{Strong Duality}
In order to estimate $\V_\delta(\pi)$, 
we first rewrite 
the inner optimization problem in Equation~\eqref{eqn:objectiveYshift} 
in its dual form using standard results in convex optimization (see e.g.,~\citet{luenberger1997optimization}).
The transformation is formalized in the following lemma, with its proof 
provided in Appendix~\ref{appx:proof_duality}.

\begin{lemma}[Strong Duality]\label{lemma:strong_duality}
Given any $\pi \in \Pi$ and any $x\in \mathcal{X}$, 
the optimal value of inner optimization problem in Equation~\eqref{eqn:objectiveYshift} equals to
\begin{align}\label{eq:dual}
-\min_{\al\geq0,\eta\in\mathbb{R}}\EE_{P}
\bigg[V_\dd(\al,\eta;\pi)\Biggiven X=x\bigg].
\end{align}
where $V_\dd(\al,\eta;\pi)=\al \exp(-\frac{Y(\pi(X))+\eta}{\al}-1)+\eta+\al\dd$.
\end{lemma}
We note that the optimization problem in~\eqref{eq:dual} 
depends on $x$ and $\pi$ --- to manifest this dependence, 
we use $(\bmalpha^*_{\pi}(x), \bmeta^*_{\pi}(x))$ to denote its 
optimizer, i.e., $\bmalpha^*_\pi$ and $\bmeta^*_\pi$ are functions of $x$ and 
\begin{align*}
\big(\bmalpha^*_\pi(x),\bmeta^*_\pi(x)\big)\in\argmin{\al\geq0,\eta\in\mathbb{R}}~\EE_{P}
\bigg[V_\dd(\al,\eta;\pi)\Biggiven X=x\bigg].
\end{align*}
 With this notation and Lemma~\ref{lemma:strong_duality}, 
the robust policy value
\begin{align}\label{eqn:V_dual}
&\V_{\delta}(\pi)= -\EE_{P}\bigg[V_\dd(\bmalpha^*_\pi(X),\bmeta^*_\pi(X);\pi)\bigg].
\end{align}
The above formulation has thus translated the original 
distributionally robust optimization problem into 
an {\em empirical risk minimization (ERM)} problem.
We note that, unlike the well-studied joint distributional 
shift formulation, the above representation admits an optimizer
pair $(\bmalpha^*_\pi(x),\bmeta^*_\pi(x))$ that 
is {\em dependent} on the context $x$ 
(i.e. $\bmalpha^*_\pi,\bmeta^*_\pi$ are functions of $x$) and the policy $\pi$. 
As we shall see shortly, our proposed policy value estimation procedure employs ERM tools 
to estimate $(\bmalpha^*_\pi,\bmeta^*_\pi)$, and then compute an estimate of $\V_\delta(\pi)$ by plugging 
$(\bmalpha^*_\pi,\bmeta^*_\pi)$ into Equation~\eqref{eqn:V_dual}. 

The remaining challenge in this proposal   
is the slow estimation rate of the optimizers --- if we na\"ively plug in 
the optimizers, the resulting policy value estimator typically has 
a convergence rate slower than root-$n$. 
To overcome this, we incorporate a novel adjustment method to debias the estimator,
which allows us to obtain a doubly-robust estimator that 
achieves root-$n$ convergence rate even when then 
nuisance parameters (e.g., $(\bmalpha^*_\pi,\bmeta^*_\pi)$) 
are converging slower than the root-$n$ rate.

We end this section by providing a sufficient condition to ensure  $\bmalpha^*_\pi(x) > 0$, which we make throughout the paper.  
\begin{assumption}\label{assum:essinf}
For $a \in [M]$ and $x\in \X$, define $\underline{y}(x;a)=\sup\{t:\PP(Y(a)<t\mid X=x,A=a)=0\}$ 
and $\tilde{p}(x;a)=\PP(Y(a)=\underline{y}(x;a)\mid X=x,A = a)$. 
It holds that $\log(1 / \tilde{p}(x;a))>\dd$ 
for $P_{X\mid A=a}$-almost all $x$.  
\end{assumption}
The above assumption is mild and can be satisfied by 
many commonly used distributions, e.g., all the continuous distributions; it requires that $P_{Y \given X,A}$ does not posit a 
large point mass at its essential infimum.
The following result from~\citet[Proposition~4]{jin2022sensitivity}, 
shows that $\als > 0$ when Assumption~\ref{assum:essinf} holds, 
ensuring that the gradient of the risk function in  
ERM has a zero mean.
\begin{proposition}[\citet{jin2022sensitivity}]\label{prop:underline-al>0} 
Under Assumption~\ref{assum:essinf}, 
the optimizer $\als$ of~\eqref{eq:dual} satisfies $\als > 0$.
\end{proposition}

\section{Policy Value Estimation under Concept Drift}\label{sec:policy-est}
\subsection{The Estimation Procedure}\label{sec:est-procedure}
Fixing a policy $\pi$, we aim to estimate the policy value $\Vd(\pi)$ 
using the training dataset $\D$. We first split~$\D$ into 
$K$ equally sized disjoint folds: $\Dk$ for $k\in[K]$,\footnote{
We assume without 
loss of generality that $n$ is divisible by $K$. In 
practice, we only need a minimum of $K=3$ folds.} where we slightly
abuse the notation and use $\cD^{(k)}$ to denote the data points or the
corresponding indices interchangeably.

For each $k \in [K]$, we first use data points in $\D^{(k+1)}$ 
to obtain the propensity score estimator $\hpik$ and 
the optimizers $(\hat \bmalpha_\pi^{(k)},\hat \bmeta_\pi^{(k)})$.\footnote{We use the convention 
that $\D^{(k)} = \D^{(k\textnormal{ mod } K)}$ for any $k$.} 
We then define
\begin{align*}
\hGpik(x,y):=&\hat{\bmalpha}_\pi^{(k)}(x) \cdot e^{-\frac{y+\hat{\bmeta}_\pi^{(k)}(x)}{\hat{\bmalpha}_\pi^{(k)}(x)}-1}
+\hat{\bmeta}_\pi^{(k)}(x)+\hat{\bmalpha}_\pi^{(k)}(x) \cdot \delta,\\
\bar{g}^{(k)}_{\pi}(x):=&\EE_{P}\Big[\hGpik\big(X,Y(\pi(X)) \big)\biggiven X=x \Big].
\end{align*}
We next use $\cD^{(k+2)}$ to obtain 
$\hgpik$ as an estimator of $\bar g_{\pi}^{(k)}$. 
The policy value estimator $\hVdk(\pi)$ for the $k$-th fold is
\begin{align}\label{eq:hatv_k}
\hVdk(\pi)=&\frac{1}{|\Dk|}\sum_{i\in\Dk}\frac{\mathbbm{1}\{\pi(X_i)=A_i\}}
{\hpik(A_i\given X_i)}\nonumber\\
&\cdot \big(\hGpik(X_i,Y_i)-\hgpik(X_i)\big) + \hgpik(X_i).
\end{align}
The policy value estimator is $\hat{\V}_\delta(\pi) := -\frac{1}{K} \sum^K_{k=1} \hat \V_\delta^{(k)}(\pi)$.
The complete procedure is summarized in Algorithm~\ref{alg:policy_est_Y_shift}.
\begin{algorithm}[tb]
    \caption{Policy estimation under concept drift}
    \label{alg:policy_est_Y_shift}
    \begin{algorithmic}
    \STATE {\bfseries Input:} Dataset $\D$; policy $\pi$; 
    uncertainty set parameter $\dd$; 
    propensity score estimation algorithm $\mathcal{C}$;
    ERM algorithm $\mathcal{E}$ for obtaining $(\bmalpha^*_\pi,\bmeta^*_\pi)$;
    regression algorithm $\mathcal{R}$ for estimating $\bar g_{\pi}$.
    \STATE Randomly split $\D$ into $K$ non-overlapping equally-sized folds $\D^{(k)}$, $k\in[K]$;
    \FOR{$k=1,\cdots, K$}
    \STATE 
    $\hpik\leftarrow\mathcal{C}(\D^{(k+1)})$, 
    $(\hat \bmalpha^{(k)}_\pi,\hat \bmeta_\pi^{(k)})\leftarrow\mathcal{E}(\D^{(k+1)})$;
    \STATE 
    $\hgpik \leftarrow\mathcal{R}
    \big(\{X_i,A_i,\hGpik(X_i,Y_i);i\in\D^{(k+2)} \}\big)$;
    \STATE 
    Compute $\hVdk(\pi)$ according to Equation~\eqref{eq:hatv_k};
\ENDFOR
\STATE \textbf{Return:} $\hVd(\pi)\leftarrow-\frac{1}{K}\sum_{k=1}^{K}\hVdk(\pi)$.
\end{algorithmic}
\end{algorithm}

\begin{remark}
The estimation procedure involves three model-fitting steps corresponding to 
$\pi_0$, $(\bmalpha^*_\pi, \bmeta^*_\pi)$, and $\bar g_\pi$, respectively.
The propensity score function $\pi_0$ can be estimated with 
off-the-shelf algorithms (e.g., logistic regression, random forest); 
the conditional mean 
$\bar g_{\pi}^{(k)}$ can be obtained by regressing $\hat{G}_{\pi}^{(k)}(X_i,Y_i)$
onto $X_i$ for the points such that $A_i = \pi(X_i)$
with standard regression algorithms, e.g.,  
kernel regression~\citep{nadaraya1964estimating,watson1964smooth}, 
local polynomial regression~\citep{cleveland1979robust,cleveland1988locally}, 
smoothing spline~\citep{green1993nonparametric}, 
regression trees~\citep{loh2011classification} and random forests~\citep{ho1995proceedings}. 
The ERM step is more complex, and will be discussed in details shortly.
\end{remark}
\begin{remark}
    The construction of the estimator $\hVd(\pi)$ employs two major techniques: cross-fitting and de-biasing. 
    The cross-fitting technique splits the training dataset $\D$ into $K$ folds equally and fits the nuisance parameters on the off-fold data. This crucially provides the convenient property of independence between the fitted nuisance parameters and policy value estimators. In contrast to the naïve plug-in estimator whose convergence rate is largely affected by the slow estimation rate of the nuisance parameters, the de-biasing technique addresses this limitation, thereby leading to the doubly-robust property of the proposed estimator.
\end{remark}

\textbf{The ERM Step.}\label{sec:ERMstep}
For notational simplicity, we denote $\theta = (\alpha,\eta)$ and write the loss function from Lemma~\ref{lemma:strong_duality} as 
\begin{equation}\label{eqn:loss-function}
\ell(x,y;\theta)=\al\exp\Big(-\frac{y+\eta}{\al}-1\Big)+\eta+\al\dd.
\end{equation}
By the notation, $\bmtheta^*_\pi(x) = (\bmalpha^*_\pi(x),\bmeta^*_\pi(x))$ 
is the optimizer of $\EE_P[\ell(x,Y(\pi(x)); \theta) \given X = x]$ with respect to $\theta$.
Throughout, we make the following assumption on $\theta^*_\pi$.
\begin{assumption}\label{assum:optimum}
For any policy $\pi$, $\exists \underline{\alpha}, \bar{\alpha}, \bar{\eta}$ 
such that $0 < \underline{\alpha} \le \bmalpha_\pi^*(x) \le \bar{\alpha}$, $\big|\bmeta_\pi^*(x)\big| \le \bar{\eta}$, $\forall x\in\X$.
\end{assumption}
The above assumption is mild and can be achieved, for example, when 
$\bmtheta^*_\pi(x)$ is continuous in $x$ and when $\X$ is compact.
We refer the readers to~\cite{jin2022sensitivity} for a more detailed discussion.

Under the unconfoundedness assumption, 
$\bmtheta^*_\pi$ is also a minimizer of 
$\EE_{P}\big[\ell(X,Y; \bmtheta(X))\indc\{A=\pi(X)\}\big]$. 
We can thus estimate $\bmtheta^*_\pi$ by minimizing the empirical risk:
\#\label{eq:erm_sol}
\hat \bmtheta_{\pi}^{(k)}\in 
\argmin{\bmtheta \in \Theta} ~\Big\{\hat\EE_{\D^{(k+1)}}[\ell(X,Y; \bmtheta(X))\indc\{A=\pi(X)\}]\Big\},
\#
where $\Theta \subseteq \{(\bmalpha, \bmeta) \mid 
\bmalpha: \X \mapsto \mathbb{R}_{\ge 0},~ 
\bmeta: \X \mapsto \mathbb{R}\}$
is to be determined.
In our implementation, we 
follow~\citet{yadlowsky2022bounds,jin2022sensitivity,sahoo2022learning},
and adopt the method of sieves~\citep{geman1982nonparametric} to solve~\eqref{eq:erm_sol}.
Specifically, we consider an increasing sequence $\Theta_1\subset\Theta_2\subset\cdots$ 
of spaces of smooth functions, and let $\Theta = \Theta_n$ in Equation~\eqref{eq:erm_sol}. 
For example, $\Theta_n$ can be a class of polynomials, splines, or wavelets.
It has been shown in~\citet[Section 3.4]{jin2022sensitivity} that under mild 
regularity conditions, $\hat \bmtheta_\pi^{(k)}$ converges to $\bmtheta^*_\pi$ 
at a reasonably fast rate. For example, if $\X = \prod^d_{j=1}\X_j \subseteq \mathbb{R}^d$ 
for some compact intervals $\X_j$ and that $\bmtheta^*_\pi$ belongs to the H\"older class of 
$p$-smooth functions --- with some other mild regularity conditions ---
then $\|\hat \bmtheta_\pi^{(k)} - \bmtheta_\pi^*\|_{L_2(P_{X\given A = \pi(X)})} 
= O_P((\frac{\log n}{n})^{-p/(2p+d)})$ and 
$\|\hat \bmtheta_\pi^{(k)} - \bmtheta_\pi^*\|_{L_\infty} = O_P((\frac{\log n}{n})^{-2p^2/(2p+d)^2})$. The solution details are given in Appendix~\ref{app:erm-solution}, and we also refer the readers to~\citet{yadlowsky2018bounds} and~\citet{jin2022sensitivity}.

\subsection{Theoretical Guarantees}
We are now ready to present the theoretical guarantees for the policy value estimator 
$\hVd(\pi)$.
To start, we assume the following for the convergence rates of the 
nuisance parameter estimators.

\begin{assumption}[Asymptotic estimation rate]\label{assum:convrate}
For any policy $\pi$, assume that for each $k\in[K]$, the estimators $\hat \pi_0^{(k)}$, $\hat g_\pi^{(k)}$ and the empirical risk optimizer $\hat \bmtheta_\pi^{(k)}$ satisfy
\begin{align*}
\|\hpik-\po\|_{L_2(P_{X\given A = \pi(X)})} &= o_P(n^{-\gamma_1}),\\
\|\hgpik- \bar{g}_{\pi}^{(k)}\|_{L_2(P_{X\given A = \pi(X)})}&=o_P(n^{-\gamma_2}),\\
\|\hat{\bmtheta}_\pi^{(k)}-\bmtheta^*_\pi\|_{L_2(P_{X \given A= \pi(X)})}&=o_P(n^{-\frac{1}{4}}),\\
\|\hat \bmtheta_\pi^{(k)}-\bmtheta^*_\pi\|_{L_\infty}&=o_P(1),
\end{align*}
for some $\gamma_1,\gamma_2 \ge 0$ and $\gamma_1 + \gamma_2 \ge \frac{1}{2}$.
\end{assumption}
Assumption~\ref{assum:convrate} (1) requires either the propensity score $\pi_0$ 
or the conditional mean of $\hat G_\pi^{(k)}(X,Y)$ is well estimated. This is 
a standard assumption in 
the double machine learning literature~\citep{chernozhukov2018double,athey2021policy,zhou2023offline,
kallus2019localized,kallus2022doubly,jin2022sensitivity} 
and can be achieved by various commonly-used machine learning methods discussed in Section~\ref{sec:est-procedure}. 
Assumption~\ref{assum:convrate} (2) requires the optimizer $\hat \bmtheta_\pi^{(k)}$ to be estimated at a 
rate faster than $n^{-1/4}$, and can be achieved by, for example, the estimators discussed in Section~\ref{sec:est-procedure}
under mild conditions. The empirical sensitivity analysis in \citet{jin2022sensitivity} also provides some 
justification for Assumption~\ref{assum:convrate}.

The following theorem states that our estimated policy value $\hVd(\pi)$ 
is consistent for estimating $\V_\delta$ and is asymptotically normal. 
Its proof is provided in Appendix~\ref{appx:proof_asympconv}. 

\begin{theorem}[Asymptotic normality]\label{thm:asymconv}
Suppose Assumptions~\ref{assum:pi0Ydistri},~\ref{assum:essinf},~\ref{assum:optimum}, and~\ref{assum:convrate} hold. 
For any policy $\pi: \X \mapsto [M]$, we have $\sqrt{n}\cdot \big(\hVd(\pi)-\Vd(\pi)\big) \stackrel{\textnormal{d}}{\rightarrow} N(0,\sigma^2_{\pi})$,
where
\begin{align*}
& \sigma^2_\pi = \textnormal{Var}\bigg(\frac{\indc\{A = \pi(X)\}}{\pi_0(A \given X)}
\cdot \big(G(X,Y) - g(X)\big) + g(X)\bigg);\\
& G_\pi(x,y)=\ell(x,y;\thes_\pi);\\
&g_\pi(x):=\mathbb{E}\big[G_\pi(X,Y(\pi(X)))\given X=x\big].
\end{align*}
\end{theorem}


\section{Policy Learning under Concept Drift}
Building on the results and methodology in Section~\ref{sec:policy-est}, we turn to 
the problem of policy learning under concept drift. 
Given a policy class $\Pi$ and 
an estimated policy value $\hVd(\pi)$ for each $\pi\in \Pi$,
it is natural to consider optimizing the estimated policy value over $\Pi$ to find the best policy. 
The biggest challenge here is that the nuisance parameter $\hat\bmtheta_\pi^{(k)}$ in defining $\hVd(\pi)$
is not only a function of $x\in\mathcal{X}$, but also a function of $\pi\in\Pi$. 
The above strategy requires carrying out the ERM step in Section~\ref{sec:est-procedure}, for 
all possible policies $\pi\in\Pi$, posing major computational difficulties.

Instead of solving $\hat \bmtheta^{(k)}_\pi$ for each $\pi\in\Pi$, we propose a computational 
shortcut that solves a similar ERM problem for each action $a\in[M]$. 
To see why this is sufficient, note that for any $\pi \in \Pi$,
\# \label{eq:decomp_obj}
&\EE\big[\ell(X,Y(\pi(X)); \theta) \given X = x\big]\nonumber\\
=& 
\sum^M_{a=1}\indc\{\pi(X) = a\} \cdot \EE[\ell(x, Y(a); \theta) \given X = x]. 
\#
Letting $\bmtheta^*_a(x)\in\argmin{\theta} \, \big\{\EE[\ell(x,Y(a);\theta)\given X=x]\big\}$, 
we can see that $\bmtheta^*_{\pi(x)}(x)$ is a minimizer of~\eqref{eq:decomp_obj}.
Then, the policy learning problem reduces to finding $\pi \in \Pi$ 
that maximizes
\begin{align*}
-\EE\bigg[\ell(X,Y(\pi(X));\bmtheta^*_{\pi(X)}(X))\bigg].
\end{align*}
\begin{remark}[Computational efficiency of the proposed shortcut]
Constructing $\hat\bmtheta_\pi(x)$ with $\hat\bmtheta_{\pi(x)}(x)$ substantially reduces the computational complexity of the policy learning task. It is virtually infeasible to estimate $\hat\bmtheta_\pi(x)$ for each $\pi$ in a policy class $\Pi$ with infinite number of policies. Alternatively, solving for $\hat\bmtheta_{\pi(x)}(x)$ transforms the infeasible task of computing a class of infinite nuisance parameters $\{\bmtheta_\pi(x):\pi\in\Pi\}$ to the feasible task of computing a finite one $\{\bm\theta_{a}(x):a\in[M]\}$. It remains an interesting future direction to extend this trick to continuous action spaces.
\end{remark}
\subsection{The Learning Algorithm}
The policy learning algorithm consists of two main steps: (1) solving for $\bmtheta^*_a$ for each $a \in [M]$ 
and constructing the policy value estimator $\hVd(\pi)$; (2) learning the optimal policy $\pi^*_\dd$ by minimizing  
$\hVd(\pi)$.

As before, we randomly split the original data set $\mathcal{D}$ into $K$ folds. 
For each fold $k \in [K]$, we use samples in the $(k+1)$-th data fold $\mathcal{D}^{(k+1)}$ 
to obtain the propensity estimator $\hpik(a\given\cdot)$ (by regression) 
and the optimizer $\bmtheta^*_a$ (by ERM) for each $a\in[M]$. 
Next, for each $a\in[M]$, define 
\begin{align*}
G_a(x,y):= \ell(x,y;\bmtheta^*_a(x)),~
\hat G^{(k)}_a(x,y) := \ell(x,y;\hat \bmtheta^{(k)}_a(x)),\\
\bar{g}^{(k)}_a(x) := \EE\big[\hat G^{(k)}_a(X,Y(a))\given X=x\big].
\end{align*}
We then obtain an estimator $\hat g^{(k)}_a$ for $\bar g^{(k)}_a$  
by regressing $\hat G^{(k)}_a(X_i,Y_i)$ onto $X_i$ with $i \in \cD^{(k+2)}$. 
Finally, we obtain the learned policy by maximizing the estimated policy value: 
\begin{align}\label{eqn:policyvaluelearn}
&\hat \pi_{\text{LN}} =\argmax{\pi\in\Pi}~ 
\hat \V^{\LN}_\delta(\pi) :=
-\frac{1}{K} \sum^K_{k=1} \hat{\mathcal{V}}_\delta^{\text{LN},(k)}(\pi);\\
&\hVd^{\text{LN},(k)}(\pi)=\frac{1}{|\Dk|}\sum_{i\in\Dk}
\frac{\mathbbm{1}\{A_i = \pi(X_i)\}}{\hpik(A_i\given X_i)}\nonumber\\
&\cdot 
\big(\hat G^{(k)}_{\pi(X_i)}(X_i,Y_i)-\hat g^{(k)}_{\pi(X_i)}(X_i)\big)+\hat g^{(k)}_{\pi(X_i)}(X_i).\nonumber
\end{align}
The above optimization problem can be solved efficiently by first-order optimization methods 
or policy tree search as in~\citet{zhou2023offline}; we shall elaborate on the implementation 
in Section~\ref{sec:experiment}. The complete policy learning procedure is summarized in Algorithm~\ref{alg:policylearning}.
\begin{algorithm}[tb]
    \caption{Policy learning under concept drift}
    \label{alg:policylearning}
\begin{algorithmic}
    \STATE {\bfseries Input:} Dataset $\mathcal{D}$; policy class $\Pi$; uncertainty set parameter $\dd$; propensity score estimation algorithm $\mathcal{C}$; ERM algorithm $\mathcal{E}(\cdot)$ for obtaining $\bmtheta^*_a$; regression algorithm $\mathcal{R}$ for estimating $\bar{g}_a$.
    \STATE Randomly split $\mathcal{D}$ into $K$ equal-sized folds;
    \FOR{$k=1,\ldots,K$}
        \STATE $\hpik\leftarrow\mathcal{C}(\D^{(k+1)})$, 
        \FOR{$a=1,\cdots,M$}
            \STATE $\hat{\bmtheta}^{(k)}_{a}\leftarrow\mathcal{E}(\D^{(k+1)})$;
            \STATE $\hat{g}_{a}^{(k)}\leftarrow\mathcal{R}(X_i,A_i,\hat{G}^{(k)}_a(X_i,Y_i);i\in\D^{(k+2)})$;
        \ENDFOR
    \ENDFOR
    \STATE {\bfseries Return:} $\hpi$ that maximizes $\hVdl(\pi)$ as in Equation~\eqref{eqn:policyvaluelearn}.
\end{algorithmic}
\end{algorithm}

\subsection{Regret Upper Bound}
In this section, we present the regret analysis of $\hpi$ obtained 
by Algorithm~\ref{alg:policylearning} (recall the definition of 
regret given in Equation~\eqref{eq:regret}). 
Before we embark on the formal analysis, 
we introduce the Hamming entropy integral $\kappa(\Pi)$, which measures the complexity of $\Pi$.
\begin{definition}[Hamming entropy integral]
Given a policy class $\Pi$ and $n$ data points $\{x_1,\ldots,x_n\}\subseteq \mathcal{X}$, 
\begin{itemize}
\itemsep0em 
\item [(1)] The \emph{Hamming distance} between two policies 
$\pi,\pi'\in\Pi$ is $d_H(\pi,\pi'):=\frac{1}{n}\sum_{i=1}^n\mathbbm{1}\{\pi(x_i)\neq\pi'(x_i)\}$.
\item [(2)] The $\ve$-\emph{covering number} of $\{x_1,\ldots,x_n\}$, 
denoted as $\mathcal{C}(\epsilon,\Pi;\{x_1,\ldots,x_n\})$, is 
the smallest number $L$ of policies $\{\pi_1,\ldots,\pi_L\}$ in $\Pi$, 
such that $\forall$ $\pi\in\Pi$, $\exists$ $\pi_\ell'$ 
such that $d_H(\pi,\pi_\ell)\leq\epsilon$. 
\item [(3)] The \emph{Hamming entropy integral} of $\Pi$ is defined as 
$\kappa(\Pi):=\int_0^1\sqrt{\log N_H(\epsilon^2,\Pi)}\, d\epsilon$, where $N_H(\epsilon,\Pi):=\sup_{n\geq1}\sup_{x_1,\ldots,x_n}
\mathcal{C}(\epsilon,\Pi;\{x_1,\ldots,x_n\})$.
\end{itemize} 
\end{definition}

The following theorem provides a regret upper bound for the policy learned by Algorithm~\ref{alg:policylearning}.
\begin{theorem}\label{thm:regret-bound}
Suppose Assumptions~\ref{assum:pi0Ydistri},~\ref{assum:essinf},~\ref{assum:optimum},~\ref{assum:convrate} hold. 
For any $\beta\in(0,1)$, there exists $N \in \mathbb{N}_+$ such that when $n\ge N$, 
we have with probability at least $1-\beta$ that 
\begin{align*}
\Regret(\hat \pi_{\LN})
\le 
\frac{C_0(\bar{\alpha}, \underline{\alpha}, \bar{\eta}, \delta,\ve)}{\sqrt{n}}
\big(65 + 8\kappa(\Pi) + \sqrt{\log(1/\beta)}\big),
\end{align*}
where $C_0(\bar{\alpha}, \underline{\alpha}, \bar{\eta}, \delta, \ve) :=
6 (\bar \alpha \cdot \exp({\bar{\eta}}/{\underline{\alpha}} -1) 
+ \bar{\eta} + \bar{\alpha}\delta)/\ve$.
\end{theorem}

The proof of Theorem~\ref{thm:regret-bound} is deferred to Appendix~\ref{sec:proofs}. 
At a high level, we decompose the regret and upper bound it with the supremum of 
the estimation error of policy values: 
\begin{align*}
\Regret(\hpi) =& \Vd(\pi^{\ast})-\Vd(\hpi)\leq \Vd(\pi^*)-\hVdl(\pi^*)\\
&+\hVdl(\pi^*)-\hVdl(\hpi)+\hVdl(\hpi)-\Vd(\hpi)\\
\le& 2\sup_{\pi \in \Pi} |\hat \V^{\LN}_\delta(\pi) - \V_\delta(\pi)|,
\end{align*}
where the last step uses the choice of $\hat \pi_\LN$.
We bound the above quantity by establishing uniform convergence 
results for the policy value estimators. 
Through a careful chaining argument, we show that the dependence of $\mathcal{R}(\hpi)$ on 
$n$ is of the order $O(n^{-\frac{1}{2}})$, which is sharper than  
the $O(n^{-\frac{1}{2}}\log n)$ dependence for that of~\citet{mu2022factored}
by a logarithmic factor. 
We also note that both regrets are asymptotic in $n$ and hold for sufficiently large $n$.

\subsection{Regret Lower Bound}
In this section, we complement the regret upper bound in Theorem~\ref{thm:regret-bound}
with a minimax lower bound that characterizes the fundamental difficulty of policy 
learning under concept drift. Our lower bound is stated in terms of the Natarajan 
dimension~\citep{natarajan1989learning}, defined as follows.
\begin{definition}[Natarajan dimension]\label{def:nat-dim}
Given an $M$-action policy class $\Pi$, we say a set of $m$ points 
$\{x_1,\ldots,x_m\}$ is \emph{shattered} by $\Pi$ if there exist two functions 
$f_{-1},f_1: \{x_1,\ldots,x_m\} \mapsto [M]$ such that $f_{-1}(x_j) \neq f_1(x_j)$ for all $j \in [m]$ and for any $\sigma \in \{\pm 1\}^m$, there exists a policy $\pi \in \Pi$
such that for any $j \in [m]$, $\pi(x_j) = f_{\sigma_j}(x_j)$. The \emph{Natarajan dimension} Ndim($\Pi$) of $\Pi$ is the size of the largest 
set shattered by $\Pi$.
\end{definition}
\begin{remark}[Connection to other complexity measures]
As can be seen from the definition, the Natarajan dimension generalizes the 
Vapnik-Chervonenkis (VC) dimension~\citep{vapnik2015uniform} 
to the multi-class classification setting.
The Natarajan dimension is also closely related to the Hamming 
entropy integral $\kappa(\Pi)$ in our upper bound, 
as $\kappa(\Pi) = O(\sqrt{\log(d)\text{Ndim}(\Pi)}$)~\citep{qu2020interpretable}.
\end{remark}

\begin{theorem}[Regret lower bound]\label{thm:lower_bnd}
Let $\mathcal{P}$ denote the set of all distributions 
of $(X,A,Y(1),\ldots,Y(M))$ that satisfy Assumption~\ref{assum:pi0Ydistri}, \ref{assum:essinf}, 
\ref{assum:optimum}, and \ref{assum:convrate}.\footnote{When we say a distribution $P$ satisfies
Assumption~\ref{assum:convrate}, we mean that under $P$ there exist $\hat \bmtheta$, 
$\hat \pi_0$, and $\hat g$ that satisfy the convergence rates in Assumption~\ref{assum:convrate}.}
Suppose that $\delta \le 0.2$, $n \ge \textnormal{Ndim}(\Pi)^2$, and $\textnormal{Ndim}(\Pi)\ge 4/(9\varepsilon)$. For any policy leaning algorithm that
 outputs $\hat \pi$ as a function of 
$\{(X_i,A_i,Y_i)\}_{i=1}^n$, there is $\sup_{P \in \mathcal{P}} \EE_{P^n}[\mathcal{R}(\hat \pi)]
\ge \frac{1}{120}\sqrt{\frac{\textnormal{Ndim}(\Pi)}{n\varepsilon}}.$
\end{theorem}
The proof of Theorem~\ref{thm:lower_bnd} is provided in Appendix~\ref{sec:proof_lower_bnd}. 
Theorem~\ref{thm:lower_bnd} implies that for any learning algorithm, there exists a 
problem instance such that the regret scales as $\Omega(\sqrt{\text{Ndim}(\Pi)/n})$. 
\begin{remark}[Optimality of Algorithm~\ref{alg:policylearning}]
Recalling the relationship between the Natarajan dimension and the Hamming entropy integral 
in the remark above, we see that our proposed algorithm achieves the minimax rate in 
the sample size and the policy class complexity.
\end{remark}

\begin{table*}[t]
\begin{center}
\begin{small}
\begin{sc}
\begin{tabular}{lcccccr}
\toprule
Metric & $\dd$ & Policy & $n=$7500 & $n=$13500 & $n=$16500 & $n=$19500 \\
\midrule
\multirow{6}{*}{$\bar{\V}_{\dd}$} & \multirow{2}{*}{0.05} & $\hpi$ & 0.2272$\pm0.002$ & 0.2299$\pm0.001$ & 0.2303$\pm0.001$ & 0.2310$\pm0.001$\\
& & $\widehat{\pi}_{\text{SNLN}}$ & 0.0554$\pm0.005$ & 0.0589$\pm0.004$ & 0.0617$\pm0.004$ & 0.0664$\pm0.003$\\
\cmidrule(l{0.5em}){2-7}
& \multirow{2}{*}{0.1}& $\hpi$ & 0.1579$\pm0.007$ & 0.1662$\pm0.002$ & 0.1663$\pm0.002$ & 0.1678$\pm0.002$ \\
& &$\widehat{\pi}_{\text{SNLN}}$ & 0.0548$\pm0.004$ & 0.0580$\pm0.004$ & 0.0583$\pm0.003$ & 0.0616$\pm0.004$\\
\cmidrule(l{0.5em}){2-7}
& \multirow{2}{*}{0.2}& $\hpi$ & 0.0781$\pm0.003$ & 0.0802$\pm0.002$ & 0.0804$\pm0.002$ & 0.0831$\pm0.002$\\
& &$\widehat{\pi}_{\text{SNLN}}$  & 0.0182$\pm0.003$ & 0.0183$\pm0.003$ & 0.0200$\pm0.003$ & 0.0219$\pm0.003$\\
\bottomrule
\end{tabular}
\end{sc}
\end{small}
\end{center}
\caption{Empirical robust policy value $\bar{\V}_{\dd}$ of policies $\hpi,\widehat{\pi}_{\text{SNLN}}$ on the simulated dataset and the corresponding, over 50 seeds.}
\label{tbl:learning-results}
\end{table*}

\begin{figure*}[t]
\begin{center}
\centerline{\includegraphics[width=0.8\textwidth]{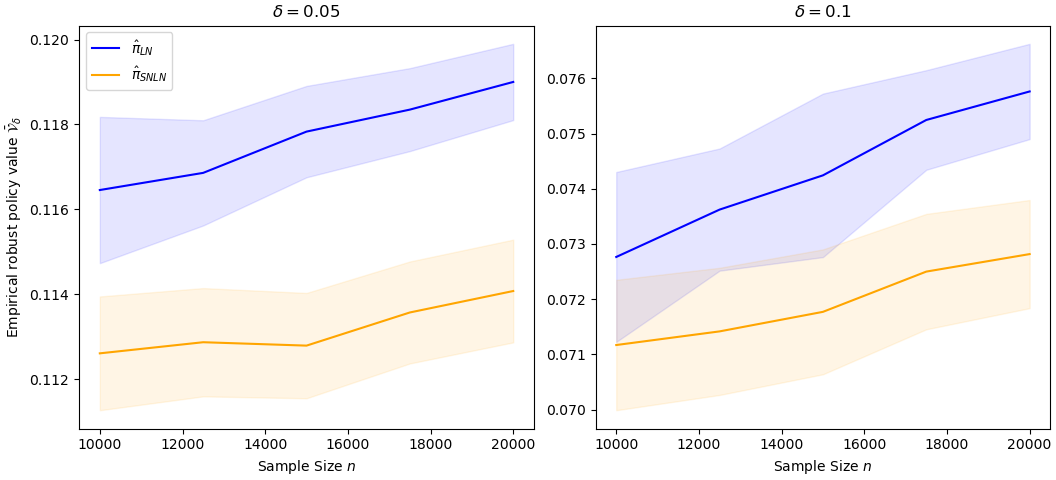}}
\caption{Empirical robust policy value $\bar{\V}_{\dd}$ of policies $\hpi,\widehat{\pi}_{\text{SNLN}}$ on the real-world dataset, over 50 seeds. 
Shading corresponds to $95\%$ 
confidence intervals.}
\label{fig:real-result}
\end{center}
\end{figure*}

\begin{table*}[t]
\begin{center}
\begin{small}
\begin{sc}
\begin{tabular}{lccccr}
\toprule
Metric & Policy  & $n=$7500 & $n=$13500 & $n=$16500 & $n=$19500 \\
\midrule
\multirow{2}{*}{$\tilde\V_{0.1}^{\min}$} & $\hpi$ & 0.2075$\pm0.015$ & 0.2139$\pm0.005$ & 0.2149$\pm0.007$ & 0.2167$\pm0.003$ \\
& $\widehat{\pi}_{\text{SNLN}}$ & 0.1884$\pm0.007$ & 0.2009$\pm0.008$ & 0.2017$\pm0.006$ & 0.2020$\pm0.004$\\
\bottomrule
\end{tabular}
\end{sc}
\end{small}
\end{center}
\caption{Empirical worst case policy reward on the KL-sphere $\tilde\V_{\delta}^{\min}$ of policies $\hpi,\widehat{\pi}_{\text{SNLN}}$, over 20 seeds.}
\label{tbl:tilde-learning-results}
\end{table*}

\section{Numerical Results}\label{sec:experiment}
We evaluated our learning algorithm in two settings: a simulated and a real-world dataset, 
against the benchmark algorithm SNLN in \citet[Algorithm 2]{si2023distributionally}.

\textbf{Simulated Dataset.}
Our data generating process follows that of the linear boundary example 
in~\citet{si2023distributionally}. We let the context set $\X$ 
to be the closed unit ball of $\mathbb{R}^5$ and let the action set to be $\{1,2,3\}$;
the rewards $Y(a)$'s are mutually independent conditioned on $X$ 
with $Y(a)\mid X$ being Gaussian, for $a \in [3]$. The training datasets $\D_{\text{train}}$ are generated with an unknown given behavior policy $\pi_0$ over 50 random seeds. We similarly generate 100 testing datasets $\D_{\text{test}}$ of size 10,000.
The details are given in Appendix~\ref{appx:exp-details}.

\textbf{Real-world Dataset.} We consider the dataset of a large-scale randomized experiment comparing assistance programs offered to French unemployed individuals provided in \citet{behaghel2014private}. The decision maker is trying to learn a personalized policy that decides whether to provide: (i) an intensive counseling program run by a public agency ($A=0$); or (ii) a similar program run by private agencies $(A=1)$, to an unemployed individual. The reward $Y$ is binary and indicates reemployment within six months. The processed dataset is provided in \citet{kallus2023treatment}.

\textbf{Implementation.}
In our implementation, the number of splits is taken to be $K=3$. 
We use the Random Forest regressor from the \texttt{scikit-learn} Python library to 
estimate $\hat{\pi}_0$ and $\hat{g}$. 
For estimating $\bmtheta^*$, we adopt the cubic spline method and 
employ the Nelder-Mead optimization method in \texttt{SciPy} 
Python library~\citep{virtanen2020scipy} to optimize the coefficients in the spline approximation,
where the obtained estimator has threshold at 0.001 to guarantee Proposition~\ref{prop:underline-al>0}. 
Finally, we optimize and find $\hat{\pi}_{\LN}$ with \texttt{policytree}~\citep{sverdrup2020policytree}.\footnote{A working example on the real-world dataset is given~in \url{https://github.com/off-policy-learning/concept-drift-robust-learning}.} 

The benchmark algorithm SNLN is adapted from~\citet[Algorithm 2]{si2023distributionally} as in \citet{kallus2022doubly}.\footnote{The implemented code of SNLN benchmark is provided by \citet{kallus2022doubly}.} 
Since \citet[Algorithm 2]{si2023distributionally} is designed for joint distribution shift formulation, we revised the original algorithm to fit our concept drift setting. The well-known KL-divergence chain rule~\citep{cover1999elements} gives
\small
\begin{align}\label{eqn:KL-chain-rule}
&D_{\text{KL}}(Q_{X,Y} \,\|\, P_{X,Y})\nonumber\\
&= D_{\text{KL}}(Q_{X} \,\|\,P_{X})+D_{\text{KL}}(Q_{Y\given X} \,\| \,P_{Y\given X}).
\end{align}
\normalsize
Therefore, given any uncertainty set radius $\dd$ and known covariate shift 
(in this experiment, we assume no covariate shift),~\citet[Algorithm 2]{si2023distributionally}
 can be used to implement policy learning under concept drift. 
 Note that SNLN admits known propensity scores. As we only consider the case 
 where the propensity scores are unknown, we complement~\citet[Algorithm 2]{si2023distributionally} 
 with estimated propensity scores from Random Forest Regressor in \texttt{scikit-learn}, 
 the same way as in the implementation of Algorithm~\ref{alg:policylearning}. 

\textbf{Evaluation.}
For a learned policy $\hat \pi$, 
we evaluate its performance by the following performance metrics. (i)
We use the testing dataset $\D_{\text{test}}$ to estimate the robust policy value $\V^*_\delta(\hat \pi)$ by the empirical robust policy value
\begin{align*}
\bar{\V}_\dd(\hat \pi)=-\frac{1}{|\D_{\text{test}}|}\sum_{i\in\D_{\text{test}}}\ell(X_i,Y_i(\pi(X_1));\bmtheta_{\hat\pi}(X_i)),
\end{align*}
where the nuisance parameters $\bmtheta_{\hat\pi}(X)$ are found via cubic spline method and employ the Nelder-Mead optimization method using the testing dataset according to policy $\hat \pi$: $\D_{\text{test},\hat \pi}=\{(X_i,Y_{i}(\hat \pi(X_i)))\}$. (ii) We also perturb the simulated dataset to mimic possible real-world distributional shift. For each testing dataset $j$ containing 10000 data points of the total 100 testing datasets, we generate a new testing dataset, with each reward distribution $(\tilde Y_i^{(j)}(1),\tilde Y_i^{(j)}(2),\tilde Y_i^{(j)}(3))$ randomly sampled on the KL-sphere centered at the reward distribution $(Y_i^{(j)}(1),Y_i^{(j)}(2),Y_i^{(j)}(3))$ of the testing data point with a radius $\delta$. Then we evaluate $\hat\pi$ using
\begin{align*}
\tilde\V_{\delta}^{\min}(\hat \pi):=\min_{j\in[100]}\bigg\{\frac{1}{10000}\sum_{i=1}^{10000}\tilde Y_i^{(j)}\big(\hat\pi(X_i^{(j)})\big)\bigg\}.
\end{align*}
This simulates a more realistic scenario where the policy performance is measured by test datasets with concept drifts.

\textbf{Results.}
Table~\ref{tbl:learning-results} and \ref{fig:real-result} reports the values $\bar\V_{\dd}$ of the learned policies $\hpi$ and $\widehat{\pi}_{\text{SNLN}}$ on the simulated and the real-wrold dataset, respectively. Table~\ref{tbl:tilde-learning-results} provides the result of  $\tilde\V_\dd^{\min}$. All results are reported with 95\% confidence intervals.
Table~\ref{tbl:learning-results} shows that $\hpi$ outperforms the benchmark $\widehat{\pi}_{\text{SNLN}}$ consistently, with higher policy values and similar 95\% confidence intervals. In Figure~\ref{fig:real-result}, $\hpi$ continues to show this advantage over $\widehat{\pi}_{\text{SNLN}}$ on the real-world dataset. With a higher $\dd$, the policy values of $\hpi,\widehat{\pi}_{\text{SNLN}}$ are smaller, due to a bigger uncertainty set. Table~\ref{tbl:tilde-learning-results} shows that $\hpi$ achieves higher worst-case rewards than $\widehat{\pi}_{\text{SNLN}}$ does, in a more realistic setting with concept drift testing datasets. Together, we see that $\hpi$ succeeds in finding a better policy under concept drift; while the performance of $\widehat{\pi}_{\text{SNLN}}$ is comprised by its conservative policy learning process, in which it considers joint distributional shifts even though it is given the information that no covariate shifts took place.

The results align with the intuition that Algorithm~\ref{alg:policylearning} admits a subset of the uncertainty set that SNLN considers, as explained in Equation~\eqref{eqn:KL-chain-rule}. Consequently, $\V_{\dd}(\widehat{\pi}_{\text{SNLN}})$ is a lower bound of $\V_{\dd}(\hpi)$ in theory, and by the results in Table~\ref{tbl:learning-results}, in practice. In real-world applications, knowing the source of the distribution shift effectively shrinks the uncertainty set, thereby yielding less conservative results. Since it is fairly easy to identify covariate shifts (comparing to detecting concept drift), when the decision maker observes none or little covariate shifts and would like to hedge against the risk of concept drift, it is suitable to apply our method which outperforms existing method designed for learning under joint distributional shifts.

One limitation of our methodology (as well as in other DRO works) is the choice of $\dd$. The parameter $\dd$ controls the size of the uncertainty set considered and thus controls the degree of robustness in our model --- the larger $\dd$, the more robust the output. The empirical performance of the algorithm substantially depends on the selection of $\dd$. A small $\dd$ leads to negligible robustification effect and the algorithm would learn an over-aggressive policy; a large $\dd$ tends to yield more conservative results. A more detailed discussion can be found in \citet{si2023distributionally}.

In Appendix~\ref{appx:exp-details}, we also provide simulation results of Algorithm~\ref{alg:policy_est_Y_shift} for a fixed target policy, which show that Algorithm~\ref{alg:policy_est_Y_shift} can estimate the distributionally robust policy value under concept drift efficiently. 

\section{Discussion}
In this paper, we study the policy learning problem under concept drift, where we propose a doubly-robust policy value estimator that is consistent 
and asymptotically normal, and then develop a minimax optimal policy learning algorithm, whose regret is $O_p(\kappa(\Pi)n^{-{1}/{2}})$ with a matching lower bound.
Numerical results show that our learning algorithm outperforms the benchmark algorithm under concept drift.
We also note that our results on pure concept drift could be extended to a more general setting where the concept drift adopts the same form as ours, but there is in addition an identifiable covariate shift as in~\citet{jin2022sensitivity}. The details can be found in Appendix~\ref{app:x-shift}.

\newpage
\section*{Acknowledgments}
This work is generously supported by the ONR grant 13983263
and the NSF grant CCF-2312205.
The authors would like to thank Miao Lu and Wenhao Yang for pointing out an error in an earlier draft of this manuscript and their helpful comments. Z.~R.~is supported by Wharton Analytics. R.~Z.~is supported by RGC grant ECS-26210324.


\section*{Impact Statement}

This paper presents work whose goal is to advance the field of 
Machine Learning. There are many potential societal consequences 
of our work, none which we feel must be specifically highlighted here.


\bibliography{mybib}
\bibliographystyle{icml2025}

\newpage
\appendix
\onecolumn

\section{Notation}
We use $[n]$ to denote the discrete set $\{1,2,\cdots,n\}$ for any $n\in\ZZ$. 
We use $\arg\!\min$ and $\arg\!\max$ to denote the minimizers and maximizers; 
if the minimizer or the maximizer cannot  be attained, we project it back to the feasible set.
We denote the usual $p$-norm as $\|\cdot\|_p$. Denote $P$ to be any probability measure defined on the probability space $(\Omega,\sigma(\Omega),P)$ and $\hat{P}$ to be the empirical distribution of $P$.
For any function $f$, 
we denote the $L_2(P)$-norm of $f$ conventionally as $\|f\|_{L_2(P)}=(\int |f(x)|^2\,dP(x))^{1/2}$
and $\|f\|_{L_\infty} = \sup_{x \in \X} |f(x)|$.
For any random variables $X,Y$, we use $X\ind Y$ to denote 
that $X$ is independent of $Y$. For a random variable/vector $X$, 
we use $\EE_{X}[\cdot]$ to indicate the expectation taken over the distribution of $X$.

\section{The ERM Solution}\label{app:erm-solution}
To solve the ERM problem, we follow~\citet{geman1982nonparametric,yadlowsky2022bounds,jin2022sensitivity}
and adopt the method of sieves:
consider an increasing sequence $\Theta_1\subset\Theta_2\subset\cdots$ 
of spaces of smooth functions, and let $(\hapik,\hepik)=\argmin{\theta\in\Theta_n}~\EE_{n}[\ell(X,Y(\pi(X));\theta)]$.
In our case, we consider the following classes of sufficiently smooth functions. For $q_1=\lceil q\rceil-1$ and $q_2=q-q_1$ (where $q$ is the smoothness parameter), define the following function class for $\etas$:
\begin{align*}
\Theta_c^q(\X)=\bigg\{&h\in C^{q_1}(\X):\sup_{\substack{x\in\X\\\sum_{l=1}^p\gamma_l<q_1}}|D^\gamma h(x)|+\sup_{\substack{x\neq x'\in\X\\\sum_{l=1}^p\beta_l=q_1}}\frac{|D^\beta h(x)-D^\beta h(x')|}{\|x-x'\|^{q_2}}\leq c\bigg\},
\end{align*}
where we denote the derivative $D^d=\sum_{l=1}^p\frac{\partial^{d_l} D}{\partial x_l}$. To ensure the non-negativeness of $\als_\pi$ in Proposition~\ref{prop:underline-al>0}, we define the truncated function class $\Theta_c^q(\X,\epsilon):=\{x\mapsto\max\{h(x),\epsilon\}:h\in\Theta_c^q(\X)\}$ for the search of $\al_\pi$. Consequently, the function class we consider is $\Theta=\Theta_c^q(\X)\times\Theta_c^q(\X,\epsilon)$.

optimizers well, we need the true optimizer $(\als_\pi,\etas_\pi)$ to be sufficiently smooth in $x$. Convexity and stability are also desirable property of the loss function for learning the optimizers. In the next assumption, we present the regularity condition on the conditional distribution $\PP_{Y\given X}$ to ensure smoothness of optimizers.

\begin{assumption}[Smooth conditional reward distribution]\label{assum:smooth-reward-distri}
The conditional reward distribution $\PP_{Y(a)\given X=x}$ is smooth in $x$, i.e. for some $h\in\X$, $\PP_{Y(a)\given X=x+th}=\PP_{Y(a)\given X=x}+t\cdot\PP_{h}$ for some measure $\PP_h$ on $\mathcal{Y}$.
\end{assumption}

Taking practical terms into consideration, Assumption~\ref{assum:smooth-reward-distri} is reasonable as we assume the conditional distributions of $Y(a)$ are close for similar covariates, for any action $a\in\A$. \citet[Appendix~B.2]{jin2022sensitivity} presents detailed discussion to justify smoothness of the optimizers under Assumption~\ref{assum:smooth-reward-distri}.

Next, we would like to discuss the regularity conditions of the loss function $\ell$ in Equation~\eqref{eqn:loss-function} and its conditional expectation $\EE[\ell(x,Y;\theta)\given X=x,A=\pi(x)]$. In particular, we require stability of the loss function and its conditional expectation so that plugging in estimators of the optimizers will not cause large errors, which is a mild condition that can be satisfied under the first-order Taylor expansion condition. Readers can refer to \cite{van2000asymptotic,jin2022sensitivity} for a more detailed discussion. Later, Lemma~\ref{lemma:regularity} summarises the regularity conditions in formal terms and indicates that in our case, with KL-divergence and the loss function defined as in Equation~\eqref{eqn:loss-function}, all the above regularity conditions are satisfied.

\section{Experiment Details}\label{appx:exp-details}
We let the context set $\X=\{x\in\mathbb{R}^{5}:\|x\|_2\leq1\}$ 
to be the closed unit ball of $\mathbb{R}^5$ and let the action set to be $[3]$;
the rewards $Y(a)$'s are mutually independent conditioned on $X$ 
with $Y(a)\mid X\sim N(\beta_a^\top X,\sigma_a^2)$, for $a \in [3]$.
We choose $\beta$'s and $\sigma$'s to be
\begin{align*}
\beta_1=(1,0,0,0,0),&&\beta_2=(-1/2,\sqrt{3}/2,0,0,0),&&\beta_3=(-1/2,-\sqrt{3}/2,0,0,0); &&\sigma=(0.2,0.5,0.8).
\end{align*}
The training dataset $\D_{\text{train}}=\{(X_i,\pi_0(X_i),Y_i(\pi_0(X_i)))\}_{i=1}^n$ is generated with a given behavior policy $\pi_0$ (unknown to policy learning algorithms), which chooses actions conditioned on context $x$ according to the following rules:
\begin{align*}
(\pi_0(1\given x),\pi_0(2\given x),\pi_0(3\given x))=\begin{cases}
(0.5,0.25,0.25),&\text{ if }\argmax{i=1,2,3}\{\beta_i^\top x\}=1,\\
(0.25,0.5,0.25),&\text{ if }\argmax{i=1,2,3}\{\beta_i^\top x\}=2,\\
(0.25,0.25,0.5),&\text{ if }\argmax{i=1,2,3}\{\beta_i^\top x\}=3.
\end{cases}
\end{align*}

We also generate 100 testing datasets, each with sample size 10,000. Each testing dataset $\D_{\text{test}}$ consists of i.i.d.~draws of data tuple $\{(X_i,Y_i(1),Y_i(2),Y_i(3))\}_{i=1}^n$, and is generated similarly to the procedure described above.

\begin{figure}[ht]
\vskip 0.2in
\begin{center}
\centerline{\includegraphics[width=\columnwidth]{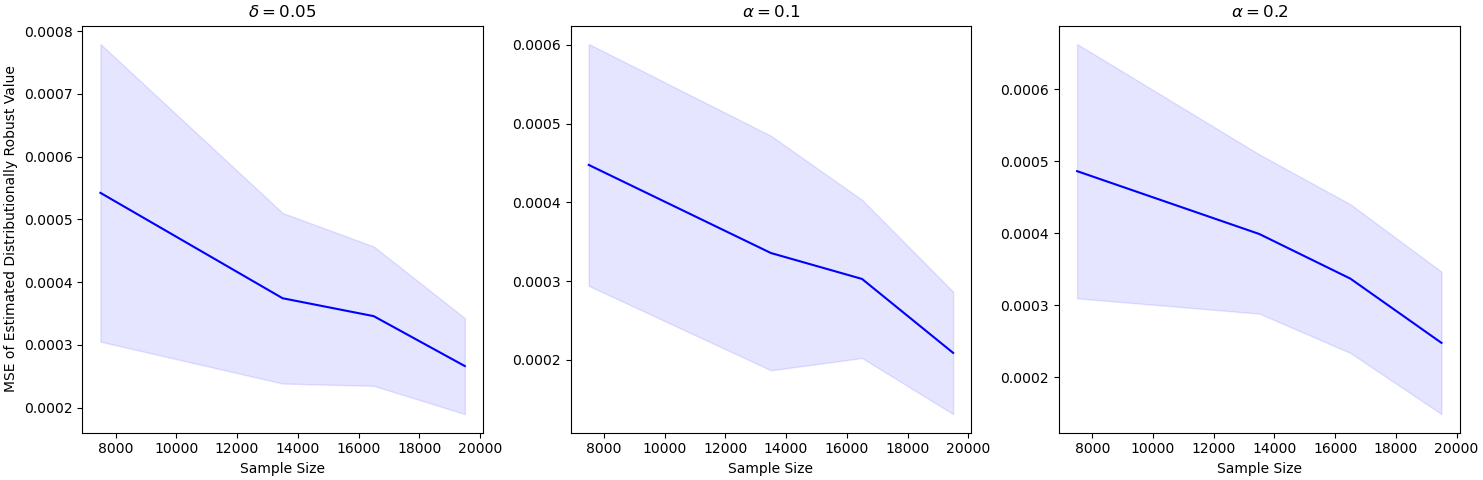}}
\caption{The Mean Square Error (MSE) of the estimated policy value by Algorithm~\ref{alg:policy_est_Y_shift}. The $x$-axis is the number of samples used by Algorithm~\ref{alg:policy_est_Y_shift}, and the $y$-axis is the mean squared error (MSE) of the policy value estimator.}
\label{fig:est_result}
\end{center}
\vskip -0.2in
\end{figure}

We present the result of the policy estimation experiments in Figure~\ref{fig:est_result}, using Algorithm~\ref{alg:policy_est_Y_shift} with inputs of the training datasets and the target policy $\pi$
\begin{align*}
\pi(x)=\begin{cases}
1,&\text{ if }\|x\|_2\in[0,1/3],\\
2,&\text{ if }\|x\|_2\in[1/3,2/3],\\
3,&\text{ if }\|x\|_2\in[2/3,1].
\end{cases}
\end{align*}
The underlying true policy value is obtained by the testing dataset $\D_{\text{test}}$. Similar to the learning experiment, we repeat the estimation experiment over 50 seeds. Figure~\ref{fig:est_result} shows that as the sample size increases, the estimated policy value by Algorithm~\ref{alg:policy_est_Y_shift} is more accurate and stable. 

\paragraph{Computation Details.}
The experiments were run on the following cloud servers: 
(i) an Intel Xeon Platinum 8160 @ 2.1 GHz with 766GB RAM and 96 CPU x 2.1 GHz; 
(ii) an Intel Xeon Platinum 8160 @ 2.1 GHz with 1.5TB RAM and 96 CPU x 2.1 GHz; 
(iii) an Intel Xeon Gold 6132 @ 2.59 GHz with 768GB RAM and 56 CPU x 2.59 GHz and 
(iv) an Intel Xeon GPU E5-2697A v4 @ 2.59 GHz with 384GB RAM and 64 CPU x 2.59 GHz.

\section{Deferred Proofs of the Main Results}\label{sec:proofs}
\subsection{Proof of Lemma~\ref{lemma:strong_duality}}
\label{appx:proof_duality}
Fix $\pi \in \Pi$ and $x \in \X$.
Letting $L = \frac{d Q_{Y\given X = x}}{dP_{Y \given X = x}}$, 
we can rewrite the inner minimization in Equation~\eqref{eqn:objectiveYshift} as
\begin{align}
\label{eq:primal_equiv}
\inf_{L\text{ measurable}}&\EE_{P_{Y\given X}}[Y(\pi(x))L\given X=x] \notag\\
\text{s.t. }&\EE_{P_{Y\given X}}[L\given X=x]=1, \\
&\EE_{P_{Y\given X}}[f_{\text{KL}}(L)\given X=x]\leq\dd,\notag
\end{align}
where the function $f_{\text{KL}}(x)=x \log x$ represents the KL divergence function.
In~\eqref{eq:primal_equiv}, the first constraint reflects that 
$L$ is an likelihood ratio, and the second constraint corresponds to 
the KL divergence 
bound.

For notational simplicity, let $\EE_{x}$ be the shorthand of $\EE_{P_{Y\given X}}[\cdot\given X=x]$. By Theorem 8.6.1 of~\cite{luenberger1997optimization}, 
the Slater's condition is satisfied and strong duality holds:
\begin{align}\label{eq:duality}
\inf_{\substack{\EE_{x}[L]=1,\\
\EE_{x}[f_{\text{KL}}(L)] \le \delta}}
\EE_{x}\big[Y(\pi(x))L\big]
=\max_{\al\geq0, \eta \in \mathbb{R}}\varphi(\al,\eta,x),
\end{align}
where
\begin{align*}
\varphi(\al,\eta,x)=\,&\inf_{L \ge 0}~\mathcal{L}(\al,\eta,L,x),\\
\mathcal{L}(\al,\eta,L,x)=\,&\EE_{x}[Y(\pi(x))L]+\eta\cdot \big(\EE_{x}
[L] - 1 \big) + \al\cdot \big(\EE_{x}[f_{\text{KL}}(L)]-\dd\big)\\
= \,&\EE_x\big[Y(\pi(x))L + \eta(L-1) +\alpha(f_\kl(L) - \delta)  \big] .
\end{align*}
We can explicitly work out the minimum of $\mathcal{L}(\al,\eta,L,x)$, and we have
\begin{align*}
\varphi(\al,\eta,x)
= \EE_{x}\bigg[-\al f^*_{\text{KL}}\bigg(-\frac{Y(\pi(x))+\eta}{\al}\bigg)-\eta-\al\dd\bigg],
\end{align*}
where $f^*_{\text{KL}}(y)=\exp(y-1)$ is the conjugate function of
$f_{\text{KL}}$. Using Equation~\eqref{eq:duality}, we arrive at 
\$ 
\inf_{\substack{\EE_{x}[L]=1,\\
\EE_{x}[f_{\text{KL}}(L)] \le \delta}}
\EE_{x}\big[Y(\pi(x))L\big] 
= - \min_{\alpha \ge 0, \eta \in \mathbb{R}}
\EE_{x}\bigg[\al \exp\bigg(-\frac{Y(\pi(x))+\eta}{\al}-1\bigg)  + \eta + \al\dd\bigg].
\$
The proof is thus completed.


\subsection{Proof of Theorem~\ref{thm:asymconv}}\label{sec:proof-of-thm-asymconv}
\label{appx:proof_asympconv}

For notational simplicity, we drop the dependence on $P$ in $\EE_P$ when the 
context is clear.
The proof of Theorem~\ref{thm:asymconv} makes use of the following lemma, which 
establishes some useful properties of the optimizer $\bmtheta^*_\pi$. The proof of 
Lemma~\ref{lemma:regularity} can be found in Appendix~\ref{appx:regularity-loss}.
\begin{lemma}\label{lemma:regularity} 
For any policy $\pi$, assume that Assumption~\ref{assum:optimum} holds. 
We have the following properties of the optimizer $\bmtheta^*_\pi$.
\begin{enumerate}
    \item [(1)] $\EE\big[\nabla_{\theta}\,\ell(x,Y(\pi(x));\theta)\given X=x\big]=0$ 
    at $\theta = \bmtheta^*_\pi(x)$ for any $x \in \X$.
    \item [(2)] There exists a constant $\xi>0$ such that for any $x$ and 
    $\theta$ satisfying $\|\theta - \bmtheta^*_\pi(x)\|_{2}\le \xi$,  
    \$
    \Big|\ell(x,y;\theta)-\ell(x,y;\bmtheta^*_\pi(x)) - 
    \nabla_\theta \ell(x,y;\bmtheta^*_\pi(x))^\top
    (\theta - \bmtheta^*_\pi(x))\big|
    \leq\bar{\ell}(x,y)\cdot \big\|\theta-\bmtheta^*_\pi(x)\big\|^2_2,
    \$ 
    for some function $\bar{\ell}(x,y)$ such that 
    $\sup_{x\in\mathcal{X}}\EE[\bar{\ell}(x,Y(\pi(x)))\given X=x]<L$ for some $L>0$.
    \item [(3)] There exists a constant $\xi_1>0$ such that
    for any $\bmtheta$ satisfying $\|\bmtheta - \bmtheta^*_\pi\|_{L_{\infty}}
    \le \xi_1$.
    \$
    \big\|\ell(X,Y(\pi(X));\bmtheta(X))-\ell(X,Y(\pi(X));\bmtheta^*_\pi(X))\big\|_{L_2(P_{X,Y(\pi(X)) \given A=\pi(X)})}
    \le C_\ell\|\bmtheta-\bmtheta^*_\pi\|_{L_2(P_{X\given A = \pi(X)})},
    \$ 
    for some constant $C_\ell >0$.
\end{enumerate}
\end{lemma}
We proceed to show the asymptotic normality of $\hat{\V}_\delta(\pi)$.
For each $k\in[K]$, we first define the following oracle quantity:
\[
\Vd^{*(k)}(\pi)=\frac{1}{|\Dk|}\sum_{i\in\Dk}\frac{\mathbbm{1}\{\pi(X_i)=A_i\}}{\po(A_i\given X_i)}
\cdot \big(G_\pi(X_i,Y_i)-g_\pi(X_i)\big)+g_\pi(X_i).
\]
In the sequel, we shall show that $\hVdk(\pi)=\Vd^{*(k)}(\pi)+o_p(n^{-\frac{1}{2}})$. 
We begin by decomposing the difference between $\hVdk(\pi)$ and $\Vd^{*(k)}$:
\begin{align*}
\hVdk&(\pi)-\Vd^{*(k)}(\pi)\\
=&\frac{1}{|\Dk|}\sum_{i\in\Dk}
\Bigg[\frac{\mathbbm{1}\{\pi(X_i)=A_i\}}{\hpik(A_i\given X_i)}\cdot 
\Big(\hat{G}^{(k)}_\pi(X_i,Y_i)-\hgpik(X_i)\Big)
-\frac{\mathbbm{1}\{\pi(X_i)=A_i\}}{\po(A_i\given X_i)} \cdot 
\Big(G_\pi(X_i,Y_i)-g_\pi(X_i)\Big)\Bigg]\\
&\qquad  +\frac{1}{|\Dk|}\sum_{i\in\Dk}\big(\hgpik(X_i)-g_\pi(X_i)\big)\\
=&\underbrace{\frac{1}{|\Dk|}\sum_{i\in\Dk}\frac{\indc\{A_i = \pi(X_i)\}}{\po(A_i\given X_i)}
\cdot \big(\hGpik(X_i,Y_i)-G_\pi(X_i,Y_i)\big)}_{\textnormal{(I)}}\\
&\qquad  
-\underbrace{\frac{1}{|\Dk|}\sum_{i\in\Dk}\bigg(\frac{\indc\{A_i = \pi(X_i)\}}{\hpik(A_i\given X_i)}-\frac{\indc\{A_i = \pi(X_i)\}}{\po(A_i\given X_i)}\bigg) \cdot \big(\hgpik(X_i)-\bgpi^{(k)}(X_i)\big)}_{\textnormal{(II)}}\\
&\qquad  
+\underbrace{\frac{1}{|\Dk|}\sum_{i\in\Dk}\bigg(\frac{\indc\{A_i = \pi(X_i)\}}{\hpik(A_i\given X_i)}-\frac{\indc\{A_i = \pi(X_i)\}}{\po(A_i\given X_i)}\bigg)\cdot \big(\hGpik(X_i,Y_i)-\bgpi^{(k)}(X_i)\big)}_{\textnormal{(III)}}\\
&\qquad  -\underbrace{\frac{1}{|\Dk|}\sum_{i\in\Dk}\frac{\indc\{A_i = \pi(X_i)\}}{\po(A_i\given X_i)}
\cdot \big(\hgpik(X_i)-g_\pi(X_i)\big)+\frac{1}{|\Dk|}\sum_{i\in\Dk}(\hgpik(X_i)-{g}_\pi(X_i))}_{\textnormal{(IV)}}.
\end{align*}

\paragraph{Bounding Term (I).}
Recall that $\bmtheta_\pi^*(x)$ is the minimizer of 
\$
\EE\Big[\ell\big(x,Y(\pi(x)); \theta\big)\biggiven X=x\Big]. 
\$
By the first-order condition established in part (1) of Lemma~\ref{lemma:regularity},  we have 
\#\label{eq:1st_taylor} 
\EE\Big[\nabla_{\theta}\ell\big(x,Y(\pi(x)); \bmtheta(x)\big) 
\biggiven X=x\Big] = 0.
\#
For any $i \in \cD^{(k)}$, by the unconfoundedness condition in Assumption~\ref{assum:pi0Ydistri}, we have 
\$
& \EE\Bigg[\frac{\mathbbm{1}\{A_i = \pi(X_i)\}}{\pi_0(A_i \given X_i)} 
\cdot \Big( \hat G_\pi^{(k)}\big(X_i,Y_i\big)  - G_\pi\big(X_i,Y_i\big) \Big) \bigggiven \cD^{(-k)}\Bigg] \notag\\
= \,&\EE\Bigg[\frac{\mathbbm{1}\{A_i = \pi(X_i)\}}{\pi_0(A_i \given X_i)} 
\cdot \Big( \hat G_\pi^{(k)}\big(X_i,Y_i(\pi(X_i))\big)  - G_\pi\big(X_i,Y_i(\pi(X_i))\big) \Big) \bigggiven \cD^{(-k)}\Bigg]\notag\\
= \,&\EE\Big[\hat G_\pi^{(k)}\big(X_i,Y_i(\pi(X_i))\big)  - G_\pi\big(X_i,Y_i(\pi(X_i))\big) \biggiven \cD^{(-k)}\Big]\\
= \, &\EE\Big[
\ell\big(X_i,Y_i(\pi(X_i)); \hat \bmtheta^{(k)}_\pi(X_i)\big)  
- \ell\big(X_i,Y_i(\pi(X_i)); \bmtheta^*_\pi(X_i)\big)
-\nabla_{\theta}\ell\big(X_i,Y(\pi(X_i)); \bmtheta_\pi^*(X_i)) \biggiven \cD^{(-k)}\Big],
\$
where the last step is due to Equation~\eqref{eq:1st_taylor}.
By Assumption~\ref{assum:convrate}, 
$\| \hat{\bmtheta}^{(k)}_\pi - \bmtheta^*_\pi\|_{L_\infty} = o_P(1)$.
Therefore, for any $\beta \in (0,1)$, there exists $N \in \mathbb{N}_+$ such that 
for $n \ge N$, $\|\hat \theta_\pi^{(k)} - \theta^*_\pi\|_{L_\infty} \le \min(\xi,\xi_1)$.
On the event that $\|\hat \theta_\pi^{(k)}(x) - \theta^*_\pi(x)\|_{L_\infty} \le 
\min(\xi,\xi_1)$
by part (2) of Lemma~\ref{lemma:regularity} and Jensen's inequality, we have 
\$ 
& \Bigg| \EE\bigg[\frac{\mathbbm{1}\{A_i = \pi(X_i)\}}{\pi_0(A_i \given X_i)} 
\cdot \Big( \hat G_\pi^{(k)}\big(X_i,Y_i\big)  - G_\pi\big(X_i,Y_i\big) \Big)\bigg] \bigggiven \cD^{(-k)} \Bigg|\\
\le \, & \EE\bigg[
\Big|\ell(X_i,Y_i(\pi(X_i)); \hat \bmtheta^{(k)}_\pi(X_i))  
- \ell(X_i,Y_i(\pi(X_i)); \bmtheta^*_\pi(X_i))
-\nabla_{\theta}\ell\big(X_i,Y(\pi(X_i)); \bmtheta_\pi^*(X_i)\big)\Big| \Biggiven \cD^{(-k)}\bigg]\\
\le \, & \EE\Big[\bar{\ell}(X_i,Y_i) \cdot \big\|\hat \bmtheta^{(k)}_\pi(X_i) - \bmtheta^*_\pi(X_i)\big\|^2_{2}\Big]
\le L \EE\Big[\big\|\hat \bmtheta_\pi^{(k)}(X_i) - \bmtheta^*_\pi(X_i)\big\|_2^2 \biggiven \cD^{(-k)}\Big]
= L\|\hat\bmtheta^{(k)}_\pi - \bmtheta^*_\pi\|_{L_2(P_X)}^2.
\$
By Chebyshev's inequality, we have for any $t >0$ that
\$ 
& \PP\Bigg(\bigg|\frac{1}{|\D^{(k)}|}\sum_{i\in \cD^{(k)}}\frac{\mathbbm{1}\{A_i = \pi(X_i)\}}
{\pi_0(A_i \given X_i)}\cdot \big(\hat G_\pi^{(k)}(X_i,Y_i) - G_\pi(X_i,Y_i)\big)\\ 
& \qquad \qquad \qquad \qquad  \qquad \qquad - 
\EE\bigg[\frac{\mathbbm{1}\{A = \pi(X)\}}{\pi_0(A\given X)}\cdot \Big(\hat G_\pi^{(k)}(X,Y) - G_\pi(X,Y)\Big) 
\Biggiven \cD^{(-k)}\bigg]\bigg| \ge t
\bigggiven \cD^{(-k)}\Bigg)\\
& \, \le \frac{1}{|\D^{(k)}|t^2}
\text{Var}\bigg(\frac{\indc\{A = \pi(X)\}}{\pi_0(A \given X)}\cdot 
\Big[\hat{G}^{(k)}_\pi(X,Y) - G_\pi(X,Y) \Big] \bigg)\\
\le \, & \frac{\big\|\hat G^{(k)}_\pi - G_\pi \big\|_{L_2(P_{X,Y\given A=\pi(X)})}^2}{\ve^2|\D^{(k)}|t^2}\\
\le \,& \frac{C_\ell\Big(\big\|\hat \theta^{(k)}_\pi - \theta^*_\pi\big\|^2_{L_2(P_{X\given A =\pi(X)})}\Big)}{\ve^2|\D^{(k)}|t^2},
\$
where the last step is due to part (3) of Lemma~\ref{lemma:regularity}.
Combining the above results, we have that
\$ 
\text{term (I)} & = O_P\big(n^{-1/2} \cdot \|\hat \bmtheta^{(k)}_\pi - \bmtheta^*_\pi\|_{L_2(P_{X})}+ 
\|\hat \bmtheta^{(k)}_\pi - \bmtheta^*_\pi\|_{L_2(P_{X})}^2 \big)
 = o_P(n^{-1/2}),
\$ 
where the 
last step is due to Assumption~\ref{assum:convrate}.

\paragraph{Bounding Term (II).}
Applying the Cauchy-Schwarz inequality to term (II), we have
\begin{align*}
& \Bigg|\frac{1}{|\Dk|}\sum_{i\in\Dk}\bigg(\frac{\indc\{A_i = \pi(X_i)\}}
{\hpik(A_i\given X_i)}-\frac{\indc\{A_i = \pi(X_i)\}}{\po(A_i\given X_i)}\bigg)
\cdot \Big(\hgpik(X_i)-\bgpi^{(k)}(X_i)\Big)\Bigg|\\
\leq\,
&\sqrt{\frac{1}{|\Dk|}\sum_{i\in\Dk}\indc\{A_i = \pi(X_i)\}  \cdot \bigg(\frac{1}
{\hpik(A_i\given X_i)}-\frac{1}{\po(A_i\given X_i)}\bigg)^2}\\
&\qquad \qquad \qquad \qquad \qquad \times \sqrt{\frac{1}{|\Dk|}\sum_{i\in\Dk}\mathbbm{1}\{A_i = \pi(X_i)\}
\cdot \big(\hgpik(X_i)-\bgpi^{(k)}(X_i)\big)^2}\\
=&O_P\bigg(\epsilon^{-2}\big\|\hpik - \po\big\|_{L_2(P_{X\given A = \pi(X)})}
\cdot\big\|\hgpik-\bgpi^{(k)}\big\|_{L_2(P_{X \given A = \pi(X)})}\bigg)
=o_P(n^{-1/2}),
\end{align*}
where the next-to-last inequality is due to the lower bound on $\pi_0$ and $\hat \pi^{(k)}$;
the last equality is due to the given convergence rate of the product estimation error 
in  Assumption~\ref{assum:convrate}.

\paragraph{Bounding Term (III).}
By Assumption~\ref{assum:convrate}, for any $\beta \in (0,1)$, there exists 
$N_1 \in \mathbb{N}_+$ such that  for $n\ge N_1$, 
\$
\PP\big(\|\hat \bmtheta^{(k)}_\pi - \bmtheta^*\|_{L_\infty} \le  
\min(\underline{\alpha}, \bar{\eta})/2\big)
\ge 1-\beta.
\$
On the event 
$\|\hat \bmtheta^{(k)}_\pi - \bmtheta^*\|_{L_\infty} \le  \min(\underline{\alpha}, \bar{\eta})/2$, 
\$ 
\big|\hat G_\pi^{(k)}(x,y)\big| = \big|\ell(x,y;\hat{\bmtheta}^{(k)}_{\pi})\big| 
\le \bar \alpha \exp\Big(\frac{\bar y + \bar \eta}{\underline{\alpha}}-1\Big)
+ \bar{\eta} + \bar{\alpha}\delta =: L_g.
\$
Next, for any $i \in \cD^{(k)}$,  
\$ 
& \EE\Bigg[\bigg(\frac{\indc\{A_i = \pi(X_i)\}}
{\hpik(A_i\given X_i)}-\frac{\indc\{A_i = \pi(X_i)\}}{\po(A_i\given X_i)}\bigg)
\cdot \Big(\hGpik(X_i,Y_i)-\bgpi^{(k)}(X_i)\Big) \bigggiven \cD^{(-k)}\Bigg]\\
= \,&
\EE\Bigg[\EE\bigg[\frac{\indc\{A_i = \pi(X_i)\}}
{\hpik(A_i\given X_i)}-\frac{\indc\{A_i = \pi(X_i)\}}{\po(A_i\given X_i)} \Biggiven X_i, \cD^{(-k)}\bigg]\\
& \qquad \qquad \qquad \times \EE\Big[\hGpik(X_i,Y(\pi(X_i)))-\bgpi^{(k)}(X_i)\biggiven X_i,\cD^{(-k)}\Big] \bigggiven \cD^{(-k)}\Bigg]
=0,
\$
where the first step is by the unconfoundedness assumption and 
the second step is due to the fact that $\bgpi^{(k)}$ is the conditional expectation of $\hGpik$.

On the event $\{\|\hat \bmtheta_\pi^{(k)} - \bmtheta^*_\pi\|_{L_\infty} \le  \min(\underline{\alpha}, 
\bar{\eta})\}$. By Chebyshev's inequality, for any $t >0$,
\begin{align*}
& \PP\Bigg(\bigg|\frac{1}{|\Dk|}\sum_{i\in\Dk}\bigg(\frac{\indc\{A_i = \pi(X_i)\}}
{\hpik(A_i\given X_i)}-\frac{\indc\{A_i = \pi(X_i)\}}{\po(A_i\given X_i)}\bigg)
\cdot \big(\hGpik(X_i,Y_i)-\bgpi^{(k)}(X_i)\big)\bigg| \ge t \bigggiven \cD^{(-k)}\Bigg)\\
\le \,& \frac{1}{|\D^{(k)}|t^2} 
\textnormal{Var}\Bigg(\bigg[\frac{\mathbbm{1}\{A_i = \pi(X_i)\}}{\hpik(A_i\given X_i)}
-\frac{\mathbbm{1}\{A_i = \pi(X_i)\}}{\po(A_i\given X_i)}\bigg]\cdot \big(\hGpik(X_i,Y_i)-\bgpi^{(k)}(X_i)\big) 
\bigggiven \cD^{(-k)}\Bigg)\\
\le \,& \frac{1}{|\D^{(k)}|t^2} 
\EE\Bigg[\bigg[\frac{\mathbbm{1}\{A_i = \pi(X_i)\}}{\hpik(A_i\given X_i)}
-\frac{\mathbbm{1}\{A_i = \pi(X_i)\}}{\po(A_i\given X_i)}\bigg]^2\cdot \big(\hGpik(X_i,Y_i)-\bgpi^{(k)}(X_i)\big)^2
\bigggiven \cD^{(-k)}\Bigg]\\
\le \, & \frac{4L_g^2}{|\cD^{(k)}|\ve^4t^2}
\|\hat \pi_0^{(k)} -\pi_0\|^2_{L_2(P_{X\given T = \pi(X)})}.
\end{align*}
The above inequality along with a union bound implies that 
\$ 
\text{term (III) }
=O_P\big(\|\hat \pi^{(k)}_0 - \pi_0\|_{L_2(P_{X\given A=\pi(X)})}/\sqrt{|\cD^{(k)}|}\big)
= o_P(n^{-1/2}),
\$
where the last step is 
by the consistency of $\hat \pi^{(k)}_0$ assumed in Assumption~\ref{assum:convrate}. 

\paragraph{Bounding Term (IV).}
We first show that term (IV) is of zero-mean:
\$ 
& \EE\Bigg[-\frac{1}{|\cD^{(k)}|} \sum_{i\in \cD^{(k)}} 
\frac{\indc\{A_i = \pi(X_i)\}}{\pi_0(A_i\given X_i)}
\cdot \big(\hat g^{(k)}_\pi(X_i) - g_\pi(X_i)\big) 
+ \frac{1}{|\cD^{(k)}|} \sum_{i\in \cD^{(k)}} 
(\hat g^{(k)}_\pi(X_i) - g_\pi(X_i)) \bigggiven \cD^{(-k)} \Bigg] \\
=\, &-\EE\Bigg[\frac{\indc\{A_i = \pi(X_i)\}}{\pi_0(A_i\given X_i)}
\cdot \big(\hat g^{(k)}_\pi(X_i) - g_\pi(X_i)\big) \bigggiven \cD^{(-k)}\Bigg] 
+ \EE\Big[\hat g_\pi^{(k)}(X_i) - g_\pi(X_i) \biggiven \cD^{(-k)}\Big] = 0.
\$
By Chebyshev's inequality, for any $t >0$,
\$ 
& \PP\Bigg(\bigg|\frac{1}{|\Dk|}\sum_{i\in\Dk}\frac{\mathbbm{1}\{\pi(X_i)=A_i\}}{\po(A_i\given X_i)}
\cdot (\hgpik(X_i)-g_\pi(X_i))-\frac{1}{|\Dk|}\sum_{i\in\Dk}\big(\hgpik(X_i)-{g}_\pi(X_i)\big)\bigg|
\ge t \bigggiven \cD^{(-k)}\Bigg)\\
\le \, & \frac{1}{|\cD^{(k)}|t^2}
\textnormal{Var}\Bigg(\frac{\indc\{A_i = \pi(X_i)\}}{\pi_0(A_i\given X_i)} 
\cdot \big(\hgpik(X_i)-g_\pi(X_i)\big) - 
\big(\hat g_\pi^{(k)}(X_i) - g_\pi(X_i)\big) \bigggiven \cD^{(-k)}\Bigg)\\
=\,& \frac{1}{|\cD^{(k)}|t^2} 
\EE\bigg[\frac{1-\pi_0(\pi(X_i) \given X_i)}{\pi_0(\pi(X_i) \given X_i)}
\cdot \big(\hat g_\pi^{(k)}(X_i)
- g_\pi(X_i)\big)^2 \Biggiven \cD^{(-k)}\bigg].
\$
As a result, term (IV) $ = O_P\big(\|\hat g_\pi^{(k)} - g_\pi\|_{L_2(P_X)} /\sqrt{n}\big)$.
Note that 
\$ 
\|\hat g_\pi^{(k)} - g_\pi\|_{L_2(P_X)} 
& = O(\|\hat g_\pi^{(k)} - g_\pi\|_{L_2(P_{X\given A = \pi(X)})})\\
& \le 
O\Big(\|\hat g_\pi^{(k)} - \bar g_\pi\|_{L_2(P_{X \given A = \pi(X)})} + 
\|\bar g_\pi^{(k)} - g_\pi\|_{L_2(P_{X \given A = \pi(X)})}\Big), 
\$
where the first inequality follows from the overlap condition.
By Assumption~\ref{assum:convrate}, 
$\|\hat g_\pi^{(k)} - \bar g_\pi\|_{L_\infty}
= o_P(1)$. Meanwhile, 
\$ 
& \|\bar g_\pi^{(k)} - g_\pi\|_{L_2(P_{X \given A = \pi(X)})}^2\\ 
= \, &\EE\big[(\bar g (X) - g(X))^2 \given A = \pi(X), \cD^{-k}\big]\\
= \, &\EE\bigg[\Big(\EE\big[\ell(X,Y(\pi(X)); \hat \bmtheta^{(k)}_\pi(X)) -
\ell(X,Y(\pi(X));\bmtheta^*_\pi(X)) \given X\big]\Big)^2 \Biggiven  A = \pi(X), \cD^{(-k)}\bigg]\\
\stackrel{\text{(i)}}{\le} \, & \EE\bigg[\Big(\ell(X,Y(\pi(X)); \hat \bmtheta^{(k)}_\pi) -
\ell(X,Y(\pi(X));\bmtheta^*_\pi) \Big)^2 \Biggiven  A = \pi(X), \cD^{(-k)}\bigg]\\
\stackrel{\text{(ii)}}{=} \, &O\Big( \|\hat \bmtheta_\pi^{(k)} - \bmtheta_\pi^*\|_{L_2(P_{X\given A = \pi(X)})}^2 \Big)= o_P(1). 
\$
Above, step (i) follows from Jensen's inequality and step (ii)
from part (3) of Lemma~\ref{lemma:regularity}. 
Combining everything, we have that term (IV) is of rate $o_P(n^{-1/2})$.

\paragraph{Putting Everything Together.}
So far we have shown that for each fold $k\in[K]$, there is 
\[
\hVdk(\pi)-\Vd^{*(k)}(\pi)=o_P(n^{-1/2}).
\]
Averaging over all $k$ folds, we have
\begin{align*}
& \sqrt{n}\cdot \big(\hat{\mathcal{V}}_\delta(\pi)-\Vd(\pi)\big)\\
= \, &\frac{1}{\sqrt{n}}
\sum_{i=1}^n\Bigg\{-\frac{\indc\{A_i = \pi(X_i)\}}{\po(A_i\given X_i)}
\cdot \big(G_\pi(X_i,Y_i)-g_\pi(X_i)\big)-
g_\pi(X_i)-\Vd(\pi)\Bigg\}  + o_P(1),
\end{align*}
By the central limit theorem and Slutsky's theorem. 
\$
\sqrt{n} \cdot \big(\hat{\mathcal{V}}_\delta(\pi)-\Vd(\pi)\big)
\stackrel{\text{d.}}{\rightarrow }\mathcal{N}(0,\sigma^2),
\$
where
\$
\sigma^2 = \text{Var}\bigg(\frac{\indc\{A = \pi(X)\}}{\pi_0(A \given X)}
\cdot \big(G_\pi(X,Y) - g_\pi(X)\big) + g_\pi(X)\bigg).
\$

\subsection{Proof of Theorem~\ref{thm:regret-bound}}\label{sec:proof-of-thm-regret-bound}
By Assumption~\ref{assum:optimum}, taking $\pi(x) \equiv a$ for any $a \in [M]$, there exist
constants $\bar{\alpha}_a, \underline{\alpha}_a, \bar{\eta}_a$ such that 
\$ 
0<\underline{\alpha}_a \le \bmalpha^*_a(x) \le \bar{\alpha}_a, \quad
|\bmeta_a(x)| \le \bar{\eta}_a, \quad \forall x \in \X.
\$
Letting $\underline{\alpha} = \min_{a \in [M]} \underline{\alpha}_a$,
$\bar{\alpha} = \max_{a \in [M]} \bar{\alpha}_a$, 
$\bar{\eta} = \max_{a \in [M]} \bar{\eta}_a$, 
it follows that 
\# \label{eq:bound-on-theta}
0<\underline{\alpha} \le \bmalpha^*_a(x) \le \bar{\alpha}, \quad
|\bmeta_a(x)| \le \bar{\eta}, \quad \forall x \in \X,\forall a \in [M].
\#
For any $a \in [M]$, 
if we take $\pi(x) \equiv a$, then by (1) of Lemma~\ref{lemma:regularity}, 
\$
\EE\big[\nabla_\theta \ell(x,Y(a);\bmtheta^*_a(x)) \given X=x\big] = 0.
\$
By (2) of Lemma~\ref{lemma:regularity}, for any $a \in [M]$,
there exists a constant $\xi_a >0$ such that for any $\|\theta - \bmtheta^*_a(x)\|_2\le \xi_a$  
\$ 
|\ell(x,y;\theta) - \ell(x,y;\bmtheta^*_a(x)) - \nabla_\theta \ell(x,y;\bmtheta^*_a(x))^\top (\theta - \bmtheta^*_a(x))| \le \bar{\ell}_a(x,y) \|\theta - \bmtheta^*_a(x)\|_2^2,
\$
for some function $\bar{\ell}_a(x,y) \le L_a$ for some constant $L_a$.
Similarly, we shall take $\xi = \min_{a \in [M]} \xi_a$, $\bar{\ell}(x,y) = \max_a \bar{\ell}_a(x,y)$,
and $L = \sum_{a \in [M]} L_a$.

By (3) of Lemma~\ref{lemma:regularity}, for any $a \in [M]$, 
there exists a constant $\xi_{1,a} >0$ such that for any $\|\bmtheta - \bmtheta^*_a\|_{L_\infty}\le \xi_{1,a}$,
\$
\big\|\ell(X,Y(a);\bmtheta(X))-\ell(X,Y(a);\bmtheta^*_a(X))\big\|_{L_2(P_{X,Y(a) \given A=a})}
\le C_{\ell,a}\|\bmtheta-\bmtheta^*_a\|_{L_2(P_{X\given A = a})}.
\$
Taking $\xi_1 = \min_{a \in [M]} \xi_{1,a}$ and $C_{\ell} = \sum_{a \in [M]} C_{\ell,a}$,
the above inequality holds for any $a \in [M]$ and any $\|\bmtheta - \bmtheta^*_a\|_{L_\infty}\le \xi_1$.

\subsubsection{Regret decomposition}
The regret bound of Algorithm~\ref{alg:policylearning} builds on the following regret decomposition:
\begin{align}\label{eqn:regretdecomp}
\Regret(\hpi)=\,&\Vd(\pi^{\ast})-\Vd(\hpi)\nonumber\\
= \,&\Vd(\pi^*)-\hVdl(\pi^*) +\hVdl(\pi^*)-\hVdl(\hpi)+\hVdl(\hpi)-\Vd(\hpi)
\nonumber\\
\le \,&\Vd(\pi^*)-\hVdl(\pi^*) +\hVdl(\hpi)-\Vd(\hpi)\nonumber\\
\le \,&2\sup_{\pi\in\Pi}\big|\hVdl(\pi)-\Vd(\pi)\big|,
\end{align}
where the second-to-last step is by the choice of $\hpi$. 
For any $\pi \in \Pi$ and any fold $k\in [K]$, 
we define an intermediate quantity
\$ 
\tilde{\V}^{(k)}_\delta := 
\frac{1}{|\cD^{(k)}|} \sum_{i \in \cD^{(k)}}
\frac{\indc\{A_i = \pi(X_i)\}}{\pi_0(A_i\given X_i)} 
\cdot \big(G_{\pi(X_i)}(X_i,Y_i) - g_{\pi(X_i)}(X_i)\big) + g_{\pi(X_i)}(X_i).
\$ 
Letting $\tilde{\V}_\delta = -\frac{1}{K}\sum_{k=1}^K \tilde{\V}^{(k)}_\delta$, 
we have
\$ 
\big|\hat{\mathcal{V}}^{\text{LN}}_\delta(\pi) -  \Vd(\pi) \big|
& =
\bigg|-\frac{1}{K}\sum^K_{k=1}
\hat{\mathcal{V}}^{\text{LN},(k)}_\delta(\pi) -  \Vd(\pi) \bigg|\\
& \le 
\bigg|\frac{1}{K}\sum^K_{k=1}
\hat{\mathcal{V}}^{\text{LN},(k)}_\delta(\pi) -  
\tilde{\mathcal{V}}_\delta(\pi) \bigg|
+ \bigg|\tilde{\mathcal{V}}_\delta(\pi)- \Vd(\pi) \bigg|\\
& \le 
\sup_{\pi \in \Pi}
\frac{1}{K}\sum^K_{k=1}
\bigg|
\hat{\mathcal{V}}^{\text{LN},(k)}_\delta(\pi) -  
\tilde{\mathcal{V}}^{(k)}_\delta(\pi) \bigg|
+ \sup_{\pi \in \Pi} \bigg|-
\tilde{\mathcal{V}}_\delta(\pi)- \Vd(\pi) \bigg|.
\$ 
Taking the supremum over all $\pi \in \Pi$, we have that
\$ 
\sup_{\pi \in \Pi} \big|\hat{\mathcal{V}}^{\text{LN}}_\delta(\pi) -  \Vd(\pi) \big|
& \le \sup_{\pi \in \Pi} \bigg|
-\tilde{\mathcal{V}}_\delta(\pi)- \Vd(\pi) \bigg|
+ \sup_{\pi \in \Pi}\frac{1}{K}\sum^K_{k=1}
\bigg|
\hat{\mathcal{V}}^{\text{LN},(k)}_\delta(\pi) -  
\tilde{\mathcal{V}}^{(k)}_\delta(\pi) \bigg|.
\$
We shall show that the first term above is $O_P(n^{-1/2})$ 
and the second term is $o_P(n^{-1/2})$.
In the following, we refer to the two terms as the effective term and the negligible 
term, respectively.
The following lemma is essential for establishing the uniform convergence results.
\begin{lemma}\label{lemma:bound-rademacher}
Suppose $h$ is a function of $(x,a,y,\pi(x))$. Given 
a set of data $\{z_i=(x_i,a_i,y_i)\}_{i=1}^n$, suppose that 
$|h(z_i,\pi(x_i))| \le c_i(z_i)$. Then  
the Rademacher complexity 
\$ 
\EE_\epsilon\bigg[\sup_{\pi \in \Pi} 
\Big|\frac{1}{n}\sum^n_{i=1} \epsilon_i h\big(x_i,a_i,y_i,\pi(x_i)\big)\Big|\bigg]
\le \frac{\sqrt{\sum_{i=1}^n c_i(z_i)^2}}{n}\cdot \big(32 + 4\kappa(\Pi)\big),
\$
where $\epsilon_i \stackrel{\textnormal{i.i.d.}}{\sim} \textnormal{Unif}\{\pm1\}$ 
are i.i.d.~Rademacher random variables and $\EE_\epsilon$ means the expectation
over $\epsilon$.
\end{lemma}

\subsubsection{The effective term}
Denote $Z_i = (X_i,A_i,Y_i)$ and take
\$
h(Z_i,\pi(X_i)) = -\frac{\indc\{A_i = \pi(X_i)\}}{\pi_0(A_i\given X_i)} 
\cdot \big(G_{\pi(X_i)}(X_i,Y_i) - g_{\pi(X_i)}(X_i)\big) - g_{\pi(X_i)}(X_i) - \V_{\delta}(\pi).
\$
Under the unconfoundedness assumption in Assumption~\ref{assum:pi0Ydistri},
$\EE[h(Z_i,\pi(X_i))] = 0$. By Equation~\eqref{eq:bound-on-theta}, 
we have 
\$ 
|h(Z_i,\pi(X_i))| \le \frac{6}{\ve} \cdot
\Big(\bar \alpha \cdot \exp\Big(\frac{\bar{\eta}}{\underline{\alpha}} -1\Big) 
+ \bar{\eta} + \bar{\alpha}\delta\Big) =: C_0(\bar{\alpha}, \underline{\alpha}, \bar{\eta}, \delta, \ve). 
\$
Meanwhile, we have write
\$ 
\sup_{\pi \in \Pi} \bigg|\frac{1}{K}\sum^K_{k=1}
-\tilde{\mathcal{V}}^{(k)}_\delta(\pi) -\V_\delta(\pi)\bigg| 
=  \sup_{\pi \in \Pi} \bigg|\frac{1}{K}\sum^K_{k=1}\frac{1}{|\cD^{(k)}|}
\sum_{i\in \cD^{(k)}} h\big(Z_i;\pi(X_i)\big)\bigg| 
= \sup_{\pi \in \Pi} \bigg|\frac{1}{n}\sum^n_{i=1}h\big(Z_i;\pi(X_i)\big)\bigg|. 
\$
Next, we define  
\$ 
f(z_1,\ldots,z_n;\pi) = \frac{1}{n}\sum_{i=1}^{n} h(z_i,\pi(x_i)).
\$
Consider two arbitrary data sets $\{z_i\}_{i=1}^n$ 
and $\{z_i'\}^n_{i=1}$.
We can check that for any $\pi \in \Pi$ and any $j \in [n]$, 
\#\label{eq:bdd_diff}
& \big|f(z_1,\ldots, z_j,\ldots,z_n; \pi)\big| - 
\sup_{\pi' \in \Pi} \big|f(z_1,\ldots,z_j', \ldots,z_n;\pi')\big| \notag \\
\le  & \, \big|f(z_1,\ldots, z_j,\ldots,z_{n}; \pi)\big| - 
\big|f(z_1,\ldots,z_j', \ldots,z_{n};\pi)\big| \notag \\
\le  & \, \sup_{\pi \in \Pi}\big|f(z_1,\ldots, z_j,\ldots,z_{n}; \pi) 
- f(z_1,\ldots,z_j', \ldots,z_{n};\pi)\big| \notag\\
= &\, \sup_{\pi \in \Pi}\frac{1}{n}
\big|
h(z_j;\pi) - h(z_j';\pi)\big| 
\le {C_0(\bar{\alpha}, \underline{\alpha}, \bar{\eta}, \delta,\ve)}/n. 
\#
Above, the first inequality is because of the definition of $\sup$ and 
the second is due to the triangle inequality; the last step is due to the boundedness of $h$.
Taking the supremum over all $\pi \in \Pi$ in~\eqref{eq:bdd_diff}, we have that
\$ 
\sup_{\pi \in \Pi} \big|f(z_1,\ldots, z_j,\ldots,z_{n}; \pi)\big| -
\sup_{\pi \in \Pi} \big|f(z_1,\ldots,z_j', \ldots,z_{n};\pi)\big| 
\le {C_0(\bar{\alpha}, \underline{\alpha}, \bar{\eta}, \delta,\ve)}/n. 
\$
By the bounded difference inequality~\citep[Corollary 2.21]{wainwright2019high},
for any $t>0$, 
\$ 
&\PP\Bigg(\sup_{\pi \in \Pi} \Big|\frac{1}{n}h\big(Z_i, \pi(X_i)\big)\Big| 
-\EE\bigg[ \sup_{\pi \in \Pi} \Big|\frac{1}{n}h\big(Z_i,\pi(X_i)\big)\Big| \bigg] \ge t \Bigg) \\
=& \PP\Bigg(\sup_{\pi \in \Pi} \big|f\big(\{Z_i\}_{i\in [n]}; \pi\big)\big| 
- \EE\bigg[\sup_{\pi \in \Pi} \big|f\big(\{Z_i\}_{i \in [n]}; \pi\big)\big|\Bigg] \ge t 
\Big) 
\le e^{-\frac{2n t^2}{C_0(\bar{\alpha},\underline{\alpha},\bar{\eta},\delta,\ve)^2}}.
\$ 
Take $t = \cbnd 
\sqrt{\frac{1}{2n} \log\big(\frac{1}{\beta}\big)}$. 
Then with probability at least $1-\beta$, 
\$ 
\sup_{\pi \in \Pi} \Big|\frac{1}{n}h\big(Z_i, \pi(X_i)\big)\Big| 
< \EE\bigg[ \sup_{\pi \in \Pi} \Big|\frac{1}{n}h\big(Z_i,\pi(X_i)\big)\Big| \bigg]
+ \cbnd \sqrt{\frac{1}{2n} \log\big(\frac{1}{\beta}\big)}.
\$
It remains to bound the expectation term.
Let $Z_1',\ldots,Z_{n}'$ be an i.i.d.~copy of $Z_1,\ldots,Z_{n}$, 
and let $\epsilon_i \stackrel{\text{i.i.d.}}{\sim} \text{Unif}(\{\pm 1\})$. Then 
\#\label{eq:symmetrization} 
& \EE\Bigg[\sup_{\pi \in \Pi} 
\bigg|\frac{1}{n}\sum_{i\in [n]}h(Z_i,\pi(X_i)) - \EE\Big[h(Z_i,\pi(X_i))\Big]\bigg|\Bigg]\notag\\
=\,& \EE\Bigg[\sup_{\pi \in \Pi} 
\bigg|\frac{1}{n}\sum_{i\in [n]}h(Z_i,\pi(X_i)) - \EE_{Z'}\Big[\frac{1}{n}
\sum_{i\in [n]} h(Z_i',\pi(X_i'))\Big]\bigg|\Bigg] \notag\\
\stackrel{\text{(i)}}{\le}\,& \EE\Bigg[\sup_{\pi \in \Pi} 
\bigg|\frac{1}{n}\sum_{i\in [n]}h(Z_i,\pi(X_i)) - \frac{1}{n}
\sum_{i\in [n]} h(Z_i',\pi(X_i'))\bigg|\Bigg]\notag \\
\stackrel{\text{(ii)}}{=} \,& \EE\Bigg[\sup_{\pi \in \Pi} 
\bigg|\frac{1}{n}\sum_{i\in [n]} \epsilon_i\big(h(Z_i,\pi(X_i))- 
h(Z_i',\pi(X_i'))\big) \bigg|\Bigg],\notag\\
\le \,& 2\EE\Bigg[\sup_{\pi \in \Pi} 
\bigg|\frac{1}{n}\sum_{i\in [n]} \epsilon_i h(Z_i,\pi(X_i)) \bigg|\Bigg] \notag \\
= \,&2\EE\Bigg[\EE_\epsilon\bigg[\sup_{\pi \in \Pi} 
\bigg|\frac{1}{n}\sum_{i\in [n]} \epsilon_i h(Z_i,\pi(X_i)) \bigg| \,\bigg]\Bigg],
\#
step (i) is by Jensen's inequality and 
step (ii) is because of the symmetry of $(Z_i,Z_i')$.

Applying Lemma~\ref{lemma:bound-rademacher}, 
\$ 
\EE_\epsilon\bigg[\sup_{\pi \in \Pi} 
\bigg|\frac{1}{n}\sum_{i\in [n]} \epsilon_i h(Z_i,\pi(X_i)) \bigg| \,\bigg]
\le \frac{2\cbnd}{\sqrt{n}}(32+4\kappa(\Pi)).
\$
Combining the above, for any $\beta \in (0,1)$, 
we have with probability at least $1-\beta$, 
\# \label{eq:bound_term0}
\sup_{\pi \in \Pi} \big|
\tilde{\mathcal{V}}_\delta(\pi) -\V_\delta(\pi)\big|
\le \frac{C_0(\bar{\alpha}, \underline{\alpha}, \bar{\eta}, \delta,\ve)}{\sqrt{n}}
\big(64 + 8\kappa(\Pi) + \sqrt{\log(1/\beta)}\big). 
\#

\subsubsection{Bounding the negligible term}
We now proceed to the negligible term.
For any $\pi \in \Pi$ and any $k \in [K]$, consider the following decomposition:
\$ 
& \hat{\mathcal{V}}^{\text{LN},(k)}_\delta(\pi) - \tilde{\mathcal{V}}^{(k)}_\delta(\pi)\\
= \, & \frac{1}{|\cD^{(k)}|}\sum_{i\in \cD^{(k)}}
\frac{\indc\{A_i = \pi(X_i)\}}{\hat \pi_0(A_i \given X_i)}
\big(\hat G^{(k)}_{\pi(X_i)}(X_i,Y_i) - \hat g^{(k)}_{\pi(X_i)}(X_i)\big) + \hat g^{(k)}_{\pi(X_i)}(X_i)\\
& \qquad  - \frac{1}{|\cD^{(k)}|}\sum_{i\in \cD^{(k)}}
\frac{\indc\{A_i = \pi(X_i)\}}{\pi_0(A_i \given X_i)}
\big(G_{\pi(X_i)}(X_i,Y_i) - g_{\pi(X_i)}(X_i)\big) - g_{\pi(X_i)}(X_i)\\
=& \,  \frac{1}{|\cD^{(k)}|}\sum_{i\in \cD^{(k)}}
\bigg(\frac{\indc\{A_i = \pi(X_i)\}}{\hat \pi_0(A_i \given X_i)} -  
\frac{\indc\{A_i = \pi(X_i)\}}{\pi_0(A_i \given X_i)}\bigg)
\big(\hat G^{(k)}_{\pi(X_i)}(X_i,Y_i) - \bar g^{(k)}_{\pi(X_i)}(X_i)\big)\\
& \, \qquad    
+\frac{1}{|\cD^{(k)}|}\sum_{i\in \cD^{(k)}}
\bigg(\frac{\indc\{A_i = \pi(X_i)\}}{\hat \pi_0(A_i \given X_i)} -  
\frac{\indc\{A_i = \pi(X_i)\}}{\pi_0(A_i \given X_i)}\bigg)
\big(\bar{g}^{(k)}_{\pi(X_i)}(X_i) - \hat g^{(k)}_{\pi(X_i)}(X_i)\big)\\
& \, \qquad    
+ \frac{1}{|\cD^{(k)}|}\sum_{i\in \cD^{(k)}}
\frac{\indc\{A_i = \pi(X_i)\}}{\pi_0(A_i \given X_i)}
\big(\hat G^{(k)}_{\pi(X_i)}(X_i,Y_i) - G_{\pi(X_i)}(X_i,Y_i)\big)\\
& \, \qquad  
-\frac{1}{|\cD^{(k)}|}\sum_{i\in \cD^{(k)}}
\frac{\indc\{A_i = \pi(X_i)\}}{\pi_0(A_i \given X_i)}
\big(\hat g^{(k)}_{\pi(X_i)}(X_i) - g_{\pi(X_i)}(X_i)\big)
+ \frac{1}{|\cD^{(k)}|}\sum_{i \in \cD^{(k)}} \Big(\hat g^{(k)}_{\pi(X_i)}(X_i) - g_{\pi(X_i)}(X_i)\Big).
\$
For notational simplicity, we denote 
\$ 
& K_1(\pi) := \frac{1}{|\cD^{(k)}|}\sum_{i\in \cD^{(k)}}
\bigg(\frac{\indc\{A_i = \pi(X_i)\}}{\hat \pi_0(A_i \given X_i)} -  
\frac{\indc\{A_i = \pi(X_i)\}}{\pi_0(A_i \given X_i)}\bigg)
\big(\hat G^{(k)}_{\pi(X_i)}(X_i,Y_i) - \bar g^{(k)}_{\pi(X_i)}(X_i)\big),\\
& K_2(\pi) := 
\frac{1}{|\cD^{(k)}|}\sum_{i\in \cD^{(k)}}
\bigg(\frac{\indc\{A_i = \pi(X_i)\}}{\hat \pi_0(A_i \given X_i)} -  
\frac{\indc\{A_i = \pi(X_i)\}}{\pi_0(A_i \given X_i)}\bigg)
\big(\bar{g}^{(k)}_{\pi(X_i)}(X_i) - \hat g^{(k)}_{\pi(X_i)}(X_i)\big),\\
&K_3(\pi) :=  \frac{1}{|\cD^{(k)}|}\sum_{i\in \cD^{(k)}}
\frac{\indc\{A_i = \pi(X_i)\}}{\pi_0(A_i \given X_i)}
\big(\hat G^{(k)}_{\pi(X_i)}(X_i,Y_i) - G_{\pi(X_i)}(X_i,Y_i)\big),\\
& K_4(\pi) := 
-\frac{1}{|\cD^{(k)}|}\sum_{i\in \cD^{(k)}}
\frac{\indc\{A_i = \pi(X_i)\}}{\pi_0(A_i \given X_i)}
\big(\hat g^{(k)}_{\pi(X_i)}(X_i) - g_{\pi(X_i)}(X_i)\big)
+ \frac{1}{|\cD^{(k)}|}\sum_{i \in \cD^{(k)}} \Big(\hat g^{(k)}_{\pi(X_i)}(X_i) - g_{\pi(X_i)}(X_i)\Big).
\$ 
We proceed to bound each term separately.
To ease the presentation, we shall write $\EE_k$ and $\PP_k$ as the expectation 
and probability conditioned on $\cD^{(-k)}$, respectively.

\paragraph{Bounding $K_1(\pi)$.}
Here, we take 
\$ 
h_1(Z_i;\pi(X_i)) := \bigg(\frac{\indc\{A_i = \pi(X_i)\}}{\hat \pi_0(A_i \given X_i)} -  
\frac{\indc\{A_i = \pi(X_i)\}}{\pi_0(A_i \given X_i)}\bigg)
\big(\hat G^{(k)}_{\pi(X_i)}(X_i,Y_i) - \bar g^{(k)}_{\pi(X_i)}(X_i)\big).
\$
Since $\bar g^{(k)}_a(X)$ is the conditional expectation of 
$\hat G^{(k)}_a(X,Y(a))$, we have 
\$
\EE_k\big[h_1(Z_i,\pi(X_i)) \given X_i\big]
& = \EE_k\Bigg[
\bigg(\frac{\indc\{A_i = \pi(X_i)\}}{\hat \pi_0(A_i \given X_i)} -  
\frac{\indc\{A_i = \pi(X_i)\}}{\pi_0(A_i \given X_i)}\bigg)\big(\hat G^{(k)}_{A_i}(X_i,Y_i) - \bar g^{(k)}_{A_i}(X_i)\big)
\bigggiven X_i\Bigg] \\
& = \bigg(\frac{\pi_0(\pi(X_i))}{\hat \pi_0(\pi(X_i)\given X_i)} -  1\bigg)
\EE_k\Big[\hat G^{(k)}_{\pi(X_i)}(X_i,Y_i) - \bar g^{(k)}_{\pi(X_i)}(X_i)
\biggiven X_i\Big]
\\
& = 0.
\$
By Assumption~\ref{assum:convrate}, there exists $N_1 \in \mathbb{N}_+$, such that when 
$n \ge N_1$, w.~p.~at least $1-\beta$, 
\$ 
\max_{a \in [M]} \|\hat \bmtheta_a^{(k)} - \bmtheta_a^*\|_{L_\infty} \le  
\max(\bar{\alpha},\underline{\alpha}, \bar{\eta}) /2.
\$
On the event $\{\max_{a \in [M]} \|\hat \bmtheta_a^{(k)} - \bmtheta_a^*\|_{L_\infty} \le  
\max(\bar{\alpha},\underline{\alpha}, \bar{\eta}) /2\}$, we have  for any $a \in [M]$
\$
|\ell(x,y,;\hat \bmtheta_a(x))| \le  
2\bar{\alpha} \exp\Big(\frac{2\bar y + 4\bar{\eta}}{\underline{\alpha}}-1\Big) + 2\bar{\eta} + 2\bar{\alpha}\delta.
\$ 
Letting $\cbndb = 4\bar{\alpha} \exp\Big(\frac{2\bar y + 4\bar{\eta}}{\underline{\alpha}}-1\Big) 
+ 4\bar{\eta} + 4\bar{\alpha}\delta)/\ve^2$, 
We can then check that 
\$ 
|h_1(Z_i;\pi(X_i))| & \le 2\cbndb \cdot \big|\hat \pi_0(\pi(X_i)\given X_i) - 
\pi_0(\pi(X_i) \given X_i)\big| \\
& \le 2\cbndb \cdot \max_{a \in [M]}  
\big|\hat \pi_0(a\given X_i) - 
\pi_0(a\given X_i)\big| =: c_1(X_i).
\$ 
The upper bound is a constant conditional on $X_i$'s and $\cD^{(-k)}$.
We now apply the bounded difference inequality conditional on $X = \{X_i\}_{i\in[n]}$:
\$ 
& \PP_k\Bigg(\sup_{\pi \in \Pi} \Big| \frac{1}{|\cD^{(-k)}|}\sum_{i\in \cD^{(k)}}
h_1(Z_i,\pi(X_i))\Big| - \EE_k\bigg[\sup_{\pi \in \Pi} 
\Big| \frac{1}{|\cD^{(k)}|}\sum_{i\in \cD^{(k)}}
h_1(Z_i,\pi(X_i))\Big| \Biggiven X\bigg] \ge t \bigggiven X\Bigg) \\
\le \, & \exp\Bigg(-\frac{2|\cD^{(k)}|^2t^2}{\sum_{i\in \cD^{(k)}} c_1(X_i)^2}\Bigg).
\$
Taking $t = \sqrt{\sum_{i\in \cD^{(k)}}c_1(X_i)^2\log(1/\beta)}/|\cD^{(k)}|$, 
we have with probability at least $1-\beta$,
\$ 
\sup_{\pi \in \Pi} \Big| \frac{1}{|\cD^{(-k)}|}\sum_{i\in \cD^{(k)}}
&h_1(Z_i,\pi(X_i))\Big| \le \EE_k\bigg[\sup_{\pi \in \Pi}
\Big| \frac{1}{|\cD^{(k)}|}\sum_{i\in \cD^{(k)}}
h_1(Z_i,\pi(X_i))\Big| \Biggiven X\bigg] \\
&
\qquad \qquad \qquad \qquad \qquad 
+ \frac{\sqrt{\sum_{i\in \cD^{(k)}}c_1(X_i)^2}}{|\cD^{(k)}|}\sqrt{\log(1/\beta)}. 
\$
For each $i \in \cD^{(k)}$, we take $A_i'$ and $Y_i'$ 
as i.i.d.~copies of $A_i$ and $Y_i$ conditional on $X_i$, respectively.
By a similar symmetrization argument as in the proof for the effective term, we have
\$ 
& \EE_k\Bigg[\sup_{\pi \in \Pi}
\bigg| \frac{1}{|\cD^{(k)}|}\sum_{i\in \cD^{(k)}}
h_1\big(Z_i,\pi(X_i)\big)\bigg| \Biggiven X\Bigg] \\
= \, &\EE_k\Bigg[\sup_{\pi \in \Pi}
\bigg| \frac{1}{|\cD^{(k)}|}\sum_{i\in \cD^{(k)}}
h_1\big(X_i,A_i,Y_i,\pi(X_i)\big) - 
\EE_{A',Z'}\Big[\frac{1}{|\cD^{(k)}|}\sum_{i \in \cD^{(k)}}
h\big(X_i, A_i',Y_i',\pi(X_i)\big)\Big]\bigg| \Biggiven X\Bigg]  \\
\le \, &\EE_k\Bigg[\sup_{\pi \in \Pi}
\bigg| \frac{1}{|\cD^{(k)}|}\sum_{i\in \cD^{(k)}}
h_1(X_i,A_i,Y_i,\pi(X_i)) - \frac{1}{|\cD^{(-k)}|}\sum_{i \in \cD^{(k)}}
h(X_i, A_i',Y_i',\pi(X_i))\bigg| \bigggiven X\Bigg]\\ 
\le \, &\EE_k\Bigg[\sup_{\pi \in \Pi}
\bigg| \frac{1}{|\cD^{(k)}|}\sum_{i\in \cD^{(k)}}
\epsilon_i \Big(h_1(X_i,A_i,Y_i,\pi(X_i)) - 
h(X_i, A_i',Y_i',\pi(X_i))\Big)\bigg| \bigggiven X\Bigg]\\ 
\le \, &2\EE_k\Big[\sup_{\pi \in \Pi}
\Big| \frac{1}{|\cD^{(k)}|}\sum_{i\in \cD^{(k)}}
\epsilon_i h_1(X_i,A_i,Y_i,\pi(X_i))  \Big| \Biggiven X\Big].
\$
Applying Lemma~\ref{lemma:bound-rademacher} with $c_i = c_1(X_i)$,  
we have that
\$ 
& \EE_\epsilon\Big[\sup_{\pi \in \Pi}
\Big| \frac{1}{|\cD^{(k)}|}\sum_{i\in \cD^{(k)}}
\epsilon_i h_1(X_i,A_i,Y_i,\pi(X_i))  \Big| \Biggiven X\Big]
\le  \frac{2\sqrt{\sum_{i \in \cD^{(k)}} c_1(X_i)^2}}{{|\cD^{(k)}|}}(32 + 4\kappa(\Pi)). 
\$ 
Combining the above, on the event 
$\{\max_{a \in [M]} \|\hat \bmtheta_a^{(k)} - \bmtheta_a^*\|_{L_\infty} \le
\max(\bar{\alpha},\underline{\alpha}, \bar{\eta}) /2\}$,
\$ 
\PP_k\bigg(\sup_{\pi \in \Pi} \big|K_1(\pi)\big| \ge 
\frac{\sqrt{\sum_{i\in \cD^{(k)}}c_1(X_i)^2}}{{|\cD^{(k)}|}}
\big(64 + 8\kappa(\Pi) + \sqrt{\log(1/\beta)}\big) \Biggiven X\bigg) \le \beta.
\$
Since $\big|\hat \pi_0(a \given X) - \pi_0(a \given X)\big|^2 \le 1$, 
\$ 
& \PP_k\bigg(\frac{1}{|\cD^{(k)}|}\sum_{i\in \cD^{(k)}}
\max_{a \in [M]} \big(\hat \pi_0(a \given X) - \pi_0(a \given X)\big)^2
-  \sum_{a \in [M]}\EE\big[(\hat \pi_0(a \given X) - \pi_0(a \given X))^2\big] \ge t\bigg)\\
\le \, &\PP_k\bigg(\frac{1}{|\cD^{(k)}|}\sum_{i\in \cD^{(k)}}
\sum_{a \in [M]} \big(\hat \pi_0(a \given X) - \pi_0(a \given X)\big)^2
-  \sum_{a \in [M]}\EE\big[(\hat \pi_0(a \given X) - \pi_0(a \given X))^2\big] \ge t\bigg)\\
\le \, &\sum_{a \in [M]}\PP_k\bigg(\frac{1}{|\cD^{(k)}|}\sum_{i\in \cD^{(k)}}
 \big(\hat \pi_0(a \given X) - \pi_0(a \given X)\big)^2
- \EE\big[(\hat \pi_0(a \given X) - \pi_0(a \given X))^2\big] \ge t\bigg)\\
\le \, & M\exp\big(-2|\cD^{(k)}|t^2\big).
\$
Taking a union bound, with probability at least $1-3 \beta$, we have that
\$ 
\sup_{\pi \in \Pi} \big|K_1(\pi)\big| & \le  
\frac{2\cbndb}{\sqrt{|\cD^{(k)|}}} 
\Big(20 + 4\kappa(\Pi) + \sqrt{2\log(1/\beta)}\Big) \notag\\
& \qquad \qquad \times \Big(\sum_{a \in[M]}\|\hat \pi_0 -\pi_0\|_{L_2(P_{X \given A = a})}
+ \Big(\frac{1}{2n}\log({M}/{\beta})\Big)^{1/4}\Big). 
\$
Since $\sum_{a \in[M]}\|\hat \pi_0 -\pi_0\|_{L_2(P_{X \given A = a})} = o_P(1)$,
there exists $N_1' \ge N_1$ such that when $n \ge N_1'$, with probability at least $1-\beta/(4K)$,
\#\label{eq:bound_term1}
\sup_{\pi \in \Pi} \big|K_1(\pi)\big| \le \frac{\cbnd}{4\sqrt{n}}.
\#

\paragraph{Bounding $K_2(\pi)$.}
We first note that by Cauchy-Schwarz inequality,
\$
& \bigg|\frac{1}{|\cD^{(k)}|}\sum_{i\in \cD^{(k)}}
\bigg(\frac{\indc\{A_i = \pi(X_i)\}}{\hat \pi_0(A_i \given X_i)} -  
\frac{\indc\{A_i = \pi(X_i)\}}{\pi_0(A_i \given X_i)}\bigg)\big(\bar g^{(k)}_{A_i}(X_i) - \hat g^{(k)}_{A_i}(X_i)\big)\bigg|\\
\le \, & 
\frac{1}{|\cD^{(k)}|\ve^2} \sqrt{\sum_{i\in \cD^{(k)}}\big(\hat \pi^{(k)}_0(\pi(X_i) \given X_i) - \pi_0(\pi(X_i) \given X_i)\big)^2}
\sqrt{\sum_{i\in \cD^{(k)}}\big(\bar g^{(k)}_{\pi(X_i)}(X_i) - \hat g^{(k)}_{\pi(X_i)}(X_i)\big)^2}\\
\le \,&  \frac{1}{|\cD^{(k)}|\ve^2} \sqrt{\sum_{i\in \cD^{(k)}}\sum^M_{a=1}
\big(\hat \pi^{(k)}_0(a \given X_i) - \pi_0(a \given X_i)\big)^2}
\sqrt{\sum_{i\in \cD^{(k)}}\sum^M_{a=1}\big(\bar g^{(k)}_{a}(X_i) - \hat g^{(k)}_{a}(X_i)\big)^2}.
\$
Then for any $t >0$, let 
\$
s = \frac{M}{t\ve^2} \max_{a \in [M]}\Big\{\|\hat \pi_a^{(k)} - \pi_{0,a}^{(k)}\|_{L_2(P_X)}\Big\} 
\max_{a \in [M]}\Big\{\|\bar g_a^{(k)} - \hat g_a^{(k)}\|_{L_2(P_X)}\Big\}.
\$ 
Then 
\$ 
& \PP_k\Bigg(
\max_{\pi \in \Pi}\bigg|\frac{1}{|\cD^{(k)}|}\sum_{i\in \cD^{(k)}}
\bigg(\frac{\indc\{A_i = \pi(X_i)\}}{\hat \pi_0(A_i \given X_i)} -  
\frac{\indc\{A_i = \pi(X_i)\}}{\pi_0(A_i \given X_i)}\bigg)\big(\hat g^{(k)}_{A_i}(X_i) - \bar g^{(k)}_{A_i}(X_i)\big) \bigg|
\ge s\Bigg)\\
\le \, & \PP_k\Bigg( 
\frac{1}{|\cD^{(k)}|\ve^2} \sqrt{\sum_{i\in \cD^{(k)}}\sum^M_{a=1}
\big(\hat \pi^{(k)}_0(a \given X_i) - \pi_0(a \given X_i)\big)^2}
\sqrt{\sum_{i\in \cD^{(k)}}\sum^M_{a=1}\big(\hat g^{(k)}_{a}(X_i) - \bar g^{(k)}_{a}(X_i)\big)^2}
\ge s \Bigg)\\
\le \, & \PP_k\Bigg(
\frac{1}{\ve} \sqrt{\frac{1}{|\cD^{(k)}|}\sum_{i\in \cD^{(k)}}\sum^M_{a=1}
\big(\hat \pi^{(k)}_0(a \given X_i) - \pi_0(a \given X_i)\big)^2} \ge 
\frac{\sqrt{M}}{\sqrt{t}\ve} \max_{a \in [M]} \Big\{\|\hat \pi_a^{(k)} - \pi_{0,a}^{(k)}\|_{L_2(P_X)}\Big\} 
\Bigg) \\
& \qquad + \PP\Bigg(\frac{1}{\ve} \sqrt{\frac{1}{|\cD^{(k)}|}\sum_{i\in \cD^{(k)}}\sum^M_{a=1}\big(\hat g^{(k)}_{a}(X_i) - \bar g^{(k)}_{a}(X_i)\big)^2} \ge
\frac{\sqrt{M}}{\sqrt{t}\ve} \max_{a \in [M]} \Big\{\|\hat g_a^{(k)} - \hat g_a^{(k)}\|_{L_2(P_X)}\Big\} \Bigg)\\  
\le \,&  2t, 
\$
where the last inequality is due to Chebyshev's inequality.
Marginalizing over the randomness of $\cD^{(-k)}$, for any $\beta \in (0,1)$,
we have  with probability at least $1-\beta$  that 
\$
\max_{\pi \in \Pi}|K_2(\pi)| & < \frac{2M}{\beta\ve^2} \max_{a \in [M]}\Big\{\|\hat \pi_a^{(k)} - \pi_{0,a}^{(k)}\|_{L_2(P_X)}\Big\} 
\max_{a \in [M]}\Big\{\|\hat g_a^{(k)} - \bar g_a^{(k)}\|_{L_2(P_X)}\Big\}.
\$
By Assumption~\ref{assum:convrate}, there exists $N_2' \in \mathbb{N}_+$ such that when $n \ge N_2'$,
with probability at least $1-\beta/(4K)$,
\# \label{eq:bound_term2}
\sup_{\pi \in \Pi}|K_2(\pi)| \le \frac{\cbnd}{4\sqrt{n}}.
\#
\paragraph{Bounding $K_3(\pi)$.}
We start by taking 
\$ 
h_3(Z_i,\pi(X_i)) = 
\frac{\indc\{A_i = \pi(X_i)\}}{\pi_0(A_i \given X_i)}\cdot 
\Big[\hat G^{(k)}_{\pi(X_i)}\big(X_i,Y_i(\pi(X_i))\big) 
- G_{\pi(X_i)}\big(X_i,Y_i(\pi(X_i))\big)\Big]. 
\$
For any $\pi \in \Pi$, 
\$ 
& \EE_k\big[h_3(Z_i,\pi(X_i))\given X_i\big]\\
= \, & \EE_k\Big[\frac{\indc\{A_i = \pi(X_i)\}}{\pi_0(A_i \given X_i)}\cdot 
\big(\hat G^{(k)}_{A_i}(X_i,Y_i(\pi(X_i))) - G_{A_i}(X_i,Y_i(\pi(X_i)))\big) \biggiven 
X_i \Big] \\
=\, & \EE_k\Big[\hat G^{(k)}_{\pi(X_i)}(X_i,Y_i(\pi(X_i))) - G_{\pi(X_i)}(X_i,Y_i(\pi(X_i)))\given X_i\Big]\\
=\, & \EE_k\Big[\ell\big(X_i,Y_i(\pi(X_i));\bmtheta^{(k)}_{\pi(X_i)}(X_i)\big) 
- \ell(X_i,Y_i; \bmtheta^*_{\pi(X_i)}(X_i))\\
&\qquad \qquad \qquad  - \nabla \ell(X_i,Y_i(\pi(X_i));\bmtheta^*_{\pi(X_i)}(X_i))^\top 
\big(\hat \bmtheta^{(k)}_{\pi(X_i)}(X_i) - \bmtheta^*_{\pi(X_i)}(X_i)\big) \biggiven X_i\Big],
\$
where the last step follows from part (1) of Lemma~\ref{lemma:regularity}.
By Assumption~\ref{assum:convrate}, for any $\beta \in (0,1)$, 
there exists $N_3 \in \mathbb{N}_+$ such that when $n \ge N_3$,
\$ 
\PP\bigg(\max_{a \in [M]} \|\hat \bmtheta^{(k)}_{a} - \bmtheta^*_a\|_{L_\infty} > 
\min\big(\xi,\bar{\alpha}, \underline{\alpha}, \bar{\eta}\big) /2\bigg)
\le \beta.
\$
On the event $\big\{\max_{a \in [M]} \|\hat \bmtheta^{(k)}_{a} - \bmtheta^*_a\|_{L_\infty}
\le  \min(\xi,\bar{\alpha},\underline{\alpha}, \bar{\eta})/2 \big\}$,
we have 
\$ 
& \Big|\ell\big(X_i,Y_i;\bmtheta^{(k)}_{\pi(X_i)}(X_i)\big) 
- \ell(X_i,Y_i; \bmtheta^*_{\pi(X_i)}(X_i)) - \nabla \ell(X_i,Y_i;\bmtheta^*_{\pi(X_i)}(X_i))^\top 
\big(\hat \bmtheta^{(k)}_{\pi(X_i)} - \bmtheta^*_{\pi(X_i)}\big)\Big|\\
\le\,& \bar{\ell}(X_i,Y_i) \cdot 
\sum_{a \in [M]}\big\|\hat \bmtheta_a(X_i) - \bmtheta^*_a(X_i)\big\|_2^2,
\$
As a result, 
\$ 
\sup_{\pi \in \Pi}
\big|\EE_k[K_3(\pi) \given X]\big| & \le
\sup_{\pi \in \Pi} \frac{1}{|\cD^{(k)}|}
\sum_{i\in \cD^{(k)}} \EE_k\big[h_3(Z_i,\pi(X_i)) \given X_i\big]\\
& \le
\frac{L}{|\cD^{(k)}|}\sum_{i \in \cD^{(k)}}
\sum_{a \in [M]}\big\|\hat \bmtheta_a(X_i) - \bmtheta^*_a(X_i)\big\|_2^2.
\$
On the same event,
\$ 
\big|h_3(Z_i,\pi(X_i))\big| 
& = \bigg|\frac{\indc\{A_i = \pi(X_i)\}}{\pi_0(A_i \given X_i)}
\cdot \Big\{\ell\big(X_i,Y_i(\pi(X_i));\bmtheta^{(k)}_{\pi(X_i)}(X_i)\big) 
- \ell(X_i,Y_i; \bmtheta^*_{\pi(X_i)}(X_i)) \Big\}\bigg|\\
& \le \frac{1}{\ve}  
\Big|\nabla \ell\big(X_i,Y_i(\pi(X_i));\tilde{\bmtheta}_{\pi(X_i)}(X_i)\big)^\top
(\bmtheta^{(k)}_{\pi(X_i)}(X_i) - \bmtheta^*_{\pi(X_i)}(X_i))\Big|\\
& \le\frac{1}{\ve} \big\|\nabla \ell\big(X_i,Y_i(\pi(X_i));\tilde{\bmtheta}_{\pi(X_i)}(X_i)\big)\big\|_2
\big\|\hat \bmtheta^{(k)}_{\pi(X_i)}(X_i)- \bmtheta^*_{\pi(X_i)}(X_i)\big\|_2\\
& \le \cbndc \max_{a \in [M]}
\big\|\hat \bmtheta^{(k)}_a(X_i)- \bmtheta^*_a(X_i)\big\|_2,
\$ 
where $C_2(\bar{\alpha}, \underline{\alpha},\bar{\eta}, \delta,\ve) =
(1+(\bar y + \bar \eta)/\underline{\alpha})e^{(\bar y + \bar \eta)/\underline{\alpha}-1}+\delta+1$ 
is a constant.
Let $\bar{h}_3(Z_i,\pi(X_i)) = h_3(Z_i,\pi(X_i))
- \EE_k[h_3(Z_i,\pi(X_i)) \given X_i]$, and 
we have that 
\$
|\bar{h}_3(Z_i,\pi(X_i))| \le 2\cbndc 
\max_{a \in [M]}\big\|\hat \bmtheta^{(k)}_{a}(X_i)- \bmtheta^*_{a}(X_i)\big\|_2.
\$
Next, we apply the bounded difference theorem conditional on $X_i$'s: 
\$
& \PP_k\Bigg( 
\sup_{\pi\in \Pi}\bigg|
\frac{1}{|\cD^{(k)}|} \sum_{i\in \cD^{(k)}} \bar{h}_3(Z_i,\pi(X_i))\bigg|
- \EE_k\bigg[\sup_{\pi\in \Pi}\bigg|
\frac{1}{|\cD^{(k)}|} \sum_{i\in \cD^{(k)}} \bar{h}_3(Z_i,\pi(X_i))\bigg| 
\bigggiven X\bigg] \ge t \bigggiven X\Bigg)\\
\le \, &\exp\Bigg(- \frac{|\cD^{(k)}|^2t^2}
{2\cbndc^2\sum_{i\in \cD^{(k)}} \max_{a \in [M]}
\big\|\hat \bmtheta^{(k)}_{a}(X_i)- \bmtheta^*_{a}(X_i)\big\|_2^2}\Bigg),
\$
for any $t > 0$. Taking $t = \cbndc\sqrt{2 \sum_{i\in \cD^{(k)}} 
\max_{a \in [M]}\big\|\hat \bmtheta^{(k)}_{a}(X_i)- \bmtheta^*_{a}(X_i)\big\|_2^2 }
/|\cD^{(k)}|$,
we have with probability at least $1-\beta$ that
\$ 
\sup_{\pi\in \Pi}\bigg|
\frac{1}{|\cD^{(k)}|} \sum_{i\in \cD^{(k)}} \bar{h}_3(Z_i,\pi(X_i))\bigg|
\le \EE_k\bigg[\sup_{\pi\in \Pi}\bigg|
\frac{1}{|\cD^{(k)}|} \sum_{i\in \cD^{(k)}} \bar{h}_3(Z_i,\pi(X_i))\bigg| 
\Biggiven X\bigg] \\
\qquad \qquad \qquad \qquad 
+ \frac{\cbndc}{|\cD^{(k)}|}\sqrt{2 \sum_{i\in \cD^{(k)}}
\sum_{a\in[M]} \big\|\hat \bmtheta^{(k)}_{a}(X_i)- \bmtheta^*_{a}(X_i)\big\|_2^2}.
\$
For the expectation term,the same symmetrization argument as in the proof for 
$K_1(\pi)$ leads to
\$
\EE_k\bigg[\sup_{\pi\in \Pi}\bigg|
\frac{1}{|\cD^{(k)}|} \sum_{i\in \cD^{(k)}} \bar{h}_3(Z_i,\pi(X_i))\bigg| 
\Biggiven X\bigg] \le 2\EE\bigg[\sup_{\pi\in \Pi}|\frac{1}{|\cD^{(k)}|}
\sum_{i \in \cD^{(k)}}\Big| \epsilon_i\bar{h}_3(Z_i,\pi(X_i)) \Big| \Biggiven X\bigg]. 
\$
Then by Lemma~\ref{lemma:bound-rademacher}, we have 
\$ 
& \EE_\epsilon\bigg[\sup_{\pi\in \Pi}\Big|\frac{1}{|\cD^{(k)}|}
\sum_{i \in \cD^{(k)}}\Big| \epsilon_i\bar{h}_3(Z_i,\pi(X_i)) \Big|\bigg]\\
\le \, &\frac{2\cbndc}{|\cD^{(k)}|}(32+\kappa(\Pi)) 
\sqrt{\sum_{i \in \cD^{(k)}} 
\sum_{a \in [M]} \|\hat \bmtheta^{(k)}_a(X_i)- \bmtheta^*_a(X_i)\|_2^2}. 
\$ 
By Hoeffding's inequality, we have that 
\$ 
\PP_k\bigg(\frac{1}{|\cD^{(k)}|}\sum_{i \in \cD^{(k)}} 
\|\hat \bmtheta^{(k)}_a(X_i)- \bmtheta^*_a(X_i)\|_2^2 - 
\|\bmtheta_a^{(k)} - \bmtheta_a\|_{L_2(P_X)}^2 \ge 
\|\bmtheta_a^{(k)} - \bmtheta_a\|_{L_\infty}^2 
\sqrt{\frac{1}{2n} \log\Big(\frac{1}{\beta}\Big)} \bigg)\le \beta.
\$
Taking a union bound, with probability at least $1-3\beta$, we have that
\$
& \sup_{\pi \in \Pi} \big|K_3(\pi)\big|\notag\\
\le\, & \frac{\cbndc(130+4\kappa(\Pi))}{\sqrt{|\cD^{(k)}|}} 
\Big(\sum_{a \in [M]}\|\hat \bmtheta_a^{(k)} - \bmtheta_a\|_{L_2(P_X)} +
\sqrt{M(\bar \alpha + \bar \eta)}\Big(\frac{1}{|\cD^{(k)}|}\log\Big(\frac{M}{\beta}\Big)\Big)^{1/4} \Big)\notag\\
&\qquad \qquad + L\Big(\sum_{a \in [M]}\|\hat \bmtheta_a^{(k)} - \bmtheta_a\|_{L_2(P_X)}^2 
+ \frac{M \|\hat \bmtheta_a - \bmtheta_a\|_{L_\infty} \sqrt{\log({M}/{\beta})} }{\sqrt{|\cD^{(k)}|}}
\Big).
\$
By Assumption~\ref{assum:convrate}, $\|\hat \bmtheta_a^{(k)} - \bmtheta_a\|_{L_2(P_X)} = o_P(n^{-1/4})$
and $\|\hat \bmtheta_a^{(k)} - \bmtheta_a\|_{L_\infty} = o_P(1)$,  
so there exists $N_3' \ge N_3$ such that when $n \ge N_3'$, with probability at least $1-\beta/(4K)$,
\# \label{eq:bound_term3}
\sup_{\pi \in \Pi} \big|K_3(\pi)\big| \le \frac{\cbnd}{4\sqrt{n}}.
\#

\paragraph{Bounding $K_4(\pi)$.}
For $K_4(\pi)$, we take 
\$ 
h_4(Z_i,\pi(X_i))
= - \frac{\indc\{A_i = \pi(X_i)\}}{\pi_0(A_i \given X_i)}
\big(\hat g^{(k)}_{\pi(X_i)}(X_i) - g_{\pi(X_i)}(X_i)\big)
+ \big(\hat g^{(k)}_{\pi(X_i)}(X_i) - g_{\pi(X_i)}(X_i)\big).
\$
and therefore $K_4(\pi) = \frac{1}{|\cD|}\sum_{i \in \cD^{(k)}} h_4(Z_i,\pi(X_i))$.
Again by the unconfoundedness assumption, 
\$
\EE_k\big[h_4(Z_i,\pi(X_i)) \big]=0.
\$
Due to the overlap condition, we further have that 
\$ 
|h_4(Z_i,\pi(X_i))| \le \frac{2}{\ve} 
\big|\hat g^{(k)}_{\pi(X_i)}(X_i) - g_{\pi(X_i)}(X_i)\big|
\le \frac{2}{\ve} 
\max_{a \in [M]}\big|\hat g^{(k)}_{a}(X_i) - g_{a}(X_i)\big|.
\$
As before, we apply the bounded difference theorem conditional on $X_i$'s and 
the symmetrization argument to obtain 
\$ 
& \PP\bigg(\sup_{\pi \in \Pi} \big|K_4(\pi) \big| 
-2 \EE_k\Big[\sup_{\pi \in \Pi} \Big|\frac{1}{|\cD^{(k)}|}\sum_{i\in \cD^{(k)}} 
\epsilon_i h_4(Z_i,\pi(X_i))\Big| \Biggiven X \Big] 
\ge t \Biggiven X\bigg)\\
\le \,& \PP\bigg(\sup_{\pi \in \Pi} \big|K_4(\pi) \big| 
-\EE_k\Big[\sup_{\pi \in \Pi} \big|K_4(\pi)\big| \biggiven X \Big] 
\ge t \given X)\\
\le\,& 
\exp\Bigg(-\frac{\ve^2|\cD^{(k)}|^2t^2}{2\sum_{i\in \cD^{(k)}} \max_{a \in [M]}(\hat g^{(k)}_a(X_i) - 
g_a(X_i))^2} \Bigg).
\$
We now apply Lemma~\ref{lemma:bound-rademacher}:
\$ 
\EE_\epsilon\Bigg[\sup_{\pi \in \Pi} \bigg|\frac{1}{|\cD^{(k)}|}\sum_{i\in \cD^{(k)}} 
\epsilon_i h_4(Z_i,\pi(X_i))\bigg| \bigggiven X \Bigg] 
\le \frac{\sqrt{\sum_{i\in \cD^{(k)}}\max_{a \in [M]}
\big(\hat g_a(X_i) - g_a(X_i)\big)^2}}{|\cD^{(k)}|\ve} (64+8\kappa(\Pi)).
\$
By Assumption~\ref{assum:convrate}, there exists $N_4 \in \mathbb{N}_+$, such that when $n \ge N_4$,
with probability at least $1-\beta$, 
\$ 
\max_{a \in [M]} \|\hat \bmtheta^{(k)}_{a} - \bmtheta^*_a\|_{L_\infty} \le 
\max(\xi, \bar{\alpha}, \underline{\alpha}, \bar{\eta})/2. 
\$
On the event $\{\max_{a \in [M]} \|\hat \bmtheta^{(k)}_{a} - \bmtheta^*_a\|_{L_\infty} \le
\max(\xi, \bar{\alpha}, \underline{\alpha}, \bar{\eta})/2\}$, we have for any $a \in [M]$ that
\$ 
\big|\ell(x,y;\hat \bmtheta_a(x))\big| \le 2 C_1(\bar{\alpha}, \underline{\alpha}, \bar{\eta},\delta,\ve).
\$
On the same event, by Hoeffding's inequality, we have that 
\$
\PP_k\bigg(\frac{1}{|\cD^{(k)}|}\sum_{i\in \cD^{(k)}} 
(\hat g_a(X_i) - g_a(X_i))^2 - \|\hat g_a^{(k)} - g_a\|_{L_2(P_X)}^2 \ge t 
\bigg)\le \exp\Big(-\frac{t^2|\cD^{(k)}|}{8 \cbndb^2}\Big).
\$
Taking a union bound, we have with probability at least $1-2\beta$ that
\$
\max_{\pi \in \Pi} \big|K_4(\pi)\big| \le & 
\frac{1}{\ve\sqrt{|\cD^{(k)}|}} \Big(128 + 16 \kappa(\Pi) +\sqrt{2\log(1/\beta)} \Big)\notag\\
& \qquad \times \Big(\sum_{a \in [M]}\|\hat g_a^{(k)} - g^*_a \|_{L_2(P_X)} 
+ 2M\sqrt{\cbndb}(\log(M/\beta)/n)^{1/4}\Big). 
\$
By Assumption~\ref{assum:convrate}, $ \sum_{a \in [M]}\|\hat g_a^{(k)} - g^*_a \|_{L_2(P_X)} = o_P(1)$, so 
there exists $N_4' \ge N_4$ such that when $n \ge N_4'$, with probability at least $1-\beta/(4K)$,
\# \label{eq:bound_term4}
\sup_{\pi \in \Pi} \big|K_4(\pi)\big| \le \frac{\cbnd}{4\sqrt{n}}.
\#

Combining~\eqref{eq:bound_term0}-\eqref{eq:bound_term4} and taking 
a union bound over $k\in[K]$, when $n \ge \max(N_1,N_2,N_3,N_4)$ 
we have that with probability at least $1-\beta$,
\$ 
\sup_{\pi \in \Pi} \big|\hat{\V}^{\text{LN}}_\delta(\pi) - \tilde{\V}_\delta(\pi)\big| \le
\frac{\cbnd}{\sqrt{n}}.
\$
We have thus completed the proof of Theorem~\ref{thm:regret-bound}.

\subsection{Proof of Theorem~\ref{thm:lower_bnd}}
\label{sec:proof_lower_bnd}
We first state somes results from~\cite{si2023distributionally}
that will be used in the proof. For any $p,q\in[0,1]$, define 
\$
D(p \,\|\, q) = p\log\Big(\frac{p}{q}\Big) + (1-p)\log\Big(\frac{1-p}{1-q}\Big),
\text{ and }g_\delta(q) = \inf_{p: D_\kl(p\,\|\, q) \le \delta} p,
\$
\begin{lemma}[Adapted from Lemma A17 of \citet{si2023distributionally}]
\label{lemma:g_property}
For $\delta \le 0.2$, $g_\delta(q)$ is differentiable and $g_\delta'(q) \ge 1/2$ for 
$q \in [0.4,0.6]$.  
\end{lemma}
Note that our definition of $g_\delta(q)$ is slightly different from that in~\cite{si2023distributionally},
so we include the proof of Lemma~\ref{lemma:g_property} in Appendix~\ref{appx:proof_g_property} 
for completeness.

For notational simplicity, we use $d$ to denote the Natarajan dimension of the policy class $\Pi$. 
By the definition of Natarajan dimension, there exists a set of $d$ data points $\{x_1,\ldots,x_d\}
\subseteq \mathcal{X}$ shattered by $\Pi$: there exist two functions $f_{-1},f_1: \{x_1,\ldots,x_d\}
\mapsto [M]$ such that $f_{-1}(x_j) \neq f_1 (x_j)$ for any $j\in[d]$ and for any $\sigma \in \{-1.1\}^d$, 
there exists $\pi \in \Pi$, such that $\pi(x_j) = f_{\sigma_j}(x_j)$ for all $j\in[d]$. 

Next, we construct a class of distributions indexed by $\sigma \in \{\pm1\}^d$ that are ``hard instances'' for the 
learning problem.  Fix any $\sigma \in \{\pm 1\}^d$, we construct distribution $P_\sigma$ as follows.
First, the covariate are drawn uniformly 
from $\{x_1,\ldots,x_d\}$, i.e., 
\$ 
X_i \stackrel{\text{i.i.d.}}{\sim} \text{Unif}\big(\{x_1,\ldots,x_d\}\big).
\$
Given $X_i$, the action $A_i$ is chosen according to the 
behavior policy $\pi_0$, where for any $j\in[d]$,
\$ 
\pi_0(f_1(x_j) \given x_j) = \pi_0(f_{-1}(x_j) \given x_j) = \frac{\varepsilon}{2},
\text{ and } \pi_0(a \given x_j) = \frac{1-\varepsilon}{K-2} \text{ for all } a \neq f_1(x_j), f_{-1}(x_j).
\$
The potential outcomes are generated as follows:
\$ 
&Y_i(f_{1}(x_j)) \given X_i = x_j \sim \bar{y} \cdot \text{Bern}\Big(\frac{1 + \sigma_j \Delta}{2}\Big),
~Y_i(f_{-1}(x_j)) \given X_i = x_j \sim \bar{y}\cdot \text{Bern}\Big(\frac{1 - \sigma_j \Delta}{2}\Big), \\
& \text{ and } Y_i(a) = \bar y \cdot \text{Bern}(1/4) \text{ for all } a \neq f_{1}(x_j), f_{-1}(x_j),
\$
where $\Delta \in (0,0.1)$ is some constant to be determined later.
Note that the distribution of $(X_i,A_i)$ does not depend on $\sigma$.
By construction, it is clear that the data-generating process satisfies 
Assumption~\ref{assum:pi0Ydistri}. For any $p\in\{(1+\Delta)/2, (1-\Delta)/2,1/4\}$, 
$ \log(1/(1-p)) > \delta$.
Therefore, the data-generating process also satisfies Assumption~\ref{assum:essinf}.
As for Assumption~\ref{assum:optimum}, it suffices to check 
the Bernoulli distributions with parameters $(1+\Delta)/2,(1-\Delta)/2,1/4$, and 
can be verified. Since $n \ge d^2$, we can obtain $\hat \bmtheta$, $\hat g$, 
and $\hat{\pi}_0$ that converges at rate $O_P(n^{-1/4})$ (by stratifying on $X$), 
thereby satisfying Assumption~\ref{assum:convrate}.

We now proceed to establish the lower bound.
For any policy learning algorithm that returns $\hat \pi$, 
the worst-case regret is lower bounded by the average 
regret over the class of hard instances we have constructed above: 
\$ 
\sup_{P \in \mathcal{P}} \EE_{P^n}[\mathcal{R}(\hat \pi)]
\ge \frac{1}{2^d} \sum_{\sigma \in \{\pm 1\}^d} \EE_{P^n_\sigma}[\mathcal{R}_\delta(\hat \pi)]. 
\$   
We now focus on the right-hand side above. Fix $\sigma \in \{\pm 1\}^d$. 
Recall that $\mathcal{R}_\delta(\hat \pi) = \mathcal{V}_\delta(\pi^*) - \mathcal{V}_\delta(\hat \pi)$.
For the optimal policy value, there is 
\#\label{eq:opt_val0} 
\mathcal{V}_\delta(\pi^*) & = 
\max_{\pi \in \Pi} ~ \EE_{P_{\sigma,X}}
\bigg[\inf_{Q_{Y\given X} \in \mathcal{P}(P_{\sigma,Y\given X},\delta)} 
\EE_{Q_{Y\given X}}\big[Y(\pi(X)) \given X\big] \bigg] \notag \\
& \stackrel{\text{(i)}}{=} \max_{\pi \in \Pi} \max_{\bmalpha,\bmeta}~
\EE_{P_\sigma}\bigg[-\bmalpha(X) \exp\Big(-\frac{Y(\pi(X)) + \bmeta(X)}{\bmalpha(X)} -1\Big)
- \bmeta(X) - \bmalpha(X)\delta\bigg] \notag \\
& = \max_{\bmalpha,\bmeta} \max_{\pi \in \Pi} ~
\EE_{P_\sigma}\bigg[-\bmalpha(X) \exp\Big(-\frac{Y(\pi(X)) + \bmeta(X)}{\bmalpha(X)} -1\Big)
- \bmeta(X) - \bmalpha(X)\delta\bigg],
\#
where the step (i) follows from the duality result in Proposition~\ref{lemma:strong_duality}.
We now take a closer look at the expectation above: by the construction of $P_\sigma$,
\$
& \EE_{P_\sigma}\bigg[-\bmalpha(X) \exp\Big(-\frac{Y(\pi(X)) + \bmeta(X)}{\bmalpha(X)} -1\Big)
- \bmeta(X) - \bmalpha(X)\delta\bigg] \\
= ~& \frac{1}{d}\sum^d_{j=1}
\EE_{P_\sigma}\bigg[-\bmalpha(x_j) \exp\Big(-\frac{Y(\pi(x_j)) + \bmeta(x_j)}{\bmalpha(x_j)} -1\Big)
- \bmeta(x_j) - \bmalpha(x_j)\delta\Biggiven X = x_j\bigg] \\
= ~& \frac{1}{d}\sum^d_{j=1}
-\bmalpha(x_j)\exp\Big(-\frac{\bmeta(x_j)}{\bmalpha(x_j)}-1\Big) 
\cdot \EE_P\bigg[\exp\Big(-\frac{Y(\pi(x_j))}{\bmalpha(x_j)} \Big)
\Biggiven X = x_j\bigg] - \bmeta(x_j) - \bmalpha(x_j)\delta.
\$
Letting $p_j = \PP(Y(\pi(x_j)) = 1\given X = x_j)$, we have 
\$ 
\EE_P\bigg[\exp\Big(-\frac{Y(\pi(x_j))}{\bmalpha(x_j)} \Big)
\Biggiven X = x_j\bigg]
& = p_j \exp(-{1}/{\bmalpha(x_j)}) + 1-p_j,
\$
which is decreasing in $p_j$ and is minimized when $\pi(x_j) = f_{\sigma_j}(x_j)$. 
By construction, such a policy $\pi$ is in $\Pi$. As a result,
\$ 
\eqref{eq:opt_val0} & = 
\max_{\bmalpha,\bmeta} ~\frac{1}{d} \sum_{j=1}^d
\EE_{P_\sigma}\bigg[-\bmalpha(x_j) \exp\Big(-\frac{Y(f_{\sigma_j}(x_j)) 
+ \bmeta(x_j)}{\bmalpha(x_j)} -1\Big) - \bmeta(x_j) - \bmalpha(x_j)\delta \Biggiven X = x_j\bigg]\\
& =\frac{1}{d} \sum_{j=1}^d  \max_{\alpha,\eta}~
\EE_{P_\sigma}\bigg[-\alpha \exp\Big(-\frac{Y(f_{\sigma_j}(x_j)) 
+ \eta}{\alpha} -1\Big) - \eta - \alpha \delta \Biggiven X = x_j\bigg]\\
& = \frac{1}{d}\sum^d_{j=1} \inf_{Q_{Y\given X} \in \mathcal{P}(P_{\sigma, Y\given X = x_j}, \delta)}
\EE_{Q_{Y\given X}}\big[Y(f_{\sigma_j}(x_j)) \given X = x_j\big] = g\Big(\frac{1 + \Delta}{2}\Big).
\$
The last step is because $Y(f_{\sigma_j}(x_j)) \given X = x_j \sim \text{Bern}((1+\Delta)/2)$.
Similarly, for $\mathcal{V}(\hat \pi)$,  we have
\$ 
\mathcal{V}_\delta(\hat \pi) & = 
\EE_{P_{\sigma, X}}\bigg[\inf_{Q_{Y\given X} \in \mathcal{P}(P_{\sigma, Y\given X}, \delta)}
\EE_{Q_{Y\given X}}\big[Y\big(\hat \pi(X)\big) \given X \big]\bigg] \\
& = \frac{1}{d}\sum^d_{j=1} 
\inf_{Q_{Y\given X} \in \mathcal{P}(P_{\sigma, Y\given X = x_j}, \delta)}
\EE_{Q_{Y\given X}}\big[Y(\hat \pi(x_j)) \given X = x_j\big]\\
& = \frac{1}{d} \sum^d_{j=1} 
\indc\big\{\hat \pi(x_j) = f_{\sigma_j}(x_j)\big\} g\Big(\frac{1+\Delta}{2}\Big)
+ \indc\big\{\hat \pi(x_j) = f_{-\sigma_j}(x_j) \big\} g\Big(\frac{1-\Delta}{2}\Big)\\
& \qquad \qquad \qquad + \indc\big\{\hat \pi(x_j) \neq f_{\sigma_j}(x_j), f_{-\sigma_j}(x_j) \big\} g(1/4).
\$
Combining the calculation above, we have
\$ 
\mathcal{R}(\hat \pi) & = 
\frac{1}{d}\sum^d_{j=1} \indc\{\hat \pi(x_j) = f_{-\sigma_j}(x_j)\} 
\cdot \bigg\{g\Big(\frac{1+\Delta}{2}\Big) - g\Big(\frac{1 - \Delta}{2}\Big)\bigg\} \\ 
& \qquad \qquad \qquad  
+ \indc\{\hat \pi(x_j) \neq f_{\sigma_j}(x_j), f_{-\sigma_j}(x_j)\} \cdot 
\bigg\{g\Big(\frac{1+\Delta}{2}\Big) - g(1/4)\bigg\}\\
& \stackrel{\text{(i)}}{\ge} 
\frac{1}{d}\sum^d_{j=1} 
+ \indc\{\hat \pi(x_j) \neq f_{\sigma_j}(x_j)\} \cdot 
\bigg\{g\Big(\frac{1+\Delta}{2}\Big) - g(1/4)\bigg\}\\
& \stackrel{\text{(ii)}}{\ge} \frac{1}{d} \sum^d_{j=1} \indc\{\hat \pi(x_j) \neq f_{\sigma_j}(x_j)\} 
\cdot g'(\xi) \Delta \stackrel{\text{(iii)}}{\ge} \frac{\Delta}{2d}\sum^d_{j=1} \indc\{\hat \pi(x_j) \neq f_{\sigma_j}(x_j)\},
\$
where  step (i) uses that $g$ is non-decreasing (c.f.~\citet[Proposition 1]{cauchois2024robust});
in step (ii), $\xi \in ((1-\Delta)/2,(1+\Delta)/2)$,
and step (iii) follows from Lemma~\ref{lemma:g_property}.

Next, we denote $\sigma[j]$ to be the 
vector $\sigma$ with the $j$-th element flipped. Then, we have
\#\label{eq:lower_bound2} 
\frac{1}{2^d}\sum_{\sigma \in \{\pm 1\}^d} \EE_{P^n_\sigma}[\mathcal{R}(\hat \pi)]
& \ge 
\frac{1}{2^d}\sum_{\sigma \in \{\pm 1\}^d} 
\EE_{P^n_\sigma}\bigg[\frac{\Delta}{2d} \sum^d_{j=1}\indc\{\hat \pi(x_j) \neq f_{\sigma_j}(x_j)\} \bigg]\notag \\
& = \frac{\Delta}{d2^{d+1}} \sum^d_{j=1}
\sum_{\sigma: \sigma_j = 1} 
\bigg\{\PP_{P_\sigma^n}\big(\hat \pi(x_j) \neq f_{1}(x_j)\big) + 
\PP_{P_{\sigma[j]}^n}\big(\hat \pi(x_j) \neq f_{-1}(x_j)\big) \bigg\}\notag \\ 
& \ge \frac{\Delta}{d2^{d+1}} \sum^d_{j=1}
\sum_{\sigma: \sigma_j = 1} 
\PP_{P_\sigma^n}\big(\hat \pi(x_j) \neq f_{1}(x_j)\big) + 
\PP_{P_{\sigma[j]}^n}\big(\hat \pi(x_j) = f_{1}(x_j)\big) \notag \\ 
& \ge \frac{\Delta}{d2^{d+1}} \sum^d_{j=1}
\sum_{\sigma: \sigma_j = 1} \big(1 - D_{\text{TV}}(P_{\sigma}^n, P_{\sigma[j]}^n)\big),
\#
where the last step follows from the definition of the TV distance.
By Pinsker's inequality, there is 
\$ 
D_{\text{TV}}^2\big(P_{\sigma}^n, P_{\sigma[j]}^n\big)
& \leq \frac{1}{2}D_{\text{KL}}\big(P_{\sigma}^n \,\|\, P_{\sigma[j]}^n\big)\\
& = \frac{1}{2}\sum^n_{i=1}\EE_{P_\sigma}
\bigg[\log\Big(\frac{\text{d}P_{\sigma}}{\text{d}P_{\sigma[j]}}(X_i,A_i,Y_i)\Big)\bigg]\\
& = \frac{1}{2}\sum^n_{i=1}\EE_{P_\sigma}
\bigg[\indc\big\{X_i = x_j, A_i = f_{\pm 1}(x_j)\big\}
\cdot \Delta \log\Big(\frac{1+\Delta}{1-\Delta}\Big)\bigg]\\
& \le \frac{3n\varepsilon}{2d}\Delta^2,
\$
where the last step follows from $x \log(\frac{1+x}{1-x}) \le 3x^2$, for $x \in (0,1/3)$.
Take $\Delta = \frac{1}{15}\sqrt{\frac{d}{n\varepsilon}}$ --- this is 
possible since $n \ge d^2$ and $d \ge 4/(9\varepsilon)$ and then
\$ 
\eqref{eq:lower_bound2} \ge \sqrt{\frac{d}{n}} \frac{d 2^{d-1}}{15 d2^{d+2}} 
= \frac{1}{120}\times \sqrt{\frac{d}{n\varepsilon}}.
\$

\section{Generalization to Identifiable Covariate 
Distribution Shift}\label{app:x-shift}
We now extend our methodology to handle situations where 
shift in $X$ and $Y\given X$ distributions are both present.
We note that, in most practical cases, the decision maker has access to the covariates in the target environment, making the shift covariate distribution identifiable and estimable --- it is therefore unnecessary to guard against the worst-case shift.

\paragraph{Method.}
Formally, suppose we have access to a training dataset $\mathcal{D}$ collected in an environment $P$, and aims at learning 
a policy that behaves well in the environment $Q$. Here, 
we assume that $Q_{Y\given X} \in \mathcal{P}(P_{Y\given X},\delta)$ for some given 
radius $\delta>0$ but do not impose any 
constraints on $X$ distribution shift
except that $Q_X$ is absolutely continuous with respect to $P_X$ and that the likelihood ratio is bounded.
Letting $r(x) = \frac{dQ_X}{dP_X}(x)$, we can estimate 
$r$ with $X$ from $P$ and $Q$ using standard tools 
(by directly estimating the density ratio or 
by means of classification algorithms).
For any policy $\pi$, the distributional robust policy value 
can be written as 
\begin{equation}\label{eqn:x-shift-obj}
\mathcal{V}_\dd(\pi):=\EE_{Q_X}\Big[\inf_{Q_{Y\given X} 
\in \mathcal{P}(P_{Y\given X},\delta)}
\EE_{Q_{Y\given X}}\big[Y(\pi(X)) \given X\big]\Big].
\end{equation}
Note that the inner expectation of~\eqref{eqn:x-shift-obj} 
is the same as 
in~\eqref{eqn:objectiveYshift}. By Lemma~\ref{lemma:strong_duality}, there is 
\begin{align*}
\eqref{eqn:x-shift-obj} 
& = \EE_{Q_X\times P_{Y\given X}}\Bigg[\bmalpha^*_\pi(X)\exp\Big(-\frac{Y(\pi(X))+\bmeta^*_\pi(X)}
{\bmalpha^*_\pi(X)}\Big)+\bmeta^*_\pi(X)+
\bmalpha^*_\pi(X)\dd\Bigg] \notag \\
& = \EE_{Q_X \times P_{Y\given X}}\Bigg[\frac{\mathbbm{1}\{A = \pi(X)\}}{\pi_0(A\given X)}
\bigg(\bmalpha^*_\pi(X)\exp\Big(-\frac{Y(A)+\bmeta^*_\pi(X)}
{\bmalpha^*_\pi(X)}\Big)+\bmeta^*_\pi(X)+
\bmalpha^*_\pi(X)\dd\bigg)\Bigg] \notag \\
& = \EE_{P}\Bigg[r(X)\frac{\mathbbm{1}\{A = \pi(X)\}}{\pi_0(A\given X)}
\bigg(\bmalpha^*_\pi(X)\exp\Big(-\frac{Y(A)+\bmeta^*_\pi(X)}
{\bmalpha^*_\pi(X)}\Big)+\bmeta^*_\pi(X)+
\bmalpha^*_\pi(X)\dd\bigg)\Bigg].
\end{align*}
The above expression ensures that the robust policy value 
is identifiable with the data accessible to the decision maker.
With this representation, subsequent policy evaluation and 
learning are similar to the pure concept drift case, 
and we provide the adaptation below.

In addition to $\mathcal{D}$, we let $\tilde{\cD} = 
\{X_i\}_{i=1}^m$ denote the covariates from environment 
$Q$, i.e., $X_i \stackrel{\text{i.i.d.}}{\sim} Q_X$.
Assume that $\lim_{m,n\rightarrow \infty } m/n = \gamma$.
As before, we adopt a $K$-fold cross-fitting scheme, 
where we split both $\cD$ and $\tilde \cD$ into $K$ 
non-overlapping equally-sized folds.
For $k\in[K]$, we use $\cD^{(k+1)}$ and $\tilde{\cD}^{(k+1)}$ 
to obtain $\hat \pi_0^{(k)}$, $\hat r^{(k)}$, and 
$(\hat \bmalpha^{(k)}_{\pi}, \hat \bmeta_\pi^{(k)})$
as estimates of $\pi_0$, $r$, and $(\bmalpha_\pi^*,\bmeta_\pi^*)$,
respectively; we then use $\cD^{(k+2)}$ and $\tilde{\cD}^{(k+2)}$
to obtain $\hat g_\pi^{(k)}$ as an estimate for $\bar g_\pi^{(k)}$. The estimator of the $k$-th fold is 
\begin{align*}
\hat{\mathcal{V}}^{(k)}_\delta(\pi) 
& = \frac{1}{\cD^{(k)}}\sum_{i \in \cD^{(k)}} 
\frac{\hat r^{(k)}(X_i) \mathbbm{1}\{A_i = \pi(X_i)\}}
{\hat \pi_0^{(k)}(A_i \given X_i)}
\cdot \big(\hat G^{(k)}_\pi(X_i,Y_i) - \hat g_\pi^{(k)}(X_i)\big) 
+\frac{1}{|\tilde{\cD}^{(k)}|}
\sum_{i\in \tilde{\cD}^{(k)}}
\hat g_\pi^{(k)}(X_i),
\end{align*}
and the final robust policy value estimator is 
$\hat{\mathcal{V}}_\delta(\pi)
= - \frac{1}{K} \sum_{k=1}^K \hat{\mathcal{V}}^{(k)}_\delta(\pi)$. 
We then obtain the learned policy via 
\[
\hat \pi_{\text{LN}} = \argmax{\pi \in \Pi}~\hat{\mathcal{V}}(\pi),
\] where we apply the same computational trick as 
in the pure concept shift case.

\paragraph{Theoretical Guarantees.}
We now extend the theoretical guarantees  
to the general case. Since the proof is 
similar to the pure concept drift case,
the proof sketch is provided.
As a prerequisite, we
modify Assumption~\ref{assum:convrate} to be:
\begin{assumption}\label{assum:convrate_general}
For any policy $\pi$, assume that for each $k\in[K]$, the estimators $\hat \pi_0^{(k)}$, $\hat r^{(k)}$, 
$\hat g_\pi^{(k)}$, and the empirical risk optimizer $\hat \bmtheta_\pi^{(k)}$ satisfy
\begin{align*}
& \|\hat r^{(k)}/\hpik-r/\po\|_{L_2(P_{X\given A = \pi(X)})} = o_P(n^{-\gamma_1}),
~\|\hgpik- \bar{g}_{\pi}^{(k)}\|_{L_2(P_{X\given A = \pi(X)})}=o_P(n^{-\gamma_2}),\\
& \|\hat{\bmtheta}_\pi^{(k)}-\bmtheta^*_\pi\|_{L_2(P_{X \given A= \pi(X)})}=o_P(n^{-\frac{1}{4}}),~
\|\hat \bmtheta_\pi^{(k)}-\bmtheta^*_\pi\|_{L_\infty}=o_P(1),
\end{align*}
for some $\gamma_1,\gamma_2 \ge 0$ and $\gamma_1 + \gamma_2 \ge \frac{1}{2}$.
\end{assumption}

Theorems~\ref{thm:asymconv_general}  
and~\ref{thm:regret-bound-general} establish the asymptotic 
normality of the policy value estimator and the 
regret upper bound.
\begin{theorem}\label{thm:asymconv_general}
Suppose Assumptions~\ref{assum:pi0Ydistri},~\ref{assum:essinf},~\ref{assum:optimum}, and~\ref{assum:convrate_general} hold. 
Additionally assume that $dQ_X/dP_X \le C$, a.s.,
for 
some constants $C>0$, and that 
$\lim_{m,n \rightarrow \infty} m/n = \gamma$.
For any policy $\pi: \X \mapsto [M]$, we have $\sqrt{n}\cdot \big(\hVd(\pi)-\Vd(\pi)\big) \stackrel{\textnormal{d}}{\rightarrow} N(0,\sigma^2_{\pi})$,
where
\begin{align*}
& \sigma^2_\pi = \textnormal{Var}\bigg(\frac{r(X)\indc\{A = \pi(X)\}}{\pi_0(A \given X)}
\cdot \big(G_\pi(X,Y) - g_\pi(X)\big)\bigg) + 
\gamma \cdot \textnormal{Var}(g_\pi(X)).
\end{align*}
\end{theorem}

\begin{theorem}\label{thm:regret-bound-general}
Suppose Assumptions~\ref{assum:pi0Ydistri},~\ref{assum:essinf},~\ref{assum:optimum},~\ref{assum:convrate_general} hold. 
Additionally assume that $dQ_X/dP_X \le C$, a.s.,
for 
some constants $C>0$, and that 
$\lim_{m,n \rightarrow \infty} m/n = \gamma$.
For any $\beta\in(0,1)$, there exists $N \in \mathbb{N}_+$ such that when $n\ge N$, 
we have with probability at least $1-\beta$ that 
\begin{align*}
\Regret(\hat \pi_{\LN})
\le 
\frac{C (\kappa(\Pi) + \sqrt{\log(1/\beta)}),
}{\sqrt{n}}
\end{align*}
where $C>0$ is a constant independent of $n$ and $\Pi$.
\end{theorem}
\paragraph{Proof Sketch.}
Recall that 
\[
G_\pi(x,y) = 
\bmalpha^*_\pi(x)\exp\Big(-\frac{y + \bmeta^*_\pi(x)}{\bmalpha_\pi^*(x)}\Big) + \bmeta^*_\pi(x) + \bmalpha_\pi^*(x)\delta 
~\text{ and }~
g_\pi(x) = \EE\big[G_\pi(X,Y(\pi(X))) \given X = x\big] 
\]
It can be checked that  
\begin{align*}
\eqref{eqn:x-shift-obj} & = 
\EE_{P}\Big[\frac{r(X)\mathbbm{1}\{A = \pi(X)\}}
{\pi_0(A\given X)}G_\pi(X,Y) \Big]\\
& = \EE_{P}\Big[\frac{r(X)\mathbbm{1}\{A = \pi(X)\}}
{\pi_0(A\given X)}\cdot 
\big(G_\pi(X,Y) - g_\pi(X)\big)\Big] 
+ \EE_{Q_X}\big[g_\pi(X)\big].
\end{align*}
So if we define 
$\bar{\mathcal{V}}_\delta(\pi) =-\frac{1}{K}\sum^K_{k=1} \bar{\mathcal{V}}^{(k)}_\delta(\pi)$, with 
\begin{align*}
\bar{{\mathcal{V}}}^{(k)}_\delta(\pi) 
& = \frac{1}{|\cD^{(k)}|}\sum_{i \in \cD^{(k)}} 
\frac{r(X_i) \mathbbm{1}\{A_i = \pi(X_i)\}}
{\pi_0(A_i \given X_i)}
\cdot \big(G_\pi(X_i,Y_i) - g_\pi(X_i)\big)
+\frac{1}{|\tilde{\cD}^{(k)}|} 
\sum_{i\in \tilde{\cD}^{(k)}}
g_\pi(X_i),
\end{align*}
we have by the central limit theorem that 
\begin{align*}
& \sqrt{n}\Big(\bar{\mathcal{V}}_\delta(\pi)
- \mathcal{V}_\delta(\pi)\Big) 
\stackrel{\textnormal{d}}{\rightarrow} N(0,\sigma^2_\pi).
\end{align*}
As in the proof of Theorem~\ref{thm:asymconv}, 
we can decompose the difference between 
$\bar{\mathcal{V}}_\delta^{(k)}(\pi)$ 
and $\hat{\mathcal{V}}_\delta^{(k)}(\pi)$ as follows: 
\begin{align}\label{eq:x-shift-decomp}
\hat{\mathcal{V}}^{(k)}_\delta(\pi) - 
\bar{\mathcal{V}}^{(k)}_\delta(\pi) = 
& \frac{1}{|\cD^{(k)}|} \sum_{i\in \cD^{(k)}}
\frac{r(X_i)\mathbbm{1}\{A_i = \pi(X_i)\}}{\pi_0(X_i)}
\big(\hat G^{(k)}_\pi(X_i,Y_i) - G_\pi(X_i,Y_i)\big)\notag \\
& + \frac{1}{|\cD^{(k)}|}\sum_{i\in \cD^{(k)}} \mathbbm{1}\{A_i = \pi(X_i)\}
\Big(\frac{\hat r(X_i)}{\hat \pi_0(X_i)} -\frac{r(X_i)}{\pi_0(X_i)}\Big)
\big(\hat G^{(k)}_\pi(X_i,Y_i) - \bar g^{(k)}_\pi(X_i)\big)\notag \\
& + \frac{1}{|\cD^{(k)}|}\sum_{i\in \cD^{(k)}} \mathbbm{1}\{A_i = \pi(X_i)\}
\Big(\frac{\hat r(X_i)}{\hat \pi_0(X_i)} -\frac{r(X_i)}{\pi_0(X_i)}\Big)
\big(\bar g^{(k)}_\pi(X_i) - \hat g_\pi(X_i)\big) \notag \\
& - \frac{1}{|\cD^{(k)}|} \sum_{i \in \cD^{(k)}}
\frac{r(X_i) \mathbbm{1}\{A_i = \pi(X_i)\}}{\pi_0(X_i)}
(\hat g^{(k)}_\pi(X_i) - g_\pi(X_i)) +\frac{1}{|\tilde{\cD}^{(k)}|}
\sum_{i\in \tilde{\cD}^{(k)}} (\hat g^{(k)}_\pi(X_i) - g_\pi(X_i)).
\end{align}
Following almost the same steps in the proof of Theorem~\ref{thm:asymconv}, 
we can show that   
\begin{align*}
\eqref{eq:x-shift-decomp} = & O_P\Big(\|\hat \bmtheta_\pi^{(k)} 
- \bmtheta^*_\pi\|_{L_2(P_{X\given A=\pi(X)})}^2\Big) +
O_P\bigg(\Big\|\frac{\hat r^{(k)}}{\hat \pi^{(k)}} - \frac{r}{\pi}\Big\|_{L_2(P_{X\given A = \pi(X)})} \cdot n^{-1/2}\bigg)\\
& \qquad  + O_P\bigg(\Big\|\frac{\hat r^{(k)}}{\hat \pi_0^{(k)}} - \frac{r}{\pi}\Big\|_{L_2(P_{X\given A=\pi(X)})}
\cdot \|\hat g^{(k)}_\pi - g_\pi\|_{L_2(P_{X\given A = \pi(X)})}
\bigg) + O_P\big(\|\hat g^{(k)}_\pi - g_\pi\|_{L_2(P_{X\given A = \pi(X)})} \cdot n^{-1/2}\big) \\ 
= &o_P(n^{-1/2}). 
\end{align*}
Taking a union bound over $k\in[K]$, we can conclude 
 $\hat {\mathcal{V}}_\delta(\pi) = \bar{\mathcal{V}}_\delta(\pi) +  o_P(n^{-1/2})$, and 
 therefore $\hat{\mathcal{V}}_\delta(\pi)$ is asymptotically normal.
The proof of Theorem~\ref{thm:regret-bound-general} follows similarly.


\section{Proof of technical lemmas}
\subsection{Proof of Lemma~\ref{lemma:regularity}}\label{appx:regularity-loss}
\paragraph{Proof of (1).}
Given $\theta$, recall that our loss function is
\begin{align*}
\ell(x,y;\theta)=\al \exp\Big(-\frac{y+\eta}{\al}-1\Big)+\eta+\al\dd.
\end{align*}
By the strong duality, $\EE[\ell(X,Y(\pi(X));\theta) \given X]$ is convex in $\theta$;
by Proposition~\ref{prop:underline-al>0}, the first-order condition of convex optimization
problem implies
\$ 
\nabla_{\theta} \EE\Big[\ell\big(x,Y(\pi(x));\bmtheta^*_\pi(x)\big) \biggiven X=x\Big]=0.
\$
Meanwhile, we can compute the gradient of $\ell(x,y;\theta)$ as
\# \label{eq:gradient}
& \frac{\partial}{\partial \alpha} \ell(x,y; \theta) = 
\Big(1 + \frac{y+\eta}{\alpha}\Big) \cdot 
\exp\Big(-\frac{y+\eta}{\alpha}-1\Big) + \delta,\notag\\
& \frac{\partial}{\partial_{\eta}} \ell(x,y; \theta) = 1 -  \exp\Big(-\frac{y+\eta}{\alpha}-1\Big). 
\#
For any $a$ such that $|a - \bmalpha^*_\pi(x)| \le \bmalpha^*_\pi(x)$, we have
\$
\bigg|\frac{\partial}{\partial \alpha} \ell(x,y;(a,\bmeta^*_\pi(x)))\bigg| 
& \le 
\Big(1 + \frac{2(\bar y+\bar \eta)}{\underline{\alpha}}\Big) \cdot 
\exp\Big(\frac{2(\bar y+\bar \eta)}{\underline{\alpha}}-1\Big) + \delta < \infty.
\$ 
By the mean value theorem and the dominated convergence theorem,
we can change the order of expectation and taking limits and therefore 
\$ 
\EE\bigg[\frac{\partial}{\partial \alpha}
\ell\big(x,Y(\pi(x));\bmtheta^*_\pi(x)\big)
\biggiven X=x\bigg]=
\frac{\partial}{\partial \alpha} 
\EE\Big[\ell\big(x,Y(\pi(x));\bmtheta^*_\pi(x)\big) \biggiven X=x\Big]=0.
\$ 
Similarly, since $\frac{\partial}{\partial_{\eta}} \ell(x,y; (\bmalpha^*_\pi(x),\eta))$
is non-decreasing in $\eta$, for $|\eta - \bmeta^*_\pi(x)|\le 1$, 
\$ 
\bigg|\frac{\partial}{\partial_{\eta}} \ell(x,y; (\bmalpha^*_\pi(x),\eta))\bigg| \le
\max\Bigg\{ \bigg|\frac{\partial}{\partial_{\eta}} \ell(x,y;(\bmalpha^*_\pi(x), \bmeta^*_\pi(x) + 1))\Big|,
\Big|\frac{\partial}{\partial_{\eta}} \ell(x,y; (\bmalpha^*_\pi(x),\bmeta^*_\pi(x) - 1))\bigg| \Bigg\},
\$
with the right-hand side being integrable under $P_{Y \given X}$. 
Agian by the mean-value theorem and the dominated convergence theorem,
\$
\EE\bigg[\frac{\partial}{\partial \eta}\ell(x,Y(\pi(x));\bmtheta^*_\pi(x)) \Biggiven X=x\bigg]=
\frac{\partial}{\partial \eta} 
\EE\bigg[\ell(x,Y(\pi(x));\bmtheta^*_\pi(x)) \Biggiven X=x\bigg]=0.
\$
We have thus completed the proof part (1) of Lemma~\ref{lemma:regularity}.

\paragraph{Proof of (2).}
We now compute the Hessian of $\ell(x,y;\theta)$:
\$ 
& \frac{\partial^2}{\partial \alpha^2} \ell(x,y; \theta) 
= \frac{(y + \eta)^2}{\alpha^3} \exp\Big(-\frac{y+\eta}{\alpha}-1\Big),\\
& \frac{\partial^2}{\partial \alpha \partial \eta} \ell(x,y; \theta) 
= -\frac{y+\eta}{\alpha^2}\exp\Big(-\frac{y+\eta}{\alpha}-1\Big),\\ 
& \frac{\partial^2}{\partial \eta^2} \ell(x,y; \theta) 
= \frac{1}{\alpha} \exp\Big(-\frac{y+\eta}{\alpha}-1\Big).
\$
By the Taylor expansion,  
\$ 
& \ell(x,y;\theta) - \ell(x,y;\bmtheta^*_\pi(x)) =
\nabla \ell(x,y;\bmtheta^*_\pi(x))^\top(\theta - \theta^*_\pi(x))
+ \frac{1}{2}(\theta - \bmtheta^*_\pi(x))^\top \nabla^2 \ell(x,y;\tilde{\theta})(\theta - \bmtheta^*_\pi(x)),\\
\Rightarrow ~&
\big|\ell(x,y;\theta) - \ell(x,y;\bmtheta^*_\pi(x))
-\nabla \ell(x,y;\bmtheta^*_\pi(x))^\top(\theta - \bmtheta^*_\pi(x))\big| \\ 
& \qquad \qquad \qquad \qquad \le \frac{1}{2}\Big(\frac{(y + \tilde{\eta})^2}{\tilde{\alpha}^3}
+\frac{1}{\tilde{\alpha}}\Big)
\exp\Big(-\frac{y+\tilde{\eta}}{\tilde{\alpha}}-1\Big)\|\theta - \bmtheta^*_\pi(x)\|_2^2,
\$
where $\tilde{\theta} = t \theta + (1-t)\bmtheta^*_\pi(x)$ for some $t \in [0,1]$ and  
the last step is because 
\$
\big\|\nabla^2 \ell(x,y;\tilde{\theta})\big\|_{\text{op}} 
\le \Big(\frac{(y + \tilde{\eta})^2}{\tilde{\alpha}^3}
+\frac{1}{\tilde{\alpha}}\Big)
\exp\Big(-\frac{y+\tilde{\eta}}{\tilde{\alpha}}-1\Big)
\$
Let $\xi = \min(\underline{\alpha}, \bar{\eta})/2$. 
For any $\theta$ such that $\|\theta - \bmtheta^*_\pi(x)\|_2 \le \xi$, we also have
$|\tilde{\alpha} - \bmalpha^*_\pi(x)| \le \xi$ and 
$|\tilde{\eta} - \bmeta^*_\pi(x)|\le \xi$. Then
\$ 
\frac{1}{2}\Big(\frac{(y + \tilde{\eta})^2}{\tilde{\alpha}^3}+\frac{1}{\tilde{\alpha}}\Big)
\exp\Big(-\frac{y+\tilde{\eta}}{\tilde{\alpha}}-1\Big) 
\le \Big(\frac{8\bar y^2 + 8 \bar \eta^2}{\underline{\alpha}^3} + \frac{2}{\underline{\alpha}}\Big)
\cdot \exp\Big(\frac{2\bar{y}+4\bar{\eta}}{\underline{\alpha}}-1\Big). 
\$ 
Letting the right-hand side be $\bar{\ell}(x,y)$, 
we have thus completed the proof of (2).

\paragraph{Proof of (3).}
By the Taylor expansion,
\$ 
\ell(x,y;\theta) - \ell(x,y;\bmtheta^*_\pi(x)) = 
\nabla \ell(x,y;\tilde{\bmtheta})^\top\big(\bmtheta(x) - \bmtheta^*_\pi(x)\big),
\$ 
where $\tilde{\bmtheta} = t\bmtheta(x) + (1-t)\bmtheta_\pi^*$ for some $t \in [0,1]$.
Let $\xi_1 = \min(\underline{\alpha}, \bar \eta)/2$. When $\|\bmtheta - \bmtheta_\pi^*\|_{L_{\infty}} \le \xi_1$, 
we have $|\tilde{\alpha} - \bmalpha^*_\pi(x)| \le \xi_1$ and 
$|\tilde{\eta} - \bmeta^*_\pi(x)|\le \xi_1$.
Plugging the expressions of the gradient in Equation~\eqref{eq:gradient}, we have 
\$ 
& \big[ \ell(x,y;\bmtheta(x)) - \ell(x,y;\bmtheta^*_\pi(x))\big]^2
=  \big[\nabla \ell(x,y;\tilde{\bmtheta}(x))^\top(\bmtheta(x) - \bmtheta^*_\pi(x))\big]^2\\
\le \, &\Bigg\{ 
\bigg[\Big(1 + \frac{y + \tilde{\bmeta}(x)}{\tilde{\bmalpha}(x)}\Big) 
\exp\Big(-\frac{y+\tilde{\bmeta}(x)}{\tilde{\bmalpha}(x)}-1\Big) + \delta\bigg]^2  
+ \bigg[1 - \exp\Big(-\frac{y+\tilde{\bmeta}(x)}{\tilde{\bmalpha}(x)}-1\Big)\bigg]^2 
\Bigg\}\cdot 
\big\|\bmtheta(x) - \bmtheta^*_\pi(x)\big\|_2^2\\
\le \,& C(\bar y, \bar{\alpha}, \underline{\alpha}, \bar{\eta}, \delta) \cdot
\big\|\bmtheta(x) - \bmtheta^*_\pi(x)\big\|_2^2,
\$
where $C(\bar y, \bar{\alpha}, \underline{\alpha}, \bar{\eta}, \delta)$
is a function of $(\bar y, \bar{\alpha}, \underline{\alpha}, \bar{\eta},  \delta$).
Taking the expectation over $P_{X,Y\given A = \pi(X)}$, we have 
\$ 
\big\|\ell(X,Y;\bmtheta(X)) - \ell(X,Y;\bmtheta^*_\pi(X)) \big\|_{L_2(P_{X,Y \given A = \pi(X)})}
\le C(\bar{y},\bar{\alpha}, \underline{\alpha}, \bar{\eta}, \delta)
\cdot \big\|\bmtheta - \bmtheta^*\big\|_{L_2(P_{X\given A = \pi(X)})},
\$
completing the proof of (3).

\subsection{Proof of Lemma~\ref{lemma:bound-rademacher}}
We first introduce the $\ell_2$ distance on the policy space $\Pi$, 
as well as the corresponding covering number.
\begin{definition}
Given a function $h$ and a set of realized data $z_1,\ldots,z_n$,
\begin{enumerate}
\item[(1)] the $\ell_2$ distance between two policies $\pi_1,\pi_2\in \Pi$ 
with respect to $\{z_1,\ldots,z_n\}$
is defined as 
\$ 
\ell_2\big(\pi_1,\pi_2;\{z_1,\ldots,z_n\}\big) =  
\sqrt{\frac{\sum_{i=1}^n\big(h(z_i,\pi_1(x_i)\big) - 
h(z_i;\pi_2(x_i))\big)^2}{4 \sum^n_{i=1} c_i(z_i)^2}}.
\$
\item [(2)] $N_2(\gamma,\Pi;\{z_1,\ldots,z_n\})$ is the minimum number 
of policies needed to $\gamma$-cover $\Pi$ under $\ell_2$ with respect $\{z_1,\ldots,z_n\}$.
\end{enumerate}
\end{definition} 
Under the $\ell_2$ distance, we define a sequence of approximation operators 
$A_j: \Pi \mapsto \Pi$ for $j \in [J]$, where $J = \lceil \log_2 n\rceil$. 
Specifically, for any $j =  0,1,\ldots,J$, 
let $S_j$ be the set of policies that $2^{-j}$-covers $\Pi$ and satisfies 
$|S_j| = N_2(2^{-j},\Pi;\{Z_1,\ldots,Z_n\})$. Specially, $S_0 = \{\bar{\pi}\}$, 
with $\bar\pi$ is an arbitrary policy in $\Pi$ --- this is a valid choice since for 
any $\pi \in \Pi$, 
\$
\ell_2(\pi,\bar{\pi};\{z_1,\ldots,z_n\})
= \sqrt{\frac{\sum_{i=1}^n\big(h(z_i,\pi(x_i)) - h(z_i,\bar{\pi}(x_i))\big)^2}
{4\sum^n_{i=1}c_i(z_i)^2}} \le 1.
\$ 
We shall let $\Lambda = 2 \sqrt{\sum^n_{i=1} c_i(z_i)^2}$
to denote the normalization factor.
The approximation operators 
are defined in a backward manner: for any $\pi \in \Pi$,
\begin{itemize}
\item [(1)]  define 
$
A_J[\pi] = \argmin{\pi' \in S_J}~\ell_2\big(\pi,\pi';\{z_1,\ldots,z_n\}\big);
$ 
\item [(2)] for $j = J-1,\ldots,0$, define 
\$ 
A_j[\pi] = \argmin{\pi' \in S_j}~\ell_2\big(A_{j+1}[\pi],\pi';\{z_1,\ldots,z_n\}\big).
\$
\end{itemize}
Using the sequential approximation operators,
we decompose the inner expectation term in~\eqref{eq:symmetrization} (Rademacher complexity) as
\$ 
& \EE_\epsilon\Bigg[\sup_{\pi \in \Pi} 
\bigg|\frac{1}{n}\sum_{i \in [n]}\epsilon_i h(Z_i,\pi(X_i))\bigg|\Bigg] \\
\le\,&  
\EE_\epsilon\Bigg[\sup_{\pi \in \Pi} \bigg|\frac{1}{n}\sum_{i \in [n]}
\epsilon_i \big[h(Z_i,\pi(X_i)) - h(Z_i,A_J[\pi](X_i)) \big]\bigg|\Bigg] \\ 
& \qquad + \EE_\epsilon\Bigg[\sup_{\pi \in \Pi} \bigg| \sum_{j=1}^J \frac{1}{n}\sum_{i \in [n]}
\epsilon_i \big[h(Z_i,A_j[\pi](X_i)) - h(Z_i,A_{j-1}[\pi](X_i)) \big]\bigg| \Bigg]\\
& \qquad +\EE_\epsilon\Bigg[\sup_{\pi \in \Pi} \bigg|\frac{1}{n}\sum_{i\in [n]} 
\epsilon_i h(Z_i,A_0[\pi](X_i))\bigg|\Bigg]\\ 
=: \, &\Xi_1 + \Xi_2 + \Xi_3.
\$
For any $\pi \in \Pi$, by the Cauchy-Schwarz inequality, 
\$ 
& \sup_{\pi \in \Pi} \bigg|\frac{1}{n}\sum_{i \in [n]}
\epsilon_i \big[h(z_i,\pi(x_i)) - h(z_i,A_J[\pi](x_i)) \big]\bigg|\\
\le \, &\frac{1}{n}\sqrt{n \sum_{i\in[n]}
\big(h\big(z_i,\pi(x_i)) - h(z_i,A_J[\pi](x_i))\big)^2}\\
= \, & \frac{\Lambda}{\sqrt{n}} \cdot \ell_2(\pi,A_J(\pi);\{z_1,\ldots,z_n\}) \\
\le \, &\frac{\Lambda}{\sqrt{n}} 2^{-J} \le  \frac{\Lambda}{n^{3/2}}, 
\$
where the second-to-last step is because $A_J(\pi)$ is $2^{-J}$-close to $\pi$
and the last step is by the choice of $J$.
As a result the above derivation, $\Xi_1 \le \Lambda/n^{3/2}$.

Next, for any $j = 1,\ldots,J$ we use $P_{j}$ to denote the projection of 
projecting a policy to $S_{j}$, i.e., $A_{j-1}[\pi] = P_{j-1}[A_j[\pi]]$. 
Once $A_j(\pi)$ is determined, $A_{j-1}(\pi)$ is also determined. 
For any $s > 0$, 
\$ 
& \PP_\epsilon\Bigg(\sup_{\pi \in \Pi} \bigg|\frac{1}{n}\sum_{i \in [n]}
\epsilon_i \big[h(z_i,A_j[\pi](x_i)) - h(z_i;A_{j-1}[\pi](x_i)) \big]\bigg|  \ge s \Bigg)\\
\le & \, \sum_{\pi' \in S_j}
\PP_\epsilon\Bigg(\bigg|\frac{1}{n}\sum_{i \in [n]}
\epsilon_i \Big[h(z_i,\pi'(x_i)) - 
h(z_i,P_{j-1}[\pi'](x_i)) \Big]\bigg|  \ge s\bigg)\\
\le  & \, \sum_{\pi' \in S_j}
2 \cdot \exp\Bigg(-\frac{2n^2s^2}
{\sum^n_{i=1}\big[h(z_i,\pi'(x_i)) - h(z_i,P_{j-1}[\pi'](x_i))\big]^2}\Bigg)\\
= & \, \sum_{\pi' \in S_j} 2 \cdot \exp\Bigg(
-\frac{2n^2s^2}{\Lambda^2\ell_2(\pi',P_{j-1}(\pi'); z)^2}\Bigg)\\
\le &\, 2 N_2(2^{-j},\Pi;Z) \cdot 
\exp\Big(-\frac{n^2s^2}{\Lambda^2 2^{-2j+1}}\Big),
\$
we $z$ is a shorthand for $\{z_1,\ldots,z_n\}$.
For any $j = 1,\ldots,J$ and $m \in \mathbb{N}$, take 
\$
s_{j,m} = \frac{\Lambda}{n2^{j-1/2}} 
\sqrt{\log\big(N_2(2^{-j},\Pi;Z)\cdot 2^{m+1} j^2)}.
\$
For a fixed $m$, with a union bound over $j = 1,\ldots,J$ we have that 
\$ 
& P_\epsilon\bigg(\sup_{\pi \in \Pi} \bigg|\sum^J_{j=1}\frac{1}{n}\sum_{i \in [n]}
\epsilon_i \big[h(z_i,A_j[\pi](x_i)) - h(z_i,A_{j-1}[\pi](x_i)) \big]\bigg|  \ge 
\sum^J_{j=1} s_{j,m}\bigg)\\
\le \, & \sum_{j=1}^J P_\epsilon\bigg(\sup_{\pi \in \Pi} 
\bigg|\frac{1}{n}\sum_{i \in [n]}
\epsilon_i \big[h(z_i,A_j[\pi](x_i)) - h(z_i,A_{j-1}[\pi](x_i))\big]\bigg|  \ge s_{j,m}\bigg)
\le  \sum^J_{j=1} \frac{1}{j^2 2^m} \le \frac{1}{2^{m-1}}. 
\$
To proceed,  we shall use the following 
lemma, whose proof is deferred to Appendix~\ref{appx:dist}. 
\begin{lemma}\label{lemma:dist}
 For any realization $z_1,\ldots,z_n$ and $\gamma >0$, there is 
 $N_2(\gamma,\Pi;z_1,\ldots,z_n) \le N_H(\gamma^2,\Pi)$. 
\end{lemma}
By Lemma~\ref{lemma:dist}, for any $m \in \mathbb{N}_+$, 
\$ 
\sum^J_{j=1}s_{j,m} = \,& \sum^J_{j=1} 
\frac{\Lambda}{2^{j-1/2}{n}} 
\sqrt{\log\big(N_2(2^{-j},\Pi;Z)\cdot 2^{m+1}j^2\big)}\\
\le\, & \sum^J_{j=1} \frac{\Lambda}{2^{j-1/2} {n}} 
\sqrt{\log(N_H(2^{-2j},\Pi))+ (m+1)\log 2 + 2\log(j)} \\
\stackrel{\text{(i)}}{\le} \, 
&\frac{2 \Lambda}{n} \sum^J_{j=1}
2^{-j} \cdot \Big(\sqrt{\log(N_H\big(2^{-2j},\Pi)\big)} + \sqrt{m+1} +\sqrt{2\log(j)}\Big)\\
\stackrel{(ii)}{\le} \, & \frac{4\Lambda}{{n}} 
\big(\kappa(\Pi) + \sqrt{m+1} + 1\big) = : u_m,
\$
where step (i) uses $\sqrt{a+b+c} \le \sqrt{a} + \sqrt{b}+\sqrt{c}$ for $a,b,c\ge 0$; 
step (ii) uses the definition of $\kappa(\Pi)$. 
Then 
\$ 
\Xi_2 & = \EE_\epsilon\Bigg[\sup_{\pi \in \Pi} \bigg| \sum_{j=1}^J 
\frac{1}{n}\sum_{i \in [n]}
\epsilon_i \Big[h(z_i,A_j[\pi](x_i)) - 
h(z_i,A_{j-1}[\pi](x_i)) \Big]\bigg| \Bigg] \notag \\
& = \int_0^\infty \PP_\epsilon\Bigg(\sup_{\pi \in \Pi} \bigg| 
\sum_{j=1}^J \frac{1}{n}\sum_{i \in [n]}
\epsilon_i \Big[h(z_i,A_j[\pi](x_i)) 
- h(z_i,A_{j-1}[\pi](x_i)) \Big]\bigg| > s\Bigg) \, \textnormal{d}s\notag \\
& \le  u_1 + \sum^\infty_{k=1} (u_{k+1} - u_k) \cdot 2^{-k+1} \notag\\
& = \frac{4\Lambda}{{n}} \cdot \Big(\kappa(\Pi) + \sqrt{2} +1
+ \sum^\infty_{k=1} (\sqrt{k+2} - \sqrt{k+1})\cdot 2^{-k+1}\Big) 
\le\frac{4 \Lambda}{{n}} \cdot \big(\kappa(\Pi) +7 \big). 
\$
Finally, we consider $\Xi_3$. Recall that $S_0 = \{\bar{\pi}\}$, and therefore 
\$ 
\Xi_3 = \EE_\epsilon\Bigg[\bigg|\frac{1}{n}\sum_{i \in [n]}
\epsilon_i h(z_i,\bar{\pi}(x_i))\bigg|\Bigg] 
& = \int^{\infty}_{0} \PP_\epsilon\Bigg(\bigg|\frac{1}{n}\sum_{i \in [n]}
\epsilon_i h(z_i,\bar{\pi}(x_i))\bigg| > s\Bigg) \, \textnormal{d}s\\
& \le\int^{\infty}_{0} 2\exp\Big(-\frac{n^2s^2}{\Lambda^2}\Big) \, \textnormal{d}s 
= \frac{3\Lambda}{n}.
\$
Putting everything together, 
\$ 
\EE_\epsilon\bigg[
\Big|\frac{1}{n}\sum^n_{i=1} \epsilon_i h\big(x_i,a_i,y_i,\pi(x_i)\big)\Big|\bigg]
& \le \frac{\Lambda}{n}\cdot(4\kappa(\Pi) + 32)\\
& = \frac{2\sqrt{\sum^n_{i=1}c_i(z_i)^2 }}{n}(4\kappa(\Pi)+32).
\$


\subsection{Proof of Lemma~\ref{lemma:dist}}
\label{appx:dist}
Fix $\gamma >0$. If $N_H(\gamma^2,\Pi) = \infty$, the lemma 
is trivially true. Otherwise, let $N_0 = N_H(\gamma^2;\Pi)$.
For any realization $z_1,\ldots,z_n$, define 
\$ 
(\pi^*_{i,1},\pi_{i,2}^*) = \argmax{\pi_1,\pi_2} \big\{|h(z_i,\pi_1(x_i)) - h(z_i,\pi_2(x_i))|\big\}.
\$
Implicitly, $(\pi^*_{i,1},\pi^*_{i,2})$ depends on $z_i$.
For an arbitrary positive integer $m$ and $i \in [n]$, we define 
\$ 
n_i = \Big\lceil \frac{m}{\Lambda^2n} \big\{h(z_i,\pi^*_{i,1}(x_i)) 
- h(z_i,\pi^*_{i,2}(x_i))\big\}^2 \Big\rceil,
\$
where we recall that 
$\Lambda^2 = 4\sum^n_{i=1} c_i(z_i)^2$.
We then construct a new set of data 
\$ 
\{\tilde{z}_1,\ldots, \tilde{z}_N\} = \{z_1,\ldots,z_1,
z_2,\ldots,z_2,\ldots,z_{n},\ldots, z_n\},
\$
where $z_i$ appears $n_i$ times and 
\$
N = \sum^n_{i=1} n_i = 
\sum^n_{i=1} \Big\lceil \frac{m}{\Lambda^2} \big\{h(z_i,\pi^*_{i,1}(x_i)) 
- h(z_i,\pi^*_{i,2}(x_i))\big\}^2 \Big\rceil
\le m+n. 
\$
By definition, there exists a policy set 
$S_0$ to be a $\gamma^2$-cover of $\Pi$ 
the Hamming distance with respect to 
$\tilde x:=\{\tilde{x}_1,\ldots,\tilde{x}_N\}$
such that $|S_0| = N_0$. As a result, for any 
$\pi \in \Pi$, there exists $\pi' \in S_0$ such that
$H(\pi,\pi';\tilde x) \le \gamma^2$. On the other hand, 
\$ 
H(\pi,\pi';\tilde x) & = 
\frac{1}{N}\sum^N_{i=1} \indc\{\pi(\tilde{x}_i) \neq \pi'(\tilde{x}_i)\}\\
& \stackrel{\text(i)}{=} \frac{1}{N}\sum^n_{i=1} n_i \indc\{\pi(x_i) \neq \pi'(x_i)\}\\
& \ge \frac{1}{N} \sum^n_{i=1} \frac{m}{\Lambda^2}
\big\{h(z_i,\pi^*_{i,1}(x_i)) - h(z_i,\pi^*_{i,2}(x_i))\big\}^2 
\cdot \indc\{\pi(x_i) \neq \pi'(x_i)\}\\
& \stackrel{\text{(ii)}}{\ge}  \frac{1}{N} \sum^n_{i=1} \frac{m}{\Lambda^2}
\big\{h(z_i,\pi(x_i)) - h(z_i,\pi'(x_i))\big\}^2 
\cdot \indc\{\pi(x_i) \neq \pi'(x_i)\}\\
& \stackrel{\text{(iii)}}{=}  \frac{1}{N} \sum^n_{i=1} \frac{m}{\Lambda^2}
\big\{h(z_i,\pi(x_i)) - h(z_i,\pi'(x_i))\big\}^2. 
\$
Above, step (i) and (ii) follow from the choice of $\tilde z$
and $(\pi^*_{i,1},\pi^*_{i,2})$, respectively; step (iii) is because when 
$\pi(x_i) = \pi'(x_i)$, $h(z_i,\pi(x_i)) = h(z_i,\pi(x_i'))$.
By the definition of the $\ell_2$ distance and that $N \le m+n$, 
we further have 
\$ 
\gamma^2 \ge H(\pi,\pi';\tilde x) \ge \frac{m}{(m+n)}\ell^2(\pi,\pi';z). 
\$ 
Since $m$ is arbitrary, we take $m$ to infinity and have $\ell_2(\pi,\pi';z) \le \gamma$.
By definition, $S_0$ is a $\gamma$-cover of $\Pi$ under $\ell_2$ 
with respect to $z_1,\ldots,z_n$, and therefore $N_2(\gamma,\Pi;z_1,\ldots,z_n) 
\le N_H(\gamma^2,\Pi)$.

\subsection{Proof of Lemma~\ref{lemma:g_property}}
\label{appx:proof_g_property}
By~\citet[Lemma B12]{yang2022toward}, $g_\delta(q)$ is differentiable in $q$, and 
\$ 
g_\delta'(q) = - \frac{\partial_q D_{\kl}(g(q) \,\|\, q)}
{\partial_p D_{\kl}(g(q) \,\|\, q)} 
= \frac{g(q)/q - (1-g(q))/(1-q)}{\log\big(g(q)/(1-g(q))\big)- 
\log\big(q/(1-q)\big)}. 
\$
Also by~\citet[Lemma B12]{yang2022toward}, $g_\delta(q)$ is convex in $q$, 
so $g_\delta'(q)$ is increasing in $q$. Since $q \in [0.4,0.6]$, 
$g'_\delta(q) \ge g'_\delta(0.4)$. From the dual form, we can check that 
$g(0.4)\ge 0.1$. Plugging in $q = 0.4$, we have 
\$ 
g'_\delta(0.4) = \frac{\frac{g(0.4)}{0.4}+ \frac{g(0.4)}{0.6} - 5/3}
{\log\big(g(0.4)/(1-g(0.4))\big)- 
\log(2/3)}
= \frac{g(0.4)/0.24 - 5/3}{\log(g(0.4) / (1-g(0.4))) - \log(2/3)}.
\$
Since the function $f(x) = \frac{x/0.24 - 5/3}{\log(x/(1-x))-\log(2/3)}$
is increasing in $x$ for $x \in(0,0.4)$, we conclude that 
\$
g_\delta'(0.4) \ge \frac{1/2.4 - 5/3}{\log(1/9)-\log(2/3)} \ge 1/2,
\$
completing the proof.

\end{document}